%% file: acl_latex.tex
\pdfoutput=1

\documentclass[11pt]{article}

\usepackage[final]{acl}

\usepackage{times}
\usepackage{latexsym}

\usepackage[T1]{fontenc}

\usepackage[utf8]{inputenc}

\usepackage{microtype}

\usepackage{inconsolata}

\usepackage{graphicx}
\input{preamble}
\title{
    \raisebox{-.5ex}{\includegraphics[width=0.55cm]{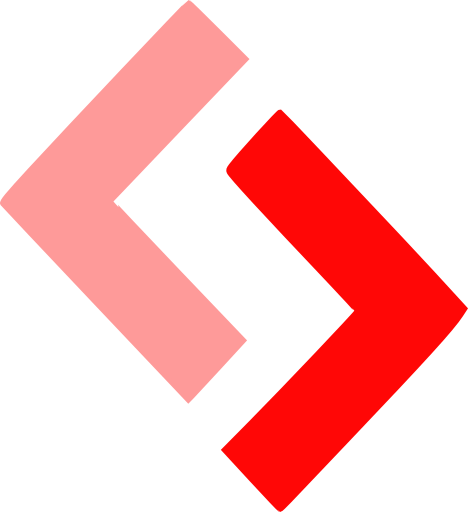}}\,%
    WebMMU: A Benchmark for Multimodal Multilingual Website Understanding and Code Generation
}

\author{First Author \\
  Affiliation / Address line 1 \\
  Affiliation / Address line 2 \\
  Affiliation / Address line 3 \\
  \texttt{email@domain} \\\And
  Second Author \\
  Affiliation / Address line 1 \\
  Affiliation / Address line 2 \\
  Affiliation / Address line 3 \\
  \texttt{email@domain} \\}

\author{\bf Rabiul Awal$^{1,2,3}$\thanks{Co-first authors $^\dagger$ Co-second authors}\hspace{1em} 
    Mahsa Massoud$^{1,2,4}$\footnotemark[1]\hspace{1em} 
    Aarash Feizi$^{1,2,4\dagger}$\hspace{1em} 
    Zichao Li$^{1,2,4\dagger}$ \hspace{1em} \\
    \bf Suyuchen Wang$^{1,2,3}$\hspace{1em} 
    Christopher Pal$^{1,2,6}$\hspace{1em}  
    Aishwarya Agrawal$^{2,3}$\hspace{1em} 
    David Vazquez$^{1}$\hspace{1em}   \\
    \bf Siva Reddy$^{1,2,4}$\hspace{1em} 
    Juan A. Rodriguez$^{1,2,5}$\hspace{1em} 
    Perouz Taslakian$^{1}$\hspace{1em} \\ 
    \bf Spandana Gella$^{1}$\hspace{1em} 
    Sai Rajeswar$^{1,2}$ \\
    $^{1}$ServiceNow, $^{2}$Mila, $^{3}$Université de Montréal, $^{4}$McGill University, \\
    $^{5}$École de Technologie Supérieure (ETS), $^{6}$ Polytechnique Montréal
}

\begin{document}
\maketitle
\input{sec/0_abstract}   
\input{sec/1_intro}
\input{sec/2_related_work}

\input{sec/3_webmmu}

\input{sec/4_experiments}

\input{sec/5_results_and_experiments}
\input{sec/6_conclusion}

\bibliography{custom}

\appendix

\input{sec/X_suppl}

\label{sec:appendix}
\end{document}

%% file: preamble.tex
\usepackage{booktabs}
\usepackage{multirow}
\usepackage{subcaption}
\usepackage{comment}
\usepackage{graphicx}      
\usepackage{booktabs}      
\usepackage{array}         
\usepackage{adjustbox}     
\usepackage{rotating}      
\usepackage{float}         
\usepackage{lscape}        
\usepackage[most]{tcolorbox}
\usepackage{placeins}
\usepackage{mathptmx}  
\usepackage{microtype} 
\usepackage{fontawesome5}
\usepackage{xcolor}
\usepackage{subcaption}
\usepackage{marginnote}  
\usepackage[textsize=tiny]{todonotes}
\usepackage{xspace}
\usepackage{dsfont}
\usepackage{colortbl}
\usepackage{wrapfig}
\usepackage{enumitem}

\newcommand{\benchmark}{WebMMU}

\definecolor{pastelblue}{RGB}{173, 216, 230}   
\definecolor{pastelgreen}{RGB}{144, 238, 144}  
\definecolor{pastelorange}{RGB}{255, 204, 153} 

\newcommand{\mockupNoStyle}{Mockup2Code}
\newcommand{\wqaNoStyle}{WebQA}
\newcommand{\codeeditNoStyle}{Web Code Editing}

\newcommand{\agenticqa}{Agenctic Action}
\newcommand{\reasoningqa}{Multi-step Reasoning}
\newcommand{\generalqa}{General Visual Comprehension}

\definecolor{open_models_below_8B}{RGB}{185, 235, 255}
\definecolor{open_models_below_12B}{RGB}{255, 219, 187}
\definecolor{open_models_over_12B}{RGB}{190, 255, 190}
\definecolor{closed_models}{RGB}{240, 240, 240}

\newcommand{\GPTfouro}{GPT4o-1120}
\newcommand{\oOne}{OpenAI-o1}

\newcommand{\GeminiTwoFlash}{Gemini-2.0-Flash}
\newcommand{\Claude}{Claude-3.5-Sonnet}
\newcommand{\QwenVLSeventyTwoB}{Qwen2VL-72B}
\newcommand{\QwenVLSevenB}{Qwen2VL-7B}

\newcommand{\PixtralTwelveB}{Pixtral-12b}
\newcommand{\InternVLTwoFiveEightB}{Internvl2.5-8b}
\newcommand{\InternVLTwoFiveThirtyEightB}{Internvl2.5-38b}
\newcommand{\PhiThreeFiveVI}{Phi3.5-VI-4b}

\newcommand{\MolmoSevenB}{Molmo-7b}
\newcommand{\FuyuEightB}{Fuyu-8b}

\newcommand{\LlamaThreeVision}{Llama-3.2-11B-Vision}
\newcommand{\UITARS}{UI-Tars-7b}

\newcommand{\FunctionalIcon}{\raisebox{-3pt}{\includegraphics[width=0.45cm]{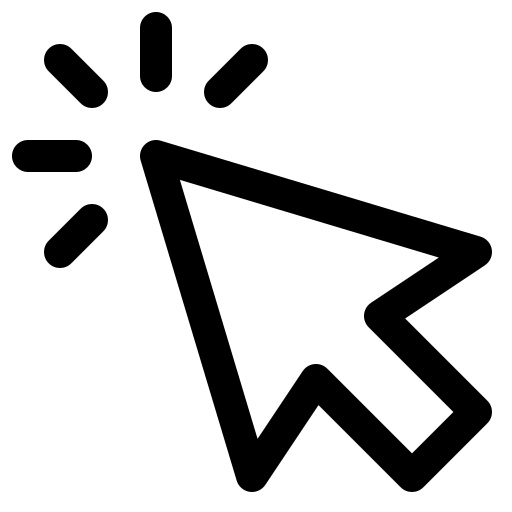}}}
\newcommand{\ReasoningIcon}{\raisebox{-3pt}{\includegraphics[width=0.45cm]{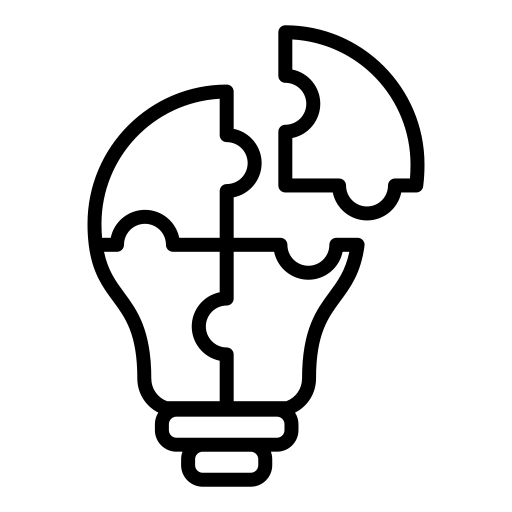}}}
\newcommand{\UnderstandingIcon}{\raisebox{-3pt}{\includegraphics[width=0.45cm]{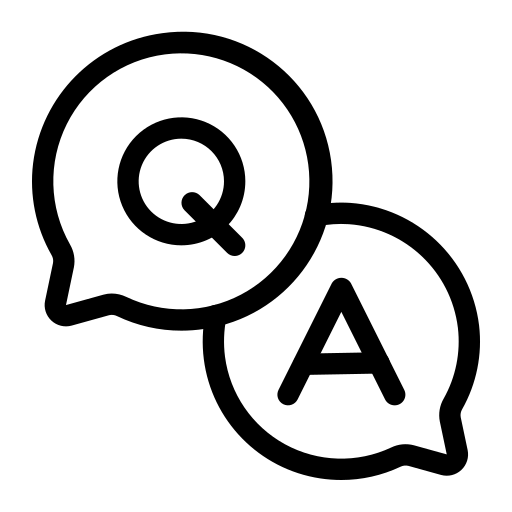}}}

\usepackage{setspace}

\tcbset{
  taskbox/.style={
    colback=green!5,
    colframe=green!10,
    boxrule=0pt,
    arc=6pt,
    auto outer arc,
    boxsep=6pt,
    left=6pt,
    right=6pt,
    enlarge left by=0mm,
    width=\textwidth,
    sharp corners=southwest,
  }
}

\definecolor{prompt1-frame}{RGB}{137, 137, 90}

\tcbset{
  mypromptstyle/.style={
    colback=white,
    colframe=prompt1-frame,
    arc=2pt,
    left=4pt, right=4pt, top=4pt, bottom=4pt,
    before skip=10pt, after skip=10pt,
    fonttitle=\bfseries,
    coltitle=white
  }
}

\definecolor{ecologygreen}{RGB}{20, 120, 120}
\definecolor{lightblue}{RGB}{240, 248, 248}
\definecolor{codebg}{RGB}{248, 248, 248}

%% file: sec/0_abstract.tex
\begin{abstract}
We present \benchmark{}, a multilingual benchmark that evaluates three core web tasks: (1) website visual question answering, (2) code editing involving HTML/CSS/JavaScript, and (3) mockup-to-code generation. Unlike prior benchmarks that treat these tasks separately, \benchmark{} unifies them using expert-annotated, real-world web data to assess models’ abilities in complex multi-step reasoning, precise element grounding, and functional UI comprehension and coding. Our evaluation shows that while multimodal large language models (MLLMs) perform well on basic information extraction, they struggle with reasoning and grounding, editing code to preserve functionality, and generating design-to-code that maintains hierarchy and supports multilingual content. These findings reveal key limitations in current MLLMs and underscore the need for improved multimodal and cross-lingual reasoning to build future web agents capable of automating diverse web development tasks. The dataset is publicly available at \href{https://webmmu-paper.github.io/}{webmmu-paper.github.io}.
\end{abstract}

\begin{figure*}[t]
    \centering
    \includegraphics[width=\textwidth]{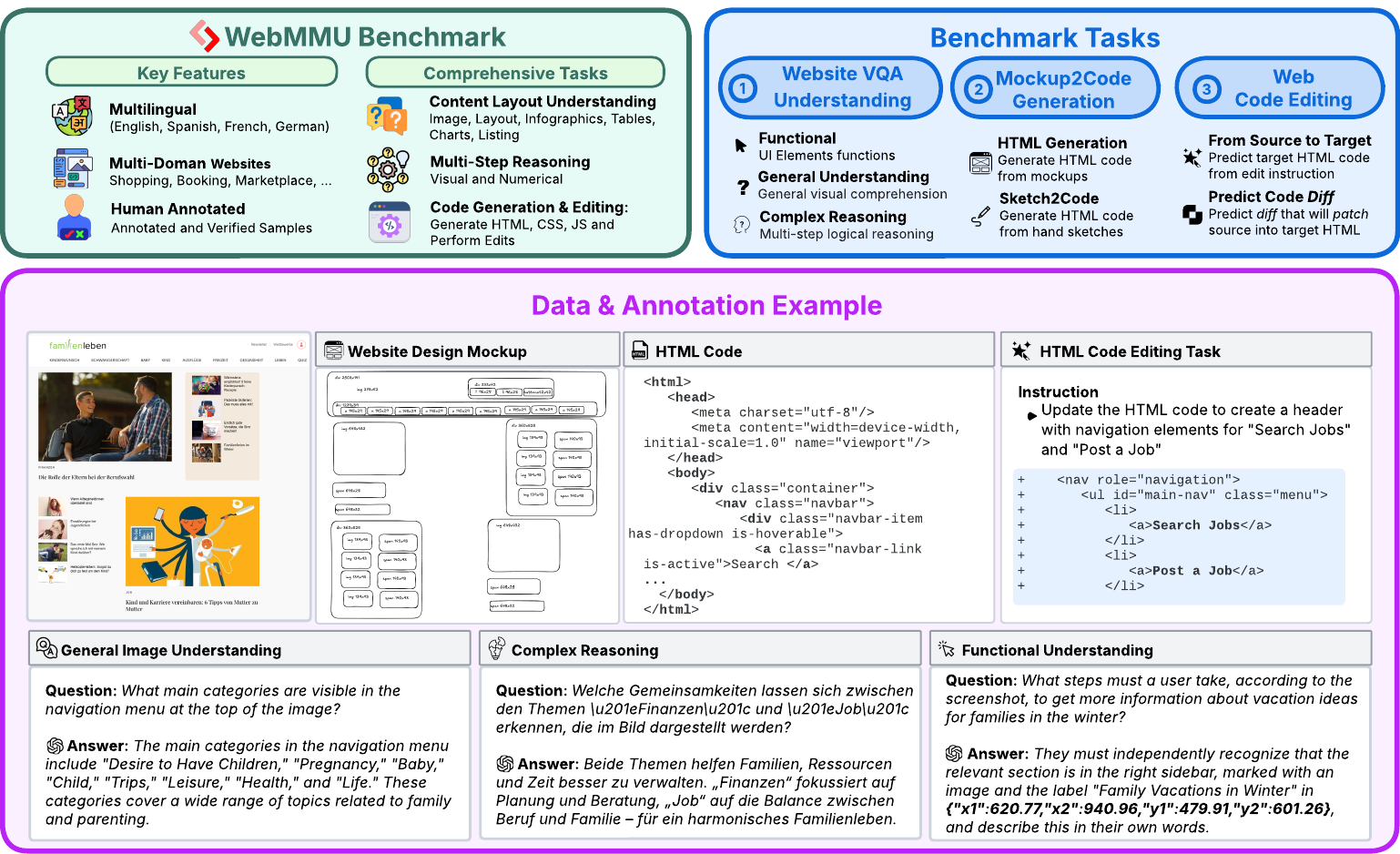}
    \caption{\textbf{\benchmark{} Benchmark Overview.} \benchmark{} evaluates models on diverse web-based tasks: \wqaNoStyle, \mockupNoStyle, and \codeeditNoStyle. Covering 20 domains and four languages, it challenges models to answer visual questions requiring multi-step reasoning and action grounding. It also assesses design-to-code generation from website layouts of varying complexity and evaluates code editing for automated web development.}
    \label{fig:\benchmark{}_overview}
\end{figure*}

%% file: sec/1_intro.tex
\section{Introduction}
\label{sec:intro}

The web is vital to daily life, enabling information access, shopping, and communication. Multimodal large language models (MLLMs)~\cite{Qwen2-VL,hurst2024gpt} that understand the \textit{Visual Web} can help users extract information, support tasks like budget-conscious shopping, and handle multiple languages~\cite{deng2024mind2web}. They also show promise in automating web design and development, including front-end layout creation, user interface (UI) editing, and code generation~\cite{anthropic2024claude}. Unlike tasks focused only on text or images~\cite{wang2024mmlu,yue2024mmmu}, visual web understanding requires combining UI structure, layouts, text, interactivity, and visuals.

Existing benchmarks target specific aspects of web tasks but remain fragmented and lack comprehensive coverage. Website VQA datasets like WebQA~\citep{chang2022webqa} and WebSRC~\citep{chen2021websrc} mainly focus on text retrieval, overlooking reasoning over UI structure, interactivity, and multilingual content. Recent web agent benchmarks evaluate online task completion \citep{koh2024visualwebarena,deng2024mind2web,he2024webvoyager}, showing promise for agentic AI but are limited to artificial websites or lack fine-grained categorization (e.g. grounding, understanding, multi-step reasoning).  In web development, design-to-code datasets such as Pix2Code~\citep{beltramelli2018pix2code} and Web2Code benchmarks~\citep{yun2024web2code}, as well as sketch-based datasets like Sketch2Code~\citep{li2024sketch2code}, cover a limited variety of UIs and often fail to capture real web variability due to automated creation. Furthermore, current benchmarks lack multilingual and cross-domain generalization, limiting applicability beyond English and specific domains. These gaps motivate a unified benchmark integrating multiple web tasks with multimodal, reasoning, and cross-lingual capabilities for effective evaluation of AI in web development and advanced web understanding.

To address these challenges, we introduce \textbf{\benchmark{}} (Figure~\ref{fig:\benchmark{}_overview}), a multimodal, \textbf{M}ultilingual, and \textbf{MU}lti-task benchmark for evaluating MLLMs on the Visual \textbf{Web} in four languages: English, Spanish, German, and French. \benchmark{} covers three core tasks: \textbf{Website VQA (\wqaNoStyle{})}, which tests functional understanding, visual comprehension, and multi-step reasoning via visual question-answering; \textbf{\mockupNoStyle{} Generation}, assessing design-to-code alignment for UI mockups and sketches, including both simple and complex nested layouts; and \textbf{\codeeditNoStyle{}}, evaluating precise, context-aware HTML/CSS/JavaScript code editing for feature additions, UI tweaks, and bug fixes. The benchmark spans 20 domains such as \textit{shopping}, \textit{booking}, \textit{sports}, and \textit{technology}, ensuring wide real-world relevance.

We benchmark state-of-the-art MLLMs across three core tasks, evaluating both open-source and closed-source models. Our results reveal significant challenges in action grounding and complex reasoning in the WebQA task, along with difficulties in structured layout understanding and accurate code generation for web development. While models (in particular, closed-source ones) exhibit strong general image understanding in \wqaNoStyle{}, they struggle with complex reasoning, with most scoring below 50\% and some as low as 2\% (e.g., Fuyu-8B in English), alongside notable multilingual performance drops (Figure~\ref{fig:vqa_failure_samples}).  In \codeeditNoStyle{}, top-performing models like \GeminiTwoFlash{} and \Claude{} outperform open-source counterparts, yet still struggle with maintaining logical structure and syntactic correctness, highlighting the need for more structure-aware code-editing techniques, particularly for complex modifications. Similarly, in \mockupNoStyle{}, models such as \oOne{} and Claude-3.5 achieve a high LLM-as-Judge score (4/5) on simple layouts but fail with nested element structures, revealing limitations in UI hierarchy comprehension. These findings emphasize the need for improved multimodal alignment, UI-aware modeling, and cross-lingual robustness to bridge the gap between vision-language models and real-world web interaction.

Our contributions are as follows:
\begin{itemize}[leftmargin=2em, itemsep=2pt, parsep=2pt]
\item \textbf{Comprehensive Multi-Task Benchmark}: A unified evaluation suite encompassing website VQA, web design-to-code generation, and code editing tasks.
\item \textbf{Diverse, Expert-Annotated Multilingual Data}: Fine-grained question-answer pairs, code edits, and UI design annotations across four languages, enabling comprehensive evaluation.
\item \textbf{Findings}: MLLMs face challenges in multi-step reasoning and grounding for WebQA, precise code editing, UI hierarchy understanding in web development, and multilingual generalization.
\end{itemize}

%% file: sec/2_related_work.tex
\section{Related Work}
\label{sec:related}

\paragraph{Web Understanding and Agentic MLLMs.}
Multimodal learning has become central to web UI understanding, integrating visual, textual, and structural modalities to support both web comprehension and agentic navigation. Early work, such as Screen2Words~\citep{wu2021screen}, parsed web screenshots into UI elements, later influencing MLLM pretraining\citep{lee2023pix2struct}. Recent advances leverage patching strategies\citep{baechler2024screenai}, grounding\citep{cheng2024seeclick}, text-structural alignment\citep{xu2024hierarchical, bai2021uibert}, and context-aware UI representations\citep{kil2024dual}. These innovations have expanded MLLM applications in web agents, enabling models to navigate and manipulate websites based on user instructions~\citep{gpt4vagent, yoran2024assistantbench, cheng2024seeclick}. However, existing benchmarks often rely on limited artificial websites\citep{deng2024mind2web, zhou2023webarena} or focus solely on English data\citep{he2024webvoyager, lu2024weblinx, zhang2024mmina, chen2024gui}, lacking diversity and real-world complexity. \benchmark{} addresses these gaps by incorporating real-world websites and multilingual queries, requiring models to perform complex reasoning and UI grounding, making it a more comprehensive evaluation framework for MLLM-driven web understanding and navigation.

\paragraph{Visual Question Answering for Web.} 
Progress in web-based VQA has been driven by benchmarks like WebSRC~\citep{chen2021websrc}, WebQA~\citep{chang2022webqa},
WebQuest~\citep{wang2024webquest}, VisualWebBench~\citep{liu2024visualwebbench}, and WebWalkerQA~\citep{wu2025webwalker} covering tasks such as captioning, webpage QA, and element grounding. Compared to traditional VQA on natural images~\cite{yue2024mmmu}, web-based VQA additionally requires understanding structured webpage layouts, the relationships between UI elements, and their functional roles within web environments. However, these benchmarks cover limited tasks, domains and languages. \benchmark{} addresses this gap by covering 20 domains in four languages and adding fine-grained categories like action grounding, multi-step reasoning, and general understanding for more thorough evaluation.

\paragraph{Automatic Web Design and Development.}  

Code generation and editing have been widely studied across programming languages, with benchmarks evaluating code generation~\citep{chen2021evaluating, jimenez2024swe, rodriguez2024starvectorgeneratingscalablevector, rodriguez2024bigdocs} and code editing based on natural language instructions~\citep{guo2024codeeditorbench, tian2024debugbench}. However, most previous studies focus on general-purpose programming, neglecting web design and development. To bridge this gap, \citet{gui2024vision2ui} and \citet{yun2024web2code} explore generating HTML/CSS from web screenshots. In contrast, \benchmark{} introduces \codeeditNoStyle{}, which involves multilingual tasks for modifying a website's visual and functional features based on user instructions, better reflecting real-world web development use cases. Additionally, \benchmark{} includes \mockupNoStyle{}; unlike prior work~\citep{jain2019sketch2code, barua2022sketch2fullstack} that relies on simplistic and artificial sketches drawn by researchers, our sketches are extracted from real-world websites, preserving complex element hierarchies through expert annotation.

%% file: sec/3_webmmu.tex
\section{\raisebox{-.5ex}{\includegraphics[width=0.5cm]{Figures/logo.png}}\,%
\benchmark{} Benchmark}
\label{sec:benchmark}

We introduce \benchmark{}, designed to evaluate MLLMs on real-world Visual Web tasks. In this section, we describe \benchmark{}’s data collection, annotation process, and present an overview of benchmark tasks.

\subsection{Data Collection and Annotation}

\paragraph{Website Selection and Data Capture.} To construct \benchmark{}, we curated a diverse set of webpage URLs from the FineWeb dataset~\citep{finewebpenedo2024the} and applied domain-specific heuristics to ensure coverage across 20 popular, content-rich, and feature-rich web domains (e.g., shopping, booking, technology). We selected webpages in four languages: English, German, French, and Spanish -- considering linguistic diversity, annotator availability, and budget constraints. To capture full browsing sessions on a single webpage, we generated collages combining multiple snapshots taken at different scroll depths and interaction states within the page. A viewport-specific snapshot was retained alongside relevant HTML and assets (e.g., CSS, JavaScript). Selection strictly adhered to web crawling policies (e.g., \texttt{robots.txt}).

\paragraph{Annotation Process.} Annotators were provided with webpage screenshots, corresponding HTML, and asset files and were tasked with three objectives: (1) generating open-ended and multiple-choice questions that capture real-world usage, including highlighting, clicking, and multi-step reasoning; (2) creating UI mockups of varying complexity and formats to support design-to-code workflows; and (3) formulating code edit requests that require programming expertise. A structured training phase ensured annotation consistency and quality. Further details on annotator guidelines are given in the Appendix \ref{app:annotator_prompts}.

\paragraph{Quality Control and Annotator Demographics.} A 100\% quality assurance framework was implemented in three stages: \textit{Trainer Review}, where experienced annotators performed initial annotations; \textit{Primary QA (QA1)}, where independent specialists verified accuracy, completeness, and adherence to guidelines; and \textit{Secondary QA (QA2)}, ensuring consistency with expert-level annotation criteria. The dataset was annotated by 127 professionals across North America, South America, Europe, Africa, and Asia, representing diverse linguistic and domain expertise. English annotators primarily came from Asia, German and French from Europe, and Spanish from Latin America. Annotators held qualifications ranging from bachelor's to advanced degrees for specialized tasks and were compensated above fair market wages, ensuring ethical labor practices and high-quality results.\looseness=-1

\subsection{Tasks Overview}


\subsubsection{Web Question Answering (\wqaNoStyle)}
The \wqaNoStyle{} task in \benchmark{} evaluates models' ability to extract, integrate, and ground structured UI elements, numerical data, and graphical components from web screenshots while reasoning over hierarchical layouts, predicting actions, and ensuring spatial grounding. It consists of three categories: \textbf{\agenticqa}, which focuses on web navigation and action execution without feedback from the environment, requiring models to understand UI elements like buttons, menus, and hyperlinks, identify elements (e.g., \textit{``Where can I find the coaching plans?''}), and execute actions (e.g., \textit{“How can I save this drill?”}) while handling \textit{spatial grounding} and distinguishing \textit{static vs. interactive elements} across multilingual UIs; many of these tasks also require \textit{coordinate-based reasoning} to localize UI components accurately. \textbf{\reasoningqa} involves multi-step inference, numerical calculations, and comparisons across UI components (e.g., \textit{``If a customer were to buy all the camera models mentioned on the bottom of this page in \"Expert Camera Reviews\" table, what would be the grand total?''}), requiring models to integrate text, numerical values, and layout structures from structured web content, where hierarchical reasoning is essential despite being constrained to single-frame snapshots; and \textbf{\generalqa}, which assesses a model’s ability to extract and synthesize structured and unstructured data from web screenshots, including OCR-extracted text, images, graphical elements, and UI components (e.g., \textit{``How many brand logos are in the Featured Brands section?''}), emphasizing \textit{semantic comprehension} beyond standard OCR-based extraction.
While existing web VQA benchmarks such as WebSRC~\cite{chen2021websrc}, WebQA~\cite{chang2022webqa}, and VisualWebBench~\cite{liu2024visualwebbench} provide useful tasks for information extraction or element grounding, they often focus on isolated subtasks, rely on static screenshots or HTML-only views, and primarily target English content. In contrast, WebMMU integrates action understanding, spatial grounding, and multi-step reasoning across four languages, making it a more comprehensive testbed for evaluating multilingual and agentic capabilities of MLLMs in real-world web environments. Unlike prior benchmarks, WebMMU includes questions with coordinate-based UI element localization, logical reasoning across DOM hierarchies, and multilingual UI semantics; all within a single benchmark and on realistic, domain-diverse web data.
\looseness=-1

\subsubsection{Mockup2Code}
The \mockupNoStyle{} task in \benchmark{} advances design-to-code by translating hand-drawn wireframes and high-fidelity digital mockups into structured code. Unlike text-based UI generation, it evaluates a model’s ability to interpret spatial hierarchies and UI structures from visual inputs. The dataset includes low-fidelity sketches and digitally created mockups, challenging models to generalize across abstraction levels in web design while tackling component recognition, spatial alignment, and structured code synthesis. Unlike prior design-to-code datasets, \benchmark{} incorporates real-world web layouts, ensuring models generate syntactically correct and semantically meaningful code aligned with modern web development practices.
Prior design-to-code efforts like Pix2Code~\cite{beltramelli2018pix2code}, Sketch2Code~\cite{jain2019sketch2code}, and WebSight~\cite{laurenccon2024unlocking} typically rely on synthetic or simplified UI inputs, often in English, and lack the layout depth and language variation present in real-world designs. WebMMU advances this space by incorporating expert-verified mockups drawn from real websites, supporting multilingual content, and emphasizing nested UI hierarchies. While prior datasets assess UI generation in isolation, our task evaluates models' ability to handle real, noisy design inputs and produce web-standard HTML/CSS outputs under spatial and structural constraints, providing a more realistic and multilingual benchmark for layout-to-code modeling.

\subsubsection{\codeeditNoStyle}
\codeeditNoStyle{} is a novel task, which evaluates a model’s ability to modify webpage code while preserving functional and structural integrity, given a screenshot, source code, and a user edit request. To perform well, models must complete three sub-tasks: (1) understand the provided inputs, including the webpage codebase, visual elements in the screenshot, and the requested modification; (2) identify the relevant code snippets that require modification; and (3) generate the appropriate HTML, CSS, or JavaScript edits to implement the requested change. These sub-tasks require an advanced understanding of webpage development and realistic code editing capabilities.
The modification requests span a broad range of visual and functional changes. Visual edits include adjusting font size and colors, repositioning elements, and adding headers or footers. Functional modifications involve adding interactive components such as buttons or forms and enhancing user experience with dynamic UI elements. The task is multilingual, aligning with the broader scope of \benchmark{}. Given the length of webpage source code, models are prompted to output only the necessary code differences rather than rewriting the entire codebase. This improves both practicality and efficiency, ensuring that the generated edits remain concise and targeted. More details on the prompt formulation are provided in Appendix~\ref{sec:code_edition_generation_prompt}. 
Existing benchmarks such as CodeEditorBench~\cite{guo2024codeeditorbench}, InstructCoder~\cite{li2023instructcoder}, and CanItEdit~\cite{cassano2023can} evaluate general-purpose code editing in languages like Python or JavaScript, often lacking visual context or domain specificity. In contrast, WebMMU focuses on precise modifications to HTML/CSS/JS code grounded in user-facing web interfaces. It uniquely incorporates multilingual edit instructions and aligns edits with visual page cues, simulating realistic web development workflows. Unlike benchmarks driven purely by unit test correctness, WebMMU emphasizes functional and structural preservation within existing codebases, thereby offering a more targeted and practical benchmark for UI-aware and multilingual web code editing.

\begin{table}[t]
  \centering
    \resizebox{\linewidth}{!}{
    \begin{tabular}{lccccc}
        \toprule
        & \textbf{En} & \textbf{Es} & \textbf{De} & \textbf{Fr} & \textbf{Total} \\
        \hline
        \textbf{Website Images}   & 834  & 418  & 416  & 391  & \textbf{2059}  \\
        \textbf{\wqaNoStyle}       & 1976 & 1521  & 1201  & 1404  & \textbf{6102} \\
        \textbf{\mockupNoStyle}    & 180  & 93   & 85   & 78   & \textbf{436}  \\
        \textbf{\codeeditNoStyle}  & 886  & 252   & 226 & 238 & \textbf{1602}  \\
        \bottomrule
    \end{tabular}
    
      }
    \caption{\textbf{Dataset Statistics.} Language-wise dataset breakdown across tasks. We report the number of web images per language. English (En), Spanish (Es), German (De) and French (Fr).}
  \label{tab:WebMMMU_language}
\end{table}

\subsection{Dataset Statistics}

\benchmark{}  covers four languages: English, Spanish, German, and French (see Table~\ref{tab:WebMMMU_language}). It contains $2059$ webpage images from domains like e-commerce, education, news, and finance. It includes $6102$ \wqaNoStyle{} samples, $436$ \mockupNoStyle{} instances, and $1602$ \codeeditNoStyle{} cases. Unlike previous datasets that focus on predefined UI layouts, \benchmark{} uses full-page snapshots, including dynamic content, nested structures, and multimodal dependencies. A small portion of images consist of multiple panels combined into a single image, capturing dense information and replicating browsing sessions.

%% file: sec/4_experiments.tex
\begin{table}[t]
    \centering
    \resizebox{\columnwidth}{!}{%
    \begin{tabular}{p{2cm} p{2.25cm} p{5cm}}
        \toprule
        \textbf{Task} & \textbf{Metric} & \textbf{Evaluation Details} \\
        \midrule
        \wqaNoStyle{} & LLM-as-Judge & Measures accuracy; 0 (incorrect) / 1 (correct). \\
        \mockupNoStyle{} & LLM-as-Judge & Assesses layout fidelity on a 1-5 scale (layout, spacing, grid). \\
        \multirow{2}{*}{Code Editing} & BLEU, TreeBLEU & Evaluates structural correctness by matching ground truth differences. \\
        & LLM-as-Judge & Scores functional accuracy on a 1-5 scale (functional correctness). \\
        \bottomrule
    \end{tabular}%
    }
    \caption{\textbf{Evaluation Metrics} used in WebMMU.}
    \label{tab:evaluation_metrics}
\end{table}

\begin{table*}[t]
  \centering
  \setlength{\tabcolsep}{6pt}  
  \renewcommand{\arraystretch}{1.1}  
  \resizebox{\textwidth}{!}{
  \begin{tabular}{lcccccccccccc}
    \toprule
    \multirow{2}{*}{Model} & \multicolumn{3}{c}{English} & \multicolumn{3}{c}{French} & \multicolumn{3}{c}{German} & \multicolumn{3}{c}{Spanish} \\
    \cmidrule(lr){2-4} \cmidrule(lr){5-7} \cmidrule(lr){8-10} \cmidrule(lr){11-13}
     & \textbf{\ReasoningIcon} & \textbf{\FunctionalIcon} & \textbf{\UnderstandingIcon} & \textbf{\ReasoningIcon} & \textbf{\FunctionalIcon} & \textbf{\UnderstandingIcon} & \textbf{\ReasoningIcon} & \textbf{\FunctionalIcon} & \textbf{\UnderstandingIcon} & \textbf{\ReasoningIcon} & \textbf{\FunctionalIcon} & \textbf{\UnderstandingIcon} \\
    
    \midrule
    \rowcolor{closed_models!50}Claude3.5 Sonnet & 48.1 & 2.9 & 47.1 & 48.2 & \bf 14.6 & 46.3 & 34.7 & \bf 16.0 & 41.0 & 62.4 & 16.1 & 49.3 \\
    \rowcolor{closed_models!50}\GeminiTwoFlash & 44.3 & 1.6 & 47.6 & 41.4 & 10.8 & 42.1 & 29.8 & 12.1 & 41.7 & 51.1 & 11.1 & 44.2 \\
    \rowcolor{closed_models!50}OpenAI-o1 & \textbf{68.2} & \bf 4.9 & \textbf{72.7} & \textbf{55.5} & 12.3 & \textbf{69.6} & \textbf{46.4} & 14.9 & \textbf{49.6} & \textbf{66.0} & 15.3 & \textbf{60.9} \\

     \midrule
     
    \rowcolor{open_models_below_8B!50}\PhiThreeFiveVI~\citep{abdin2024phi3} & 8.0 & 0.8 & 15.7 & 2.0 & 6.2 & 21.6 & 6.9 & 10.9 & 21.4 & 9.9 & 4.1 & 23.7 \\
    \rowcolor{open_models_below_8B!50}\UITARS~\cite{qin2025ui} & 17.3 & 0.5 & 42.3 & 9.4 & 4.1 & 38.0 & 9.6 & 5.6 & 38.5 & 11.9 & 2.3 & 33.2 \\
    \rowcolor{open_models_below_8B!50}\MolmoSevenB~\citep{molmo2024} & 16.5 & 0.0 & 40.6 & 6.4 & 4.2 & \bf 49.6 & 5.1 & 9.6 & 29.1 & 6.5 & 4.3 & 27.7 \\
    \rowcolor{open_models_below_8B!50}\QwenVLSevenB~\citep{Qwen2-VL} & 19.3 & 1.3 & 43.4 & 14.9 & 8.5 & 37.7 & 19.4 & 11.0 & 36.4 & 17.2 & 8.4 & 38.9 \\

    \rowcolor{open_models_below_8B!50}Llava-OV-7B~\cite{li2024llava}  & 18.9 & 2.3 & \bf 39.9 & 8.7 & 8.3 & 36.3 & 12.0 & 10.6 & 35.6 & 7.5 & 6.8 & 33.2 \\

    \rowcolor{open_models_below_8B!50} Qwen2.5VL 7B ~\citep{bai2025qwen2} & \bf  35.2 & \bf 30.8 & \bf 51.4 &  \bf 27.2 &  \bf 25.4 &  41.6 &  \bf 23.6 & \bf 22.7 & \bf 40.6 & \bf 32.6 & \bf 20.5 & \bf 44.8 \\
    
    \midrule

    \rowcolor{open_models_below_12B!50}\FuyuEightB~\citep{fuyu-8b} & 0.7 & 0.3 & 5.7 & 0.0 & 0.6 & 5.1 & 0.2 & 0.6 & 2.5 & 0.2 & 0.2 & 3.3 \\
    \rowcolor{open_models_below_12B!50}\InternVLTwoFiveEightB~\citep{chen2024internvl} & 23.3 & 1.4 & 36.2 &  13.1 & 6.0 & 30.5 & 15.7 & 10.9 & 34.8 & 14.2 & 6.2 & 33.8 \\
    
    \rowcolor{open_models_below_12B!50}\PixtralTwelveB~\citep{agrawal2024pixtral12b} & 24.5 & \bf 2.9 & 38.5 & \bf 19.2 & \bf 10.1 & \bf 53.5 & 12.8 & \bf 14.2 & 26.4 & \bf 21.6 & \bf 13.7 & 32.3 \\
    \rowcolor{open_models_below_12B!50}\LlamaThreeVision~\citep{dubey2024llama} & \bf 29.9 &  1.8 &  38.8 & 11.0 &  8.0 &  33.3 & \bf 17.9 &  12.2 & \bf 36.6 &  20.3 & 5.8 & \bf 34.7 \\

    \midrule

    \rowcolor{open_models_over_12B!50}Llava-OV-72B~\cite{li2024llava} & 31.7 & 1.9 & 40.2 & 16.4 & 8.2 & 39.0 & 23.6 & 11.7 & 42.4 & 23.3 & 7.4 & 39.2 \\
    \rowcolor{open_models_over_12B!50} Molmo 72B~\citep{molmo2024} & 25.8 & 1.1 & 40.1 & 18.5 & 7.1 & 40.4 & 23.3 & 12.0 & 36.1 & 26.0 & 7.5 & 38.6 \\
    
    \rowcolor{open_models_over_12B!50}\QwenVLSeventyTwoB~\citep{Qwen2-VL} & 33.7 & 1.0 & 44.2 & 28.1 & 11.6 & 42.0 & 28.2 & 11.8 & 42.9 & 37.5 & 10.3 & 43.6 \\
    \rowcolor{open_models_over_12B!50}\InternVLTwoFiveThirtyEightB~\citep{chen2024internvl} & 33.6 & 1.4 & 45.0 & 30.2 & 9.9 & 47.9 & 28.1 & 13.4 & 42.5 & 36.7 & 11.8 & 45.7 \\
    \rowcolor{open_models_over_12B!50} Qwen2.5VL 72B~\citep{bai2025qwen2} & \bf 43.4 & \textbf{37.1} & \bf 52.8 & \bf 36.8 & \textbf{36.5} & \bf 47.9 & \bf 31.2 & \textbf{28.7} & \bf 44.5 & \bf 44.9 & \textbf{41.4} & \bf 49.5 \\
    \bottomrule

    \end{tabular}
    }

    \caption{\textbf{\wqaNoStyle~ Performance.} Model accuracy (\%) by question type and language. \ReasoningIcon: \reasoningqa, \FunctionalIcon: \agenticqa, \UnderstandingIcon: \generalqa. Best models per size category are in \textbf{bold}. Model sizes: \colorbox{open_models_below_8B!50}{blue} (<8B params), \colorbox{open_models_below_12B!50}{orange} (8–12B), \colorbox{open_models_over_12B!50}{green} (>12B), \colorbox{closed_models!70}{gray} proprietary.}

  \label{tab:web_vqa_result}
\end{table*}

\begin{figure*}[h]
    \centering
    \includegraphics[width=1.0\linewidth]{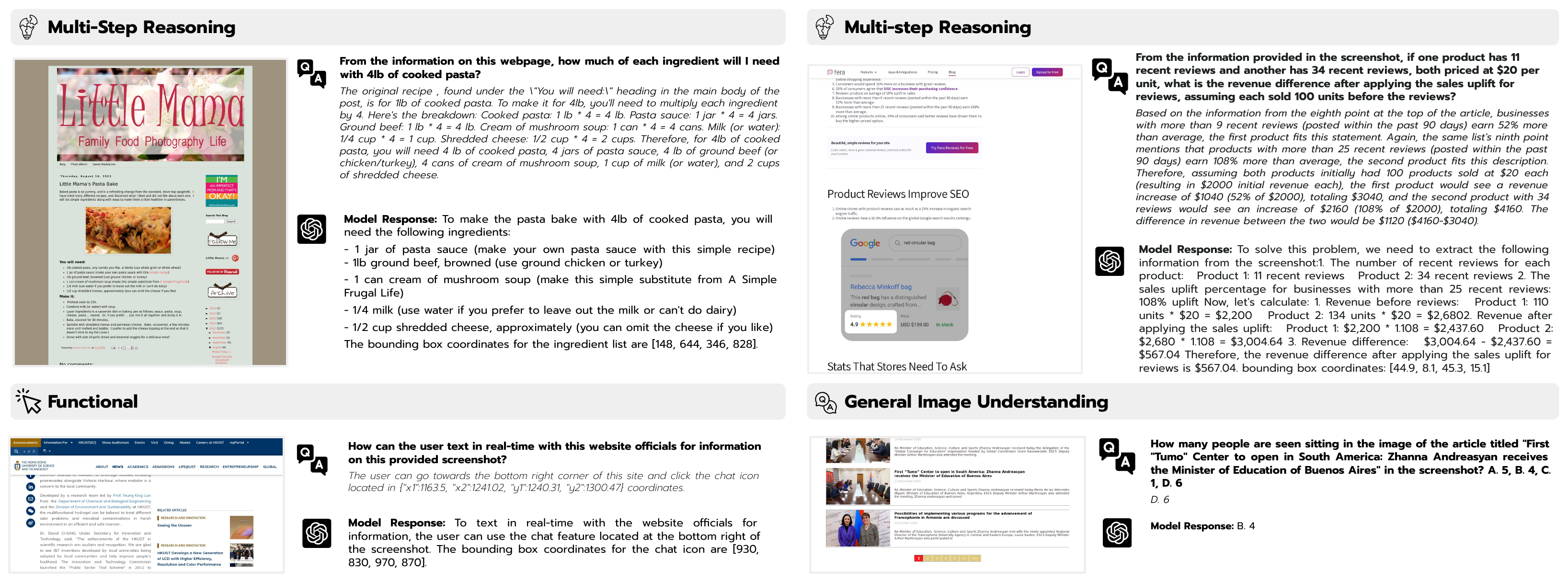}
    \caption{\textbf{Failure Cases in \wqaNoStyle} for one of the top-performing open-source model (InternVL-38B). Most prominent errors occur in grounding actions (e.g., identifying clickable link coordinates) and multi-step reasoning tasks, such as detailed step-by-step calculations.}
    
    \label{fig:vqa_failure_samples}
\end{figure*}

\section{Results}
\label{sec:results}

\begin{figure*}[h]
    \centering
    \includegraphics[width=\linewidth]{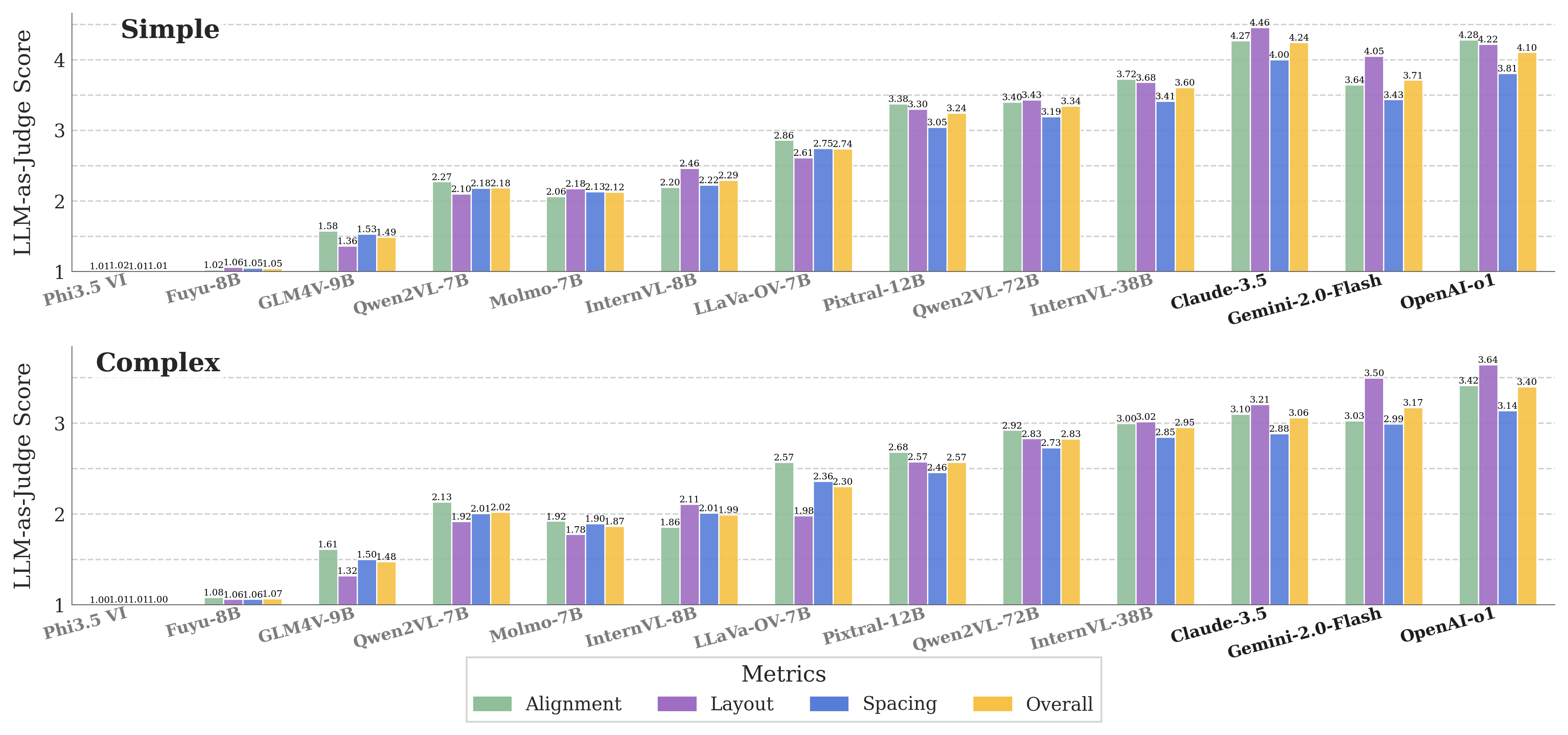}
    \caption{\textbf{Mockup2Code Performance.} LLM-as-Judge evaluation scores for simple and complex UI mockups across three key dimensions: \textit{alignment, layout, and spacing}, along with overall performance. Higher scores indicate better fidelity between the generated and reference web designs. Closed-source models outperform open-source alternatives, particularly in complex cases, yet challenges remain in high-fidelity code generation.}
    \label{fig:Mockup2Code_result}
\end{figure*}

\begin{figure*}[t]
    \centering
    \includegraphics[width=\linewidth]{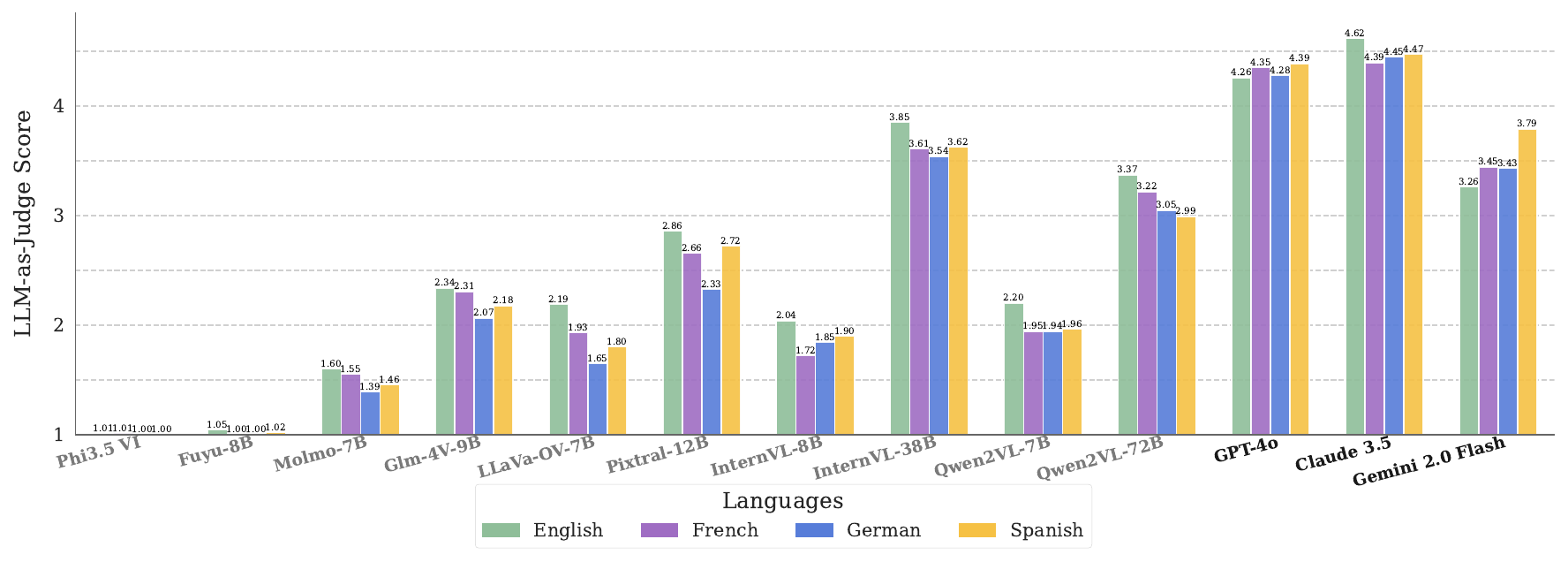}
    \caption{\textbf{Performance on Code Edits.} LLM-as-Judge metric, on a scale of 1-5, used to evaluate functional correctness of code edits. All models, including closed-source models, struggle with the \codeeditNoStyle{} task of \benchmark{}. Refer to Table~\ref{tab:codeedit} for full results, including BLEU and TreeBLEU scores, of all models.}
    \label{fig:codeedit-mlang}
\end{figure*}

\section{Evaluation}
We evaluate state-of-the-art MLLMs across both closed-source and open-source categories. Model inference for \wqaNoStyle{}, \mockupNoStyle{}, and \codeeditNoStyle{} follows standardized prompts (Appendix~\ref{appendix:prompt_formulation}). Evaluation combines LLM-as-Judge~\citep{zheng2023judging} scoring with established automatic metrics, as summarized in Table~\ref{tab:evaluation_metrics}. 

LLM-as-Judge is used to evaluate \wqaNoStyle{}, where model responses receive binary correctness scores (0 or 1) based on predefined criteria for semantic accuracy and reasoning completeness (Appendix~\ref{app:judge_prompts}). This structured approach ensures consistency and prevents arbitrary grading.  Inspired by automated evaluation in image synthesis \cite{ku2023viescore}, \mockupNoStyle{} uses LLM-as-Judge, assessing the alignment between input sketches and rendered outputs across three key dimensions: \textit{layout structure, spacing, and grid consistency} (Appendix~\ref{sec:mockup_evaluation_prompt}). Each aspect follows well-defined scoring guidelines, ensuring reproducible and fair assessments.  For \codeeditNoStyle{}, we evaluate both \textit{structural correctness} and \textit{functional accuracy}. The former is measured using BLEU~\citep{papineni2002bleu} and TreeBLEU~\citep{gui2024vision2ui}, ensuring syntactic validity and adherence to coding conventions. The latter relies on LLM-as-Judge, where functional equivalence between reference and predicted edits is rated on a 1-5 scale. To avoid arbitrary scoring, rating criteria explicitly define correctness levels based on functional preservation and intended user modifications. Since web functionalities can be implemented in multiple ways, the evaluation accounts for semantically valid alternatives, preventing undue penalization of syntactically different but functionally correct edits.  For all LLM-as-Judge evaluations, we use \GPTfouro{}, which has demonstrated strong alignment with human judgment and diverse scoring behavior~\citep{feizi2025pairbench}, ensuring robustness across tasks.

%% file: sec/5_results_and_experiments.tex
\subsection{WebQA Performance}  
Table~\ref{tab:web_vqa_result} presents model accuracy for three question types. Closed-source models, such as Gemini 2.0 Flash and Claude 3.5 Sonnet, outperform open-source alternatives across all tasks but still struggle with agentic action, particularly in predicting spatial coordinates for interactive elements. Among open-source models, larger architectures ($>$30B parameters) like Qwen2.5VL-72B and Internvl2.5-38B perform better in general image understanding and UI recognition, while smaller models ($<$8B) exhibit poor generalization across tasks.

Performance varies by question type. General image understanding is easiest, relying mainly on visual recognition. Complex reasoning is harder, with most models scoring below 50\% and some as low as 2\% (e.g., Fuyu-8b in English), showing difficulties in retrieving and reasoning over structured webpage content. \textbf{Agentic action is the hardest, with top models rarely surpassing 10\% accuracy}, as it requires precise spatial grounding, such as recognizing interactive elements (e.g., ``About Me'' in a menu) and predicting \textbf{approximate bounding boxes}. While many models detect interactive parts, they struggle with localization, resulting in low scores. \textbf{Error Analysis.} Figure~\ref{fig:vqa_failure_samples} reveals common failures: models often \textbf{miscalculate numbers or fail in multi-step reasoning}. In agentic action, inaccurate bounding boxes hurt performance. Multilingual generalization also remains weak despite resource-rich languages. These issues highlight the need for better spatial reasoning, numerical understanding, and cross-lingual adaptation to close the gap between vision-language models and real web interaction.

\subsection{Mockup2Code Generation}  
Figure~\ref{fig:Mockup2Code_result} evaluates the Mockup2Code task, reporting scores for each dimension and overall performance. Open MLLMs such as Phi3.5-VI, Fuyu-8B, and GLM4V-9B generally perform poorly across all metrics. Notably, Phi3.5-VI and Fuyu-8B score nearly 1 across all dimensions, indicating a complete failure on this task. Nevertheless, performance improves with model scale. For instance, Qwen2VL's score rises from 1.90 to 3.39 when scaling from 7B to 72B, while InternVL2.5 improves from 2.34 to 3.61 when scaling from 8B to 38B. Additionally, Pixtral-12B outperforms all 7B/8B models. Still, even the best open MLLMs struggle, especially with complex designs -- InternVL2.5-38B, the highest performer, scores only 2.98 out of 5. In contrast, proprietary models like Claude-3.5, Gemini-2.0-Flash, and \oOne{} perform significantly better, particularly on simple UI designs, where they achieve LLM-as-Judge scores above 4. \textbf{However, their performance declines in complex variants, with top scores reaching only 3.4 out of 5.} Across all evaluation dimensions, both proprietary and large-scale open MLLMs struggle most with spacing, which requires accurately setting element dimensions and margins based on sketch input. 
\begin{figure*}[t]
  \begin{subfigure}{\textwidth}
    \begin{tcolorbox}[
      colback=white,
      colframe=black!5,
      arc=0pt,
      boxrule=0.2pt,
      width=\textwidth,
      top=0pt,
      bottom=0pt,
      left=1pt,
      right=1pt
    ]
    \begin{tcolorbox}[
      colback=lightblue,
      colframe=lightblue,
      width=\textwidth,
      boxsep=0.5pt,
      top=0pt,
      bottom=0pt,
      left=2pt,
      right=2pt
    ]
      \textcolor{ecologygreen}{\textbf{\oOne}}
    \end{tcolorbox}
    
    \begin{minipage}{\textwidth}
      \begin{minipage}{0.53\textwidth}
        \begin{minipage}{0.32\textwidth}
          \centering
          \parbox[c][2.5cm][c]{\textwidth}{
            \includegraphics[height=2.2cm]{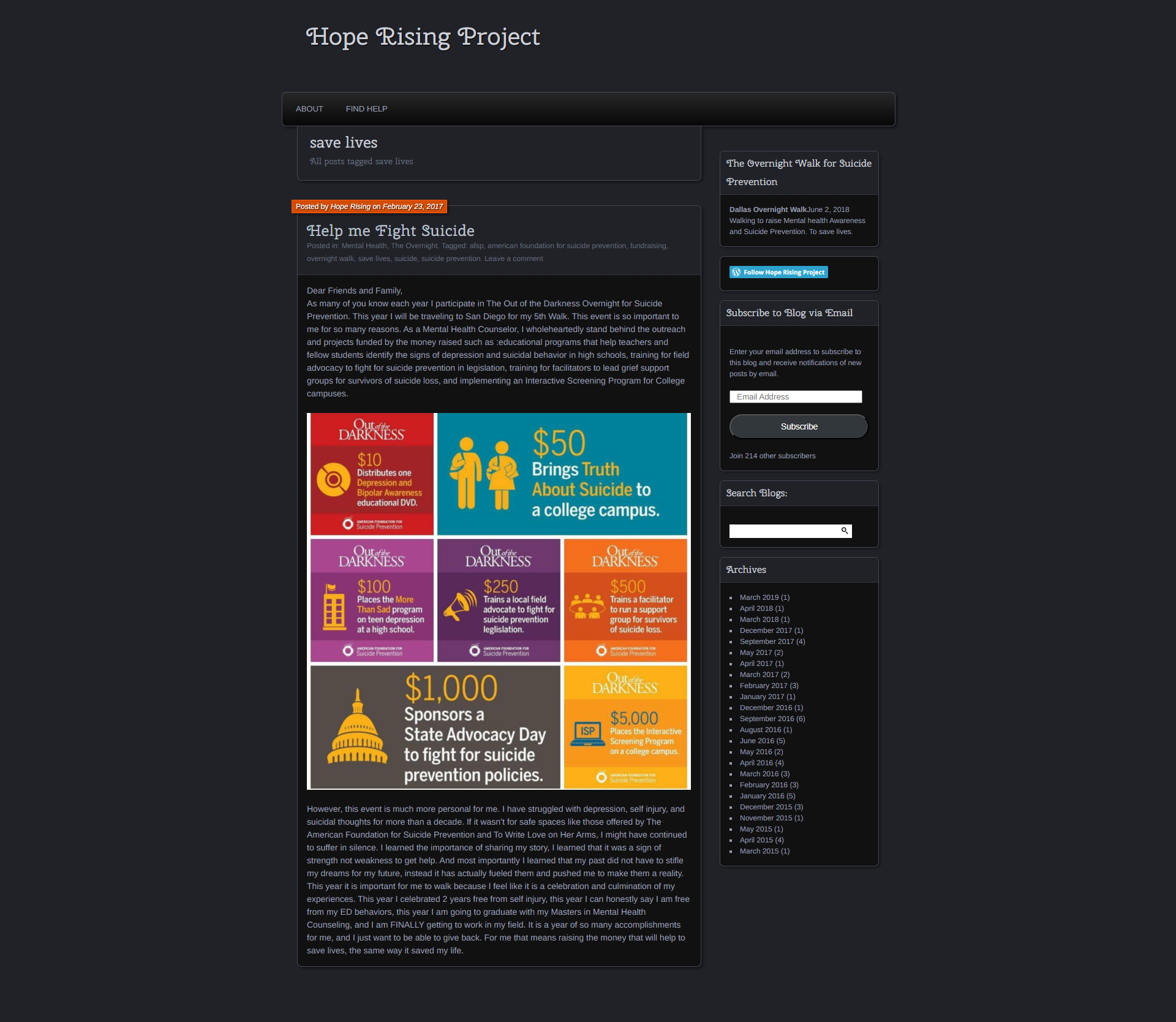}
          }
          \vspace{-0.2cm}\centering\textbf{\tiny Original Page}
        \end{minipage}%
        \hspace{0.004\textwidth}%
        \begin{minipage}{0.28\textwidth}
          \centering
          \parbox[c][2.5cm][c]{\textwidth}{
            \centering\includegraphics[height=2.4cm]{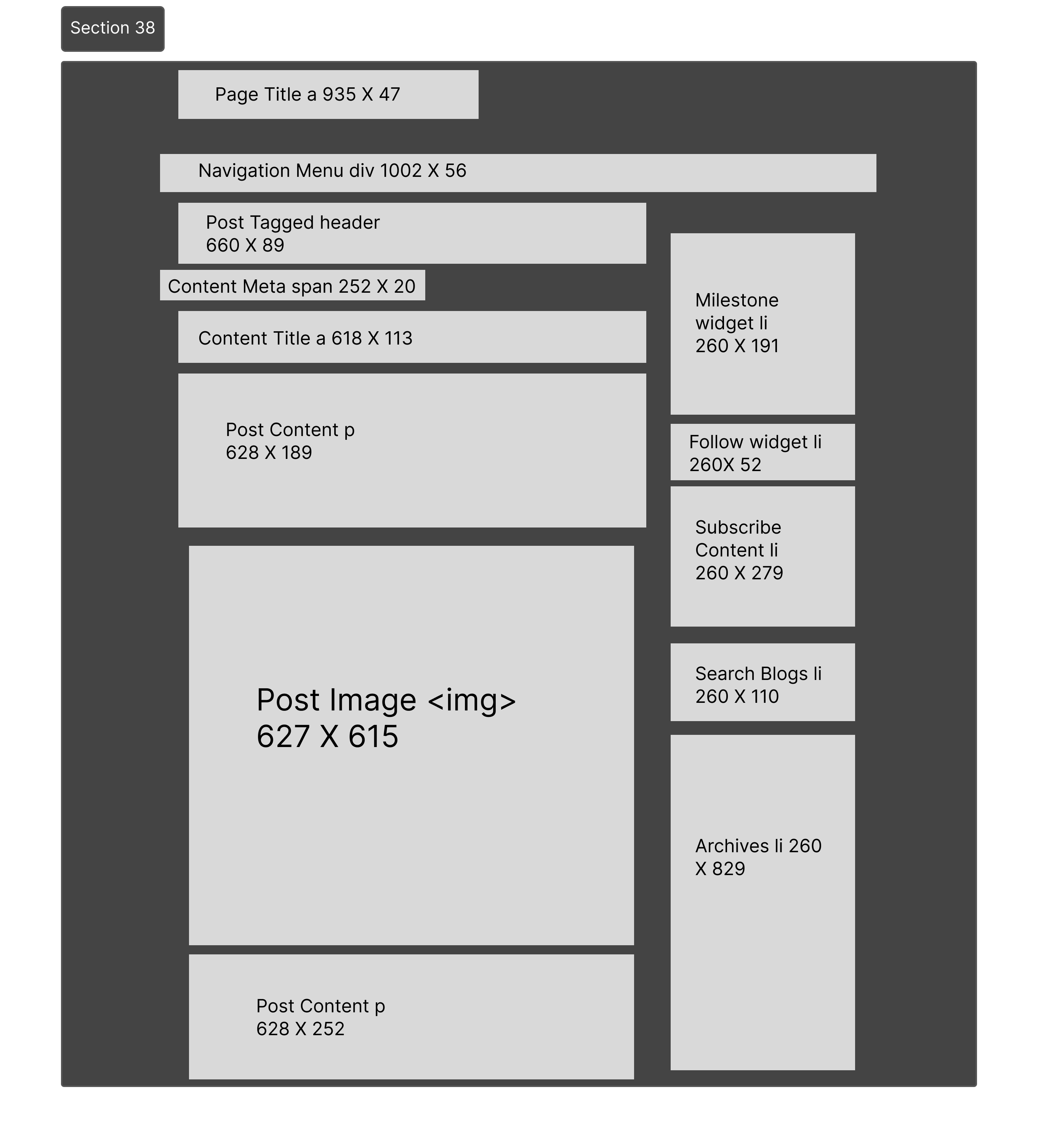}
          }
          \vspace{-0.2cm}\centering\textbf{\tiny Mockup Image}
        \end{minipage}%
        \hspace{0.002\textwidth}%
        \begin{minipage}{0.32\textwidth}
          \centering
          \parbox[c][2.5cm][c]{\textwidth}{
            \includegraphics[height=1.8cm]{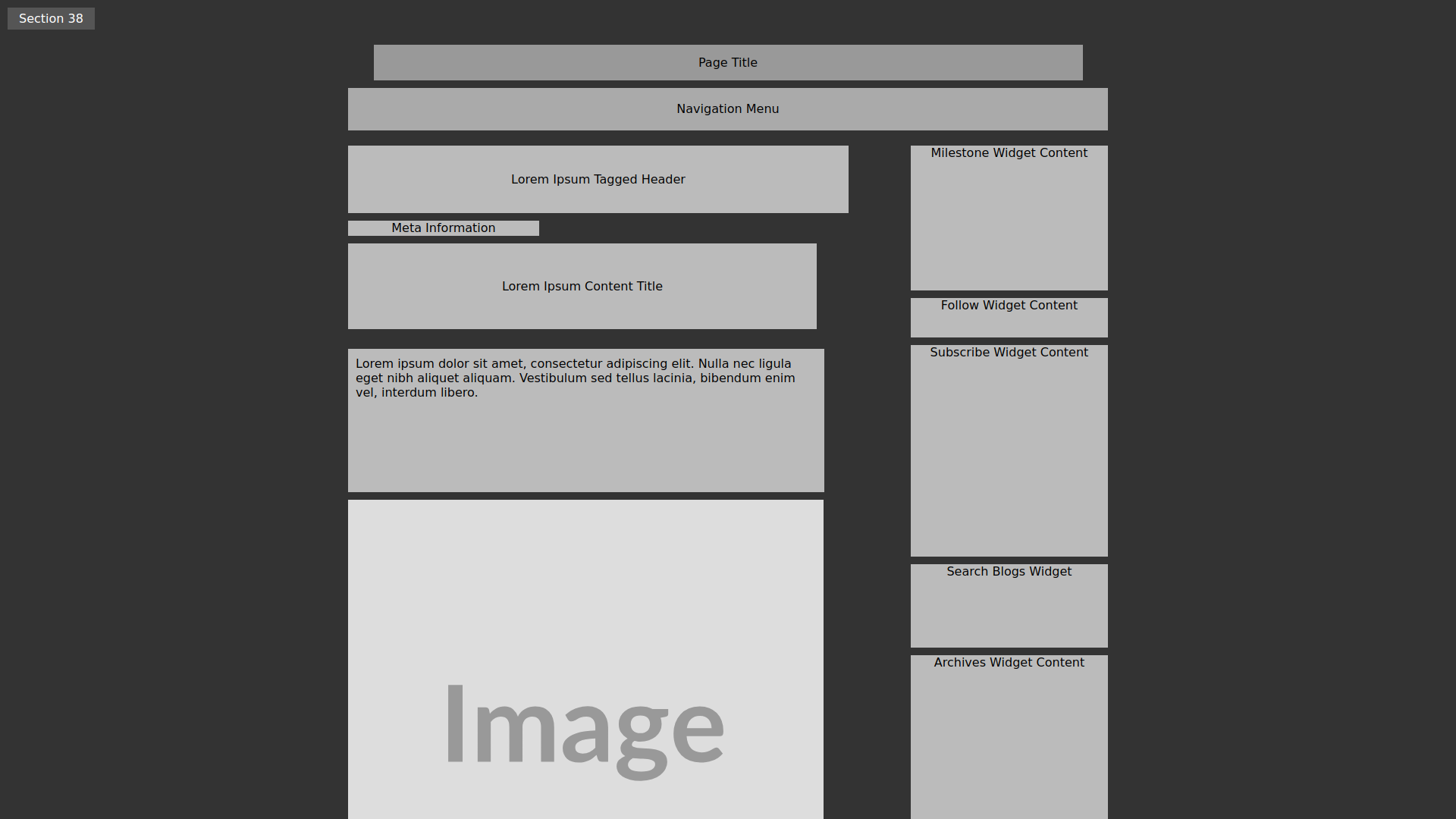}
          }
          \vspace{-0.2cm}\centering\textbf{\tiny Generation}
        \end{minipage}%
      \end{minipage}%
      \hspace{0.01\textwidth}%
      \begin{minipage}{0.44\textwidth}
        \begin{tcolorbox}[
          colback=codebg,
          colframe=gray!15,
          width=\textwidth,
          boxsep=0.5pt,
          top=0pt,
          bottom=0pt,
          left=1pt,
          right=1pt,
          title={\textcolor{black}{\textbf{\tiny LLM-as-Judge Evaluation}}},
          fonttitle=\tiny
        ]
          \begin{lstlisting}[basicstyle=\tiny\ttfamily, escapeinside={(*}{*)}]
(*\textbf{Alignment: 5.}*) The rendered design achieves perfect alignment with the sketch - text elements and sections are centered and positioned exactly as specified.
(*\textbf{Layout: 5.}*) The structure mirrors the input sketch flawlessly.
(*\textbf{Spacing: 5.}*) Spacing and proportions are consistent and balanced.
(*\textbf{Overall Score: 5}*)
          \end{lstlisting}
        \end{tcolorbox}
      \end{minipage}
    \end{minipage}
    \end{tcolorbox}
  \end{subfigure}
  
  \vspace{0.1cm}
  
  \begin{subfigure}{\textwidth}
    \begin{tcolorbox}[
      colback=white,
      colframe=black!5,
      arc=0pt,
      boxrule=0.2pt,
      width=\textwidth,
      top=0pt,
      bottom=0pt,
      left=1pt,
      right=1pt
    ]
    \begin{minipage}{\textwidth}
      \begin{minipage}{0.53\textwidth}
        \begin{minipage}{0.32\textwidth}
          \centering
          \parbox[c][2.5cm][c]{\textwidth}{
            \includegraphics[height=2.2cm]{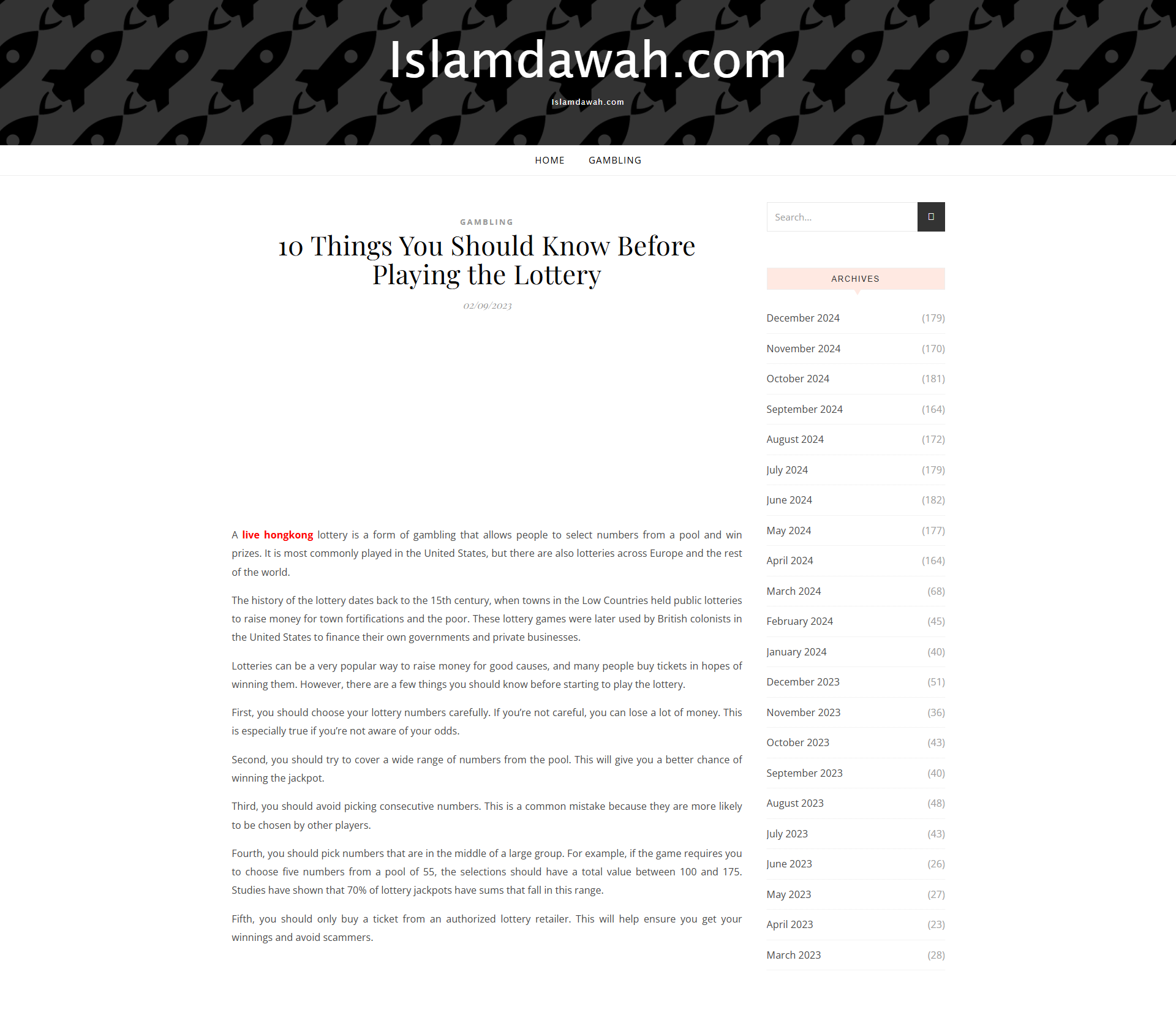}
          }
          \vspace{-0.2cm}\centering\textbf{\tiny Original Page}
        \end{minipage}%
        \hspace{0.004\textwidth}%
        \begin{minipage}{0.28\textwidth}
          \centering
          \parbox[c][2.5cm][c]{\textwidth}{
            \centering\includegraphics[height=2.4cm]{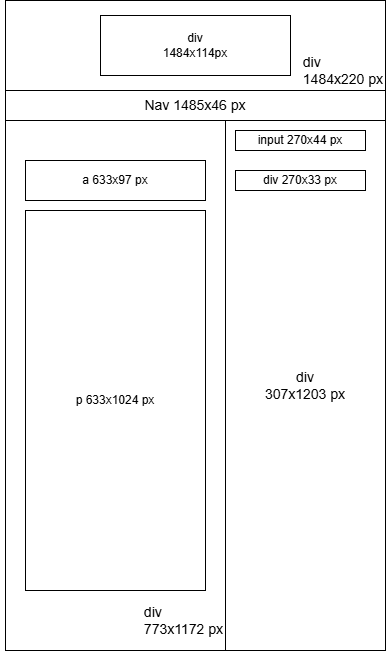}
          }
          \vspace{-0.2cm}\centering\textbf{\tiny Mockup Image}
        \end{minipage}%
        \hspace{0.002\textwidth}%
        \begin{minipage}{0.32\textwidth}
          \centering
          \parbox[c][2.5cm][c]{\textwidth}{
            \includegraphics[height=1.8cm]{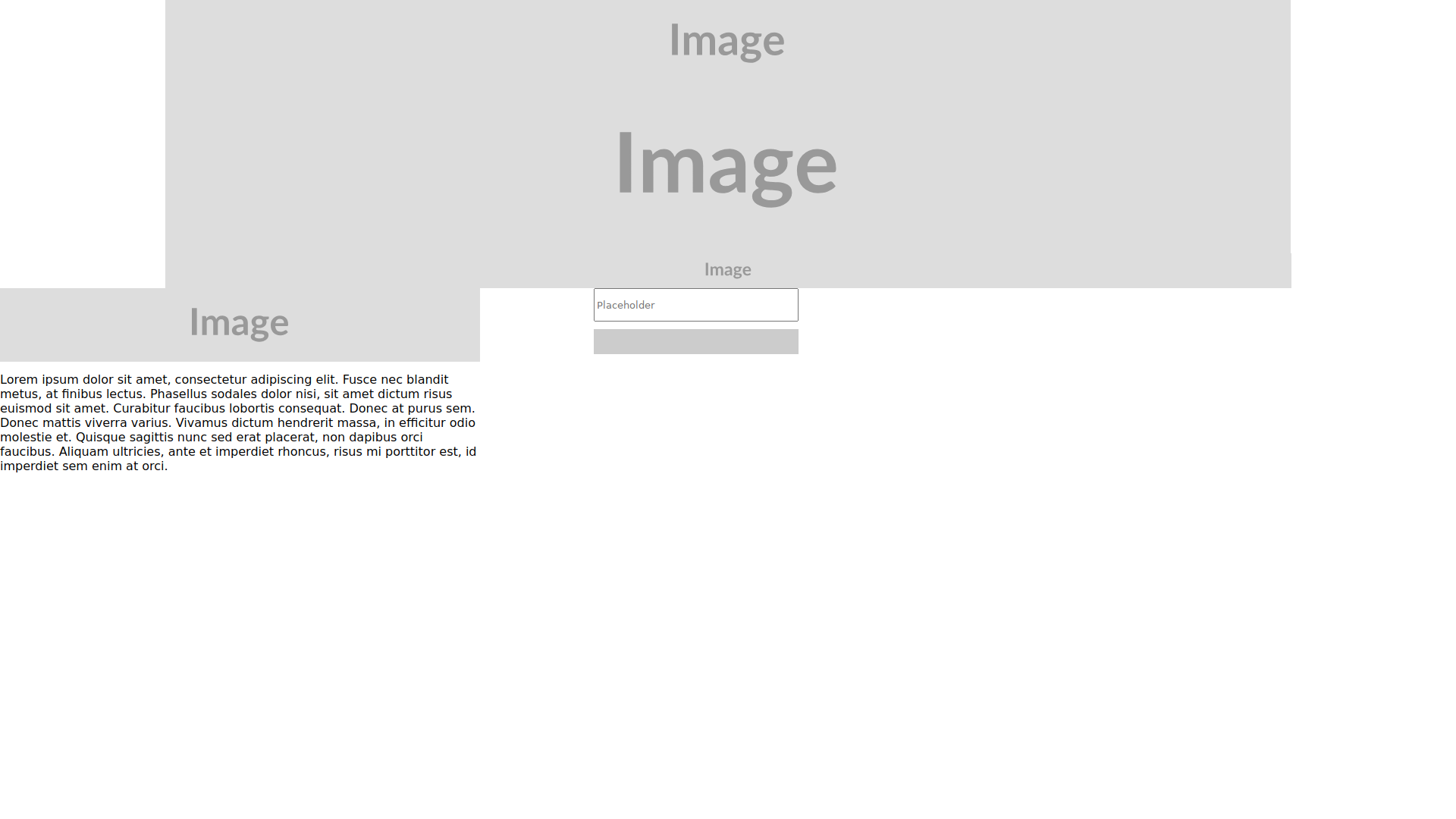}
          }
          \vspace{-0.2cm}\centering\textbf{\tiny Generation}
        \end{minipage}%
      \end{minipage}%
      \hspace{0.01\textwidth}%
      \begin{minipage}{0.44\textwidth}
        \begin{tcolorbox}[
          colback=codebg,
          colframe=gray!15,
          width=\textwidth,
          boxsep=0.5pt,
          top=0pt,
          bottom=0pt,
          left=1pt,
          right=1pt,
          title={\textcolor{black}{\textbf{\tiny LLM-as-Judge Evaluation}}},
          fonttitle=\tiny
        ]
          \begin{lstlisting}[basicstyle=\tiny\ttfamily, escapeinside={(*}{*)}]
(*\textbf{Alignment: 2.}*) Key elements (e.g., input box) misaligned, deviating from intended grid.
(*\textbf{Layout: 2.}*) Two-column structure poorly represented, essential sections missing/merged.
(*\textbf{Spacing: 2.}*) Uneven element distribution results in inconsistent spacing and imbalance.
(*\textbf{Overall Score: 2}*)
          \end{lstlisting}
        \end{tcolorbox}
      \end{minipage}
    \end{minipage}
    \end{tcolorbox}
  \end{subfigure}
  
  \caption{Success (top) and failure (bottom) cases for Mockup2Code Generation from \oOne.}
  \label{fig:mockup2code_success_case}
\end{figure*}

\textbf{Case Analysis.} Figure~\ref{fig:mockup2code_success_case} shows both success and failure cases of the top model \oOne{} on \mockupNoStyle{}. \oOne{} handles simple, flat layouts well, even with moderate element variety and count. However, it struggles with nested structures, often misaligning and failing to preserve element hierarchy and spacing, especially when \texttt{<p>} and \texttt{<a>} tags are nested within \texttt{<div>}. Similar issues occur in other samples and models, as seen in Figures~\ref{fig:mockup2code_failure_case_o1_and_internvl} and~\ref{fig:mockup2code_failure_case_internvl}.

\subsection{Code Editing Performance}  

Figure~\ref{fig:codeedit-mlang} shows \codeeditNoStyle{} results evaluated by LLM-as-Judge (metrics in Table~\ref{tab:codeedit}). Proprietary models achieve the highest functional accuracy, but only marginally outperform large open-source models, \textbf{indicating both struggle to preserve functional correctness alongside syntactic consistency}. Smaller models like Phi3.5-VI and \FuyuEightB{} perform worst, often failing to generate valid code (score $<$1.5). Performance improves with size; Qwen2VL-72B and InternVL2.5-38B rival closed-source models. Yet, even the strongest exhibit clear limitations producing structurally correct edits that fully preserve functionality. Multilingual performance is stable for top models but more variable for smaller ones, reflecting challenges in adapting edits across languages. Crucially, all models -- especially open-source -- fail to \textbf{automatically generate valid patch files} for seamless source integration. Despite access to full source files, none produced patch content directly usable without manual fixes, making human oversight essential and highlighting a core challenge in automating web code edits.

\subsection{Metric-Human Alignment}

We sampled 100 examples per task and enlisted PhD students and researchers as annotators to evaluate alignment between human judgments and the automatic metric (LLM judge). For \wqaNoStyle, humans agreed with the LLM judge in 89\% of cases. Most disagreements involved functional questions, where the judge required exact bounding boxes, but humans were more lenient -- accepting answers that correctly identified the clickable link location without a precise bounding box (e.g., the “about us” link in the navbar). We consider the judge’s stricter criteria correct since models were prompted to provide exact bounding boxes for such questions (see Appendix \ref{sec:wqa_generation_prompt}). For  \mockupNoStyle{} Spearman correlations were 0.39 (layout), 0.33 (spacing), and 0.46 (alignment), averaging 0.43 overall. Pearson correlations were slightly higher: 0.42 (layout), 0.41 (spacing), 0.48 (alignment), with an overall average of 0.50. These correlations, while moderate, reflect the task’s subjectivity and support the reliability of the automatic evaluation. For \codeeditNoStyle, expert annotators validated the LLM judge’s assessments with 91\% accuracy, demonstrating both the reliability of the evaluation and the validity of the associated judge's rationales assigned.\looseness=-1 

%% file: sec/6_conclusion.tex
\section{Conclusion}
\label{sec:conclusion}
\benchmark{} evaluates MLLMs on a real-world, challenging web question answering task and two code generation tasks: front-end design and code editing. Our tasks cover four languages and a wide variety of domains, sourced from human annotators. Our results show that Web VQA models struggle with interpreting complex UIs, reasoning, and multilingual generalization. Code editing models often generate syntactically correct but logically inconsistent code. UI generation models face a trade-off between precise element placement and preserving the original design’s meaning.  These challenges underscore the need for enhanced multimodal alignment, UI-aware architectures, and robust cross-lingual adaptation to develop future web agents capable of effectively performing a wide range of human tasks on the web.

\section*{Limitations}

While \benchmark{} provides a comprehensive evaluation of web-based AI reasoning and code generation, it has several limitations. First, it is restricted to single-screenshot web reasoning, capturing static snapshots rather than supporting interactive environments or multi-turn navigation. Although multi-step reasoning tasks are included, they rely solely on single-image (including multiple panels of a browsing session), limiting evaluation in dynamic web exploration. Second, linguistic coverage is constrained to four languages: English, French, German, and Spanish; due to annotator availability, which may limit generalization to underrepresented languages and regional web structures. Third, while \mockupNoStyle{} and \codeeditNoStyle{} cover core web technologies such as HTML, CSS, and JavaScript, modern frontend frameworks like React, Angular, and Vue.js are not explicitly evaluated. Finally, the automatic LLM judge metric, though reliable and fast, does not fully replicate human evaluation. Future work could explore improved automatic metrics or hybrid evaluation approaches to better capture nuanced human judgments. 

\section*{Ethical Considerations}

\benchmark{} is a benchmarking resource designed strictly for research purposes in multimodal and multilingual web understanding and generation. All tasks are created by human annotators using everyday web content and undergo thorough validation, so we do not anticipate misuse or harmful content. Compared to prior work, \benchmark{} expands evaluation across multiple languages, though coverage remains limited by annotator availability. To the best of our knowledge, the dataset contains no NSFW or harmful content. We commit to promptly removing any data upon valid requests once publicly released.

%% file: sec/X_suppl.tex
\section{Human Annotator Instruction}
\label{app:annotator_prompts}

\subsection{WebQA Annotations Guideline}
You will be provided with screenshots of websites. Your task is to create challenging questions that test deep understanding and reasoning about the image content. Each question should fall into one of the three categories described below, and be designed to encourage a detailed analysis of the screenshot.
\textbf{Important Note:}
If a screenshot lacks sufficient content or context for creating questions in any of the categories, mark the image as “Not enough content” and move to the next.

\paragraph{Agenctic Action}
Purpose: Focus on the interactive elements and navigation aspects of the website. These questions should prompt the viewer to interpret or locate specific functional elements, like buttons, menus, or links, and understand their purpose.
Example: “Where would a user click to access their saved items?”
Guidelines: Create questions that require the viewer to understand how different elements work or what actions they might trigger. Avoid overly simple questions that don’t involve interaction or navigation. Do provide the bounding box location or hint on how to navigate.
\paragraph{Multi-step Reasoning}
Purpose: These questions should require multi-step thinking, involving the analysis of multiple parts of the image, comparisons, or drawing inferences from the content.
Example: “How does the timing of updates in different news sources on this page provide insights into the event’s coverage?”
Guidelines: Formulate questions that connect elements across the image or require interpretation of trends, relationships, or content hierarchy. These should not be answerable from a single part of the image. If answerable, then should be difficult e.g. solving a math question (see example) or asking what will happen if the cart is doubled (see example).
\paragraph{General Visual Comprehension}
Purpose: Assess the viewer’s ability to identify and comprehend basic information displayed in the image, such as titles, labels, or the overall structure.
Example: “What is the main title or header of this page?”
Guidelines: Keep these questions straightforward, focusing on textual or visual elements that convey the primary purpose or information displayed. Aim for questions that require attention to specific details rather than general impressions. Highlight the region of answer with bounding box if needed (upto your choice).

\subsection{Performing Code Editing on Websites}
\paragraph{Understanding the Scope of Edits}
Before starting, identify the specific task or issue with clarity and precision. Ensure you fully understand the requested visual or functional changes before proceeding. Examples of tasks by difficulty are outlined below

\paragraph{Basic Changes}
\begin{itemize}
    \item Change the button color from blue to green.
    \item Fix a typo in the homepage headline.
    \item Remove the underlined style from all hyperlinks.
    \item Add a border to images in the gallery section.
\end{itemize}

\paragraph{Intermediate Enhancements}
\begin{itemize}
    \item Replace the navigation bar font with 'Roboto' and ensure it matches the design mockup.
    \item Add a hover effect to all buttons, changing their background to light gray.
    \item Update the footer links to open in a new tab and add appropriate ARIA labels for accessibility.
    \item Create a consistent color scheme for all headings on the page.

\end{itemize}

\paragraph{Advanced Functional or Design Tasks}
\begin{itemize}
    \item  Add a new section to the homepage to showcase recent blog posts, styled to match the website theme.
    \item Refactor the JavaScript for the carousel to improve performance and fix the sliding bug.
    \item Optimize the CSS for faster page load times by combining redundant rules and removing unused classes.
    \item Implement a lightbox feature for viewing images in the gallery.
    \item Create a visually engaging header with a full-width background image and overlay text for the homepage.
    \item Design a custom 404 error page with an animated illustration and a link back to the homepage.
    \item Develop a visually interactive pricing table with hover effects to highlight selected options.
    \item Redesign the "About Us" section using a card layout for team member profiles, including images and bios.
    \item Update the contact form with a modern design, including floating labels and inline validation.
    \item Animate the scrolling experience for anchor links to smoothly transition between sections of the page.
\end{itemize}

\textbf{Key Principles:} a) Focus on Instructions. b) Only address the requested tasks and avoid unrelated changes unless explicitly instructed. c) Document Changes Clearly and d) For every modification, provide a clear record that includes:
\begin{itemize}
    \item What was changed?
    \item Why was it changed?
    \item The location of the change (e.g., file name and line numbers, or element location in the inline HTML).
\end{itemize}

\subsection{Performing Sketch Task}
The distinction between simpler and more complex sketches typically depends on the number of components and the level of detail in the specifications. Simpler sketches usually have fewer elements (e.g., basic shapes, minimal labels), while complex sketches include multiple, interrelated components and detailed instructions (e.g., specifying dimensions, class names like ‘div nav,’ or explicit layout details). To differentiate, consider:
\textbf{Simple}: Basic wireframes or mockups with minimal annotations (e.g., a rectangle representing a button).
\textbf{Complex}: Detailed designs specifying attributes (e.g., ‘button 200px wide, div with class=“nav”’) or involving hierarchical or nested components.

\section{Task Samples}
Tables~\ref{tab:vqa_task_samples}, \ref{tab:sketch2html_edit_task_samples}, and \ref{tab:code_edit_task_samples} present representative examples from the WebMMU dataset, covering \wqaNoStyle{}, \mockupNoStyle{}, and \codeeditNoStyle{} tasks. The \wqaNoStyle{} task (Table~\ref{tab:vqa_task_samples}) evaluates a model’s ability to interact with webpage elements, recognize visual content, and perform complex reasoning based on structured UI components. The \mockupNoStyle{} task (Table~\ref{tab:sketch2html_edit_task_samples}) illustrates how webpage screenshots are converted into structured HTML representations, distinguishing between basic layout sketches and detailed UI component mappings. The \codeeditNoStyle{} task (Table~\ref{tab:code_edit_task_samples}) demonstrates automated HTML modifications, providing before-and-after visual transformations based on functional and design-driven prompts. These task samples comprehensively showcase the challenges in webpage understanding, layout structuring, and automated UI refinement within the WebMMU benchmark.

\begin{table*}[h]
    \centering
    \resizebox{\textwidth}{!}{ 
    \begin{tabular}{>{\centering\arraybackslash}m{4.5cm} >{\arraybackslash}m{6cm} >{\arraybackslash}m{6cm} >{\arraybackslash}m{6cm}}
        \toprule
         & \FunctionalIcon & \UnderstandingIcon & \ReasoningIcon \\
        \midrule

        \includegraphics[width=4.5cm, keepaspectratio]{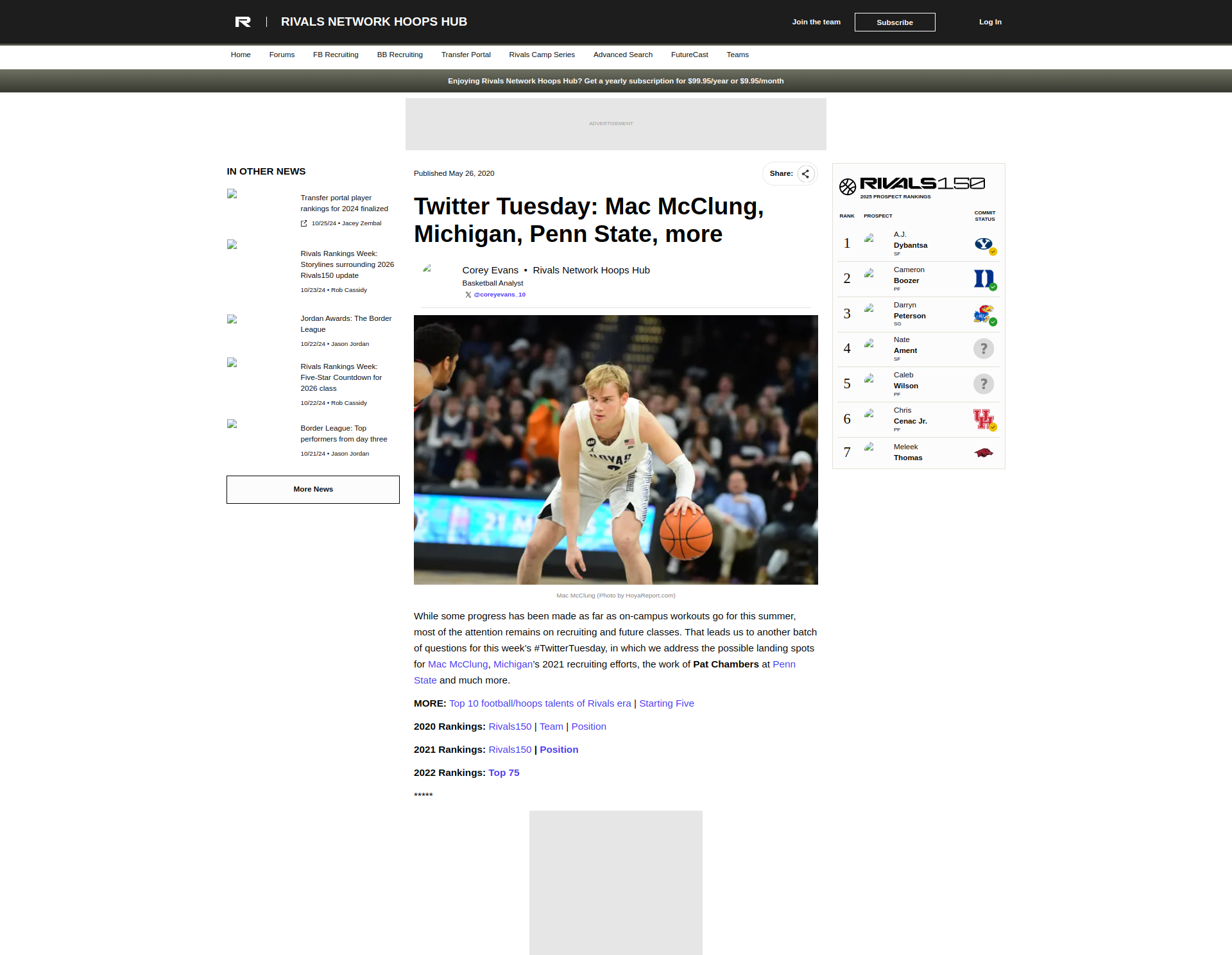} 
        & 
        How can I find more information about the player A.J. Dybantsa?  
        \newline Answer: Click on the "RIVALS150 ranking" at the lower left and select "A.J. Dybantsa" at (x1:230.34, x2:297.32, y1:1049.92, y2:1083.07).
        & 
        How many players are visible in the 4th image on the left side?  
        \newline A) 2 \quad B) 1 \quad C) 4 \quad D) 3  
        \newline Answer: D) 3
        &  
        Which user pays less when subscribing annually vs. monthly?  
        \newline Answer:  
        - Yearly: \$99.95  
        - Monthly: \$9.95 × 12 = \$119.40  
        - Savings: \$19.45 \\
        \midrule

        \includegraphics[width=4.5cm, keepaspectratio]{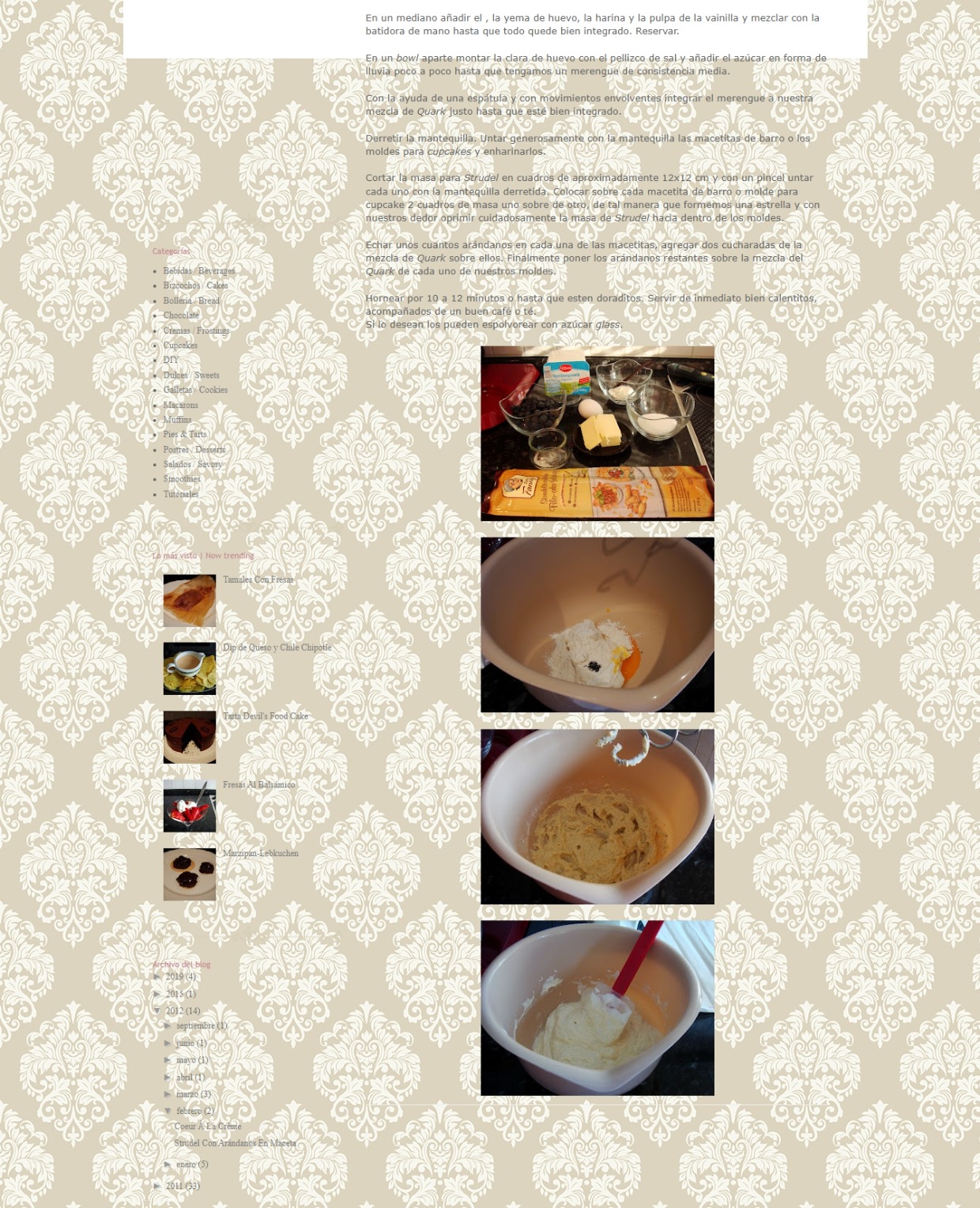} 
        &  
        ¿En qué área puede el usuario hacer clic para ver publicaciones anteriores?  
        \newline A) Categorías \quad B) Lo más visto \quad C) Archivo del blog  
        \newline Respuesta: C) Archivo del blog.
        &  
        ¿A qué tipo de público está dirigido este blog de recetas?  
        \newline Respuesta: Personas interesadas en repostería y cocina casera.
        &  
        ¿Cuál sería el efecto en la navegación si la sección "Lo más visto recientemente" estuviera al inicio?  
        \newline Respuesta: Aumentaría la accesibilidad de los artículos populares. \\
        \midrule

        \includegraphics[width=4.5cm, keepaspectratio]{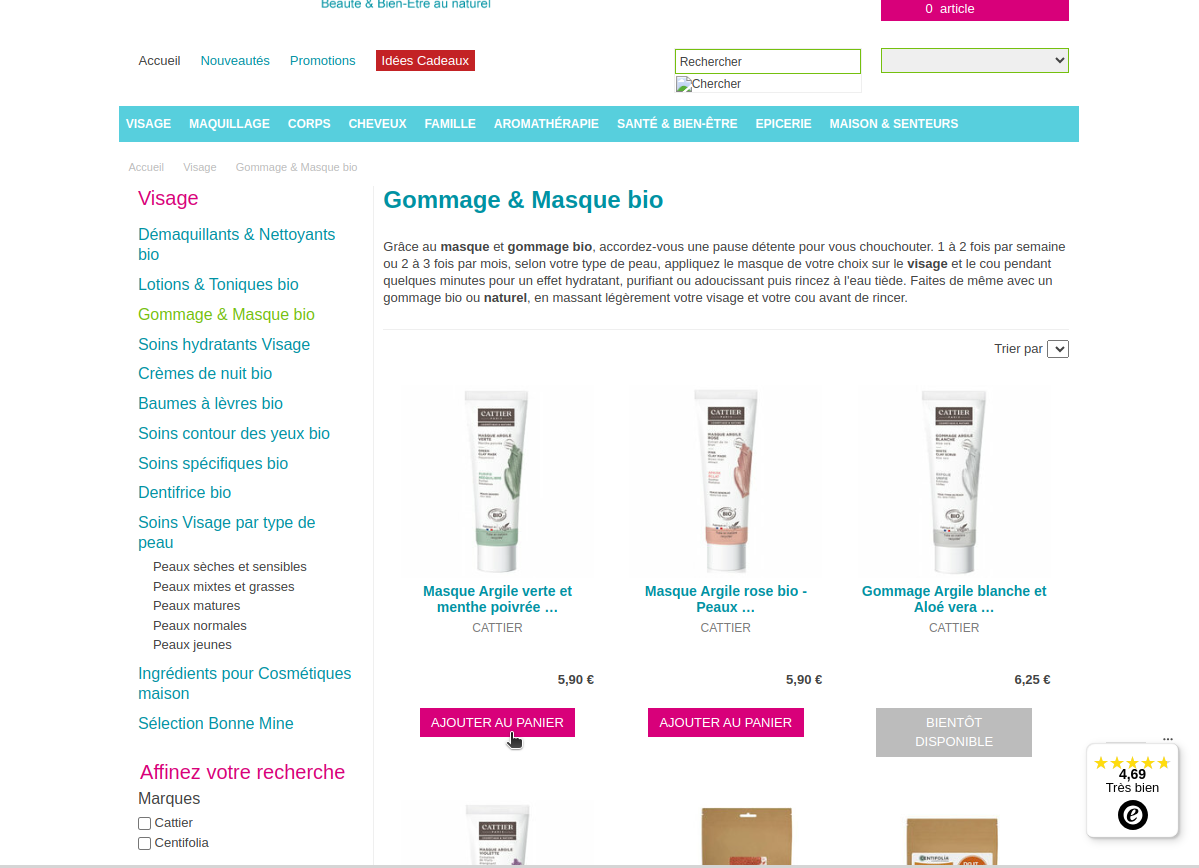} 
        &  
        Je cherche un soin pour le visage. Où dois-je aller ?  
        \newline Réponse: Dans la barre latérale sous "Soins visage par type de peau".
        &  
        Quel est le prix total des articles si l’on exclut celui avec 5 étoiles ?  
        \newline A) 58,70 € \quad B) 62,85 € \quad C) 45,50 € \quad D) 51,90 €  
        \newline Réponse: B) 62,85 €.
        &  
        Avec un budget de 15 €, quels produits puis-je acheter ?  
        \newline Réponse:  
        - Masque Argile verte et menthe poivrée bio (5,90 €)  
        - Masque Argile rose bio - Peaux sensibles (5,90 €) \\
        \midrule

        \includegraphics[width=4.5cm, keepaspectratio]{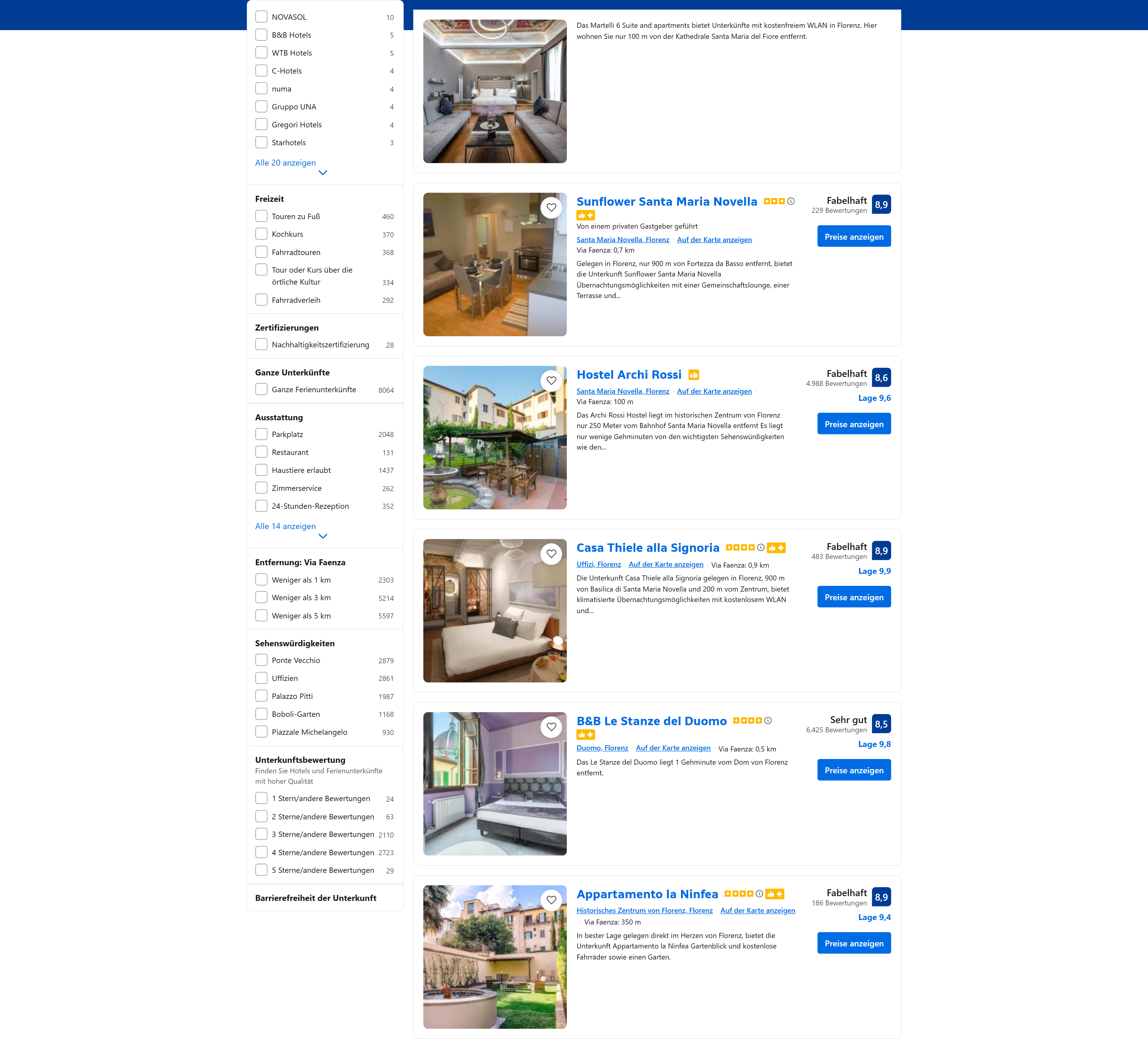} 
        &  
        Comment filtrer les hôtels qui acceptent les animaux ?  
        \newline Réponse: Dans la section "Ausstattung", cliquez sur "Haustiere erlaubt".
        &  
        Combien de chaises sont visibles dans l’image de "Sunflower Santa Maria Novella" ?  
        \newline Réponse: 4 chaises.
        &  
        Quelles sont les meilleures options d'hébergement près de Via Faenza ?  
        \newline Réponse: "Sunflower Santa Maria Novella" avec une note de 8.9. \\

        \bottomrule
    \end{tabular}
    } 
    \caption{\textbf{WebMMU VQA Task Samples.} This table presents diverse Visual Question Answering (VQA) task samples from the WebMMU dataset, categorized into three types: (1) Functional (interaction with webpage elements), (2) General Understanding (visual recognition within webpage images), and (3) Complex Reasoning (logical inference and numerical computation). Each row showcases an input webpage image alongside representative questions and answers.}
    \label{tab:vqa_task_samples}
\end{table*}

\begin{table*}[h]
    \centering
    \renewcommand{\arraystretch}{1.5} 
    \setlength{\tabcolsep}{6pt} 
    \begin{tabular}{>{\centering\arraybackslash}m{4.8cm} >{\centering\arraybackslash}m{4.8cm} >{\centering\arraybackslash}m{4.8cm}}
        \toprule
        \textbf{Input Image} & \textbf{Basic Layout Sketch} & \textbf{Detailed UI Representation} \\
        \midrule

        \begin{subfigure}[t]{4.8cm}
            \centering
            \includegraphics[width=\linewidth, keepaspectratio]{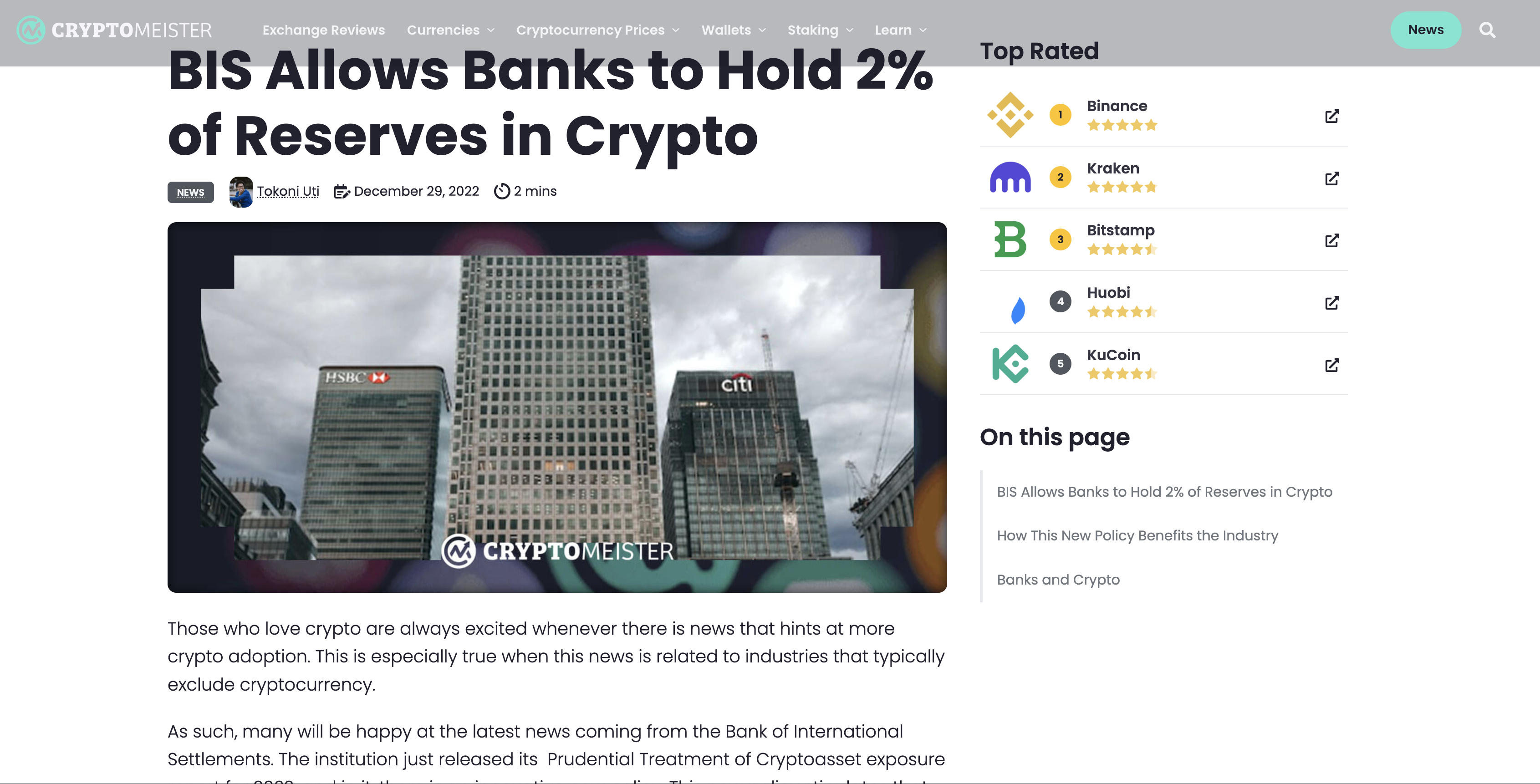}
        \end{subfigure} &
        \begin{subfigure}[t]{4.8cm}
            \centering
            \includegraphics[width=\linewidth, keepaspectratio]{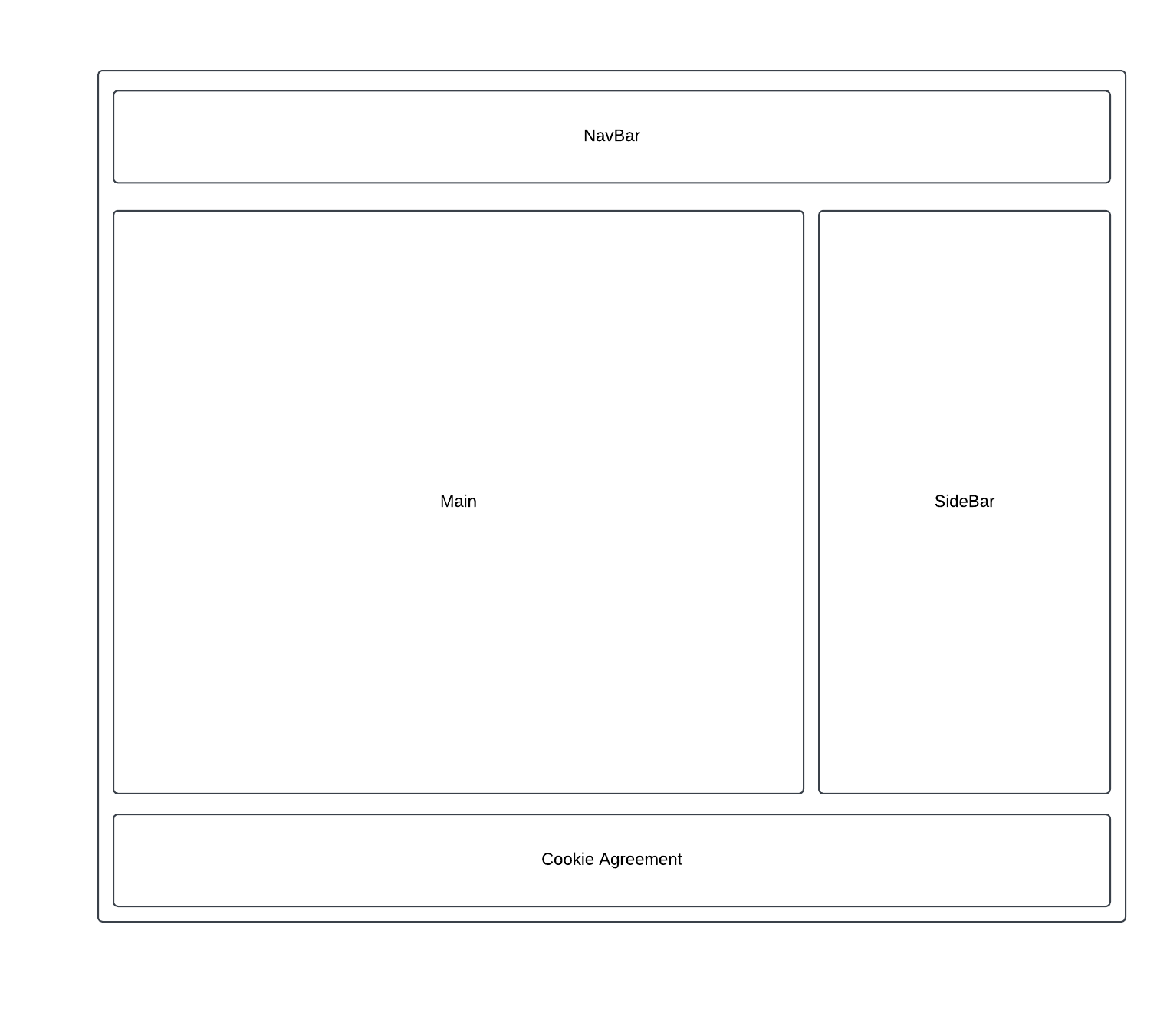}
        \end{subfigure} &
        \begin{subfigure}[t]{4.8cm}
            \centering
            \includegraphics[width=\linewidth, keepaspectratio]{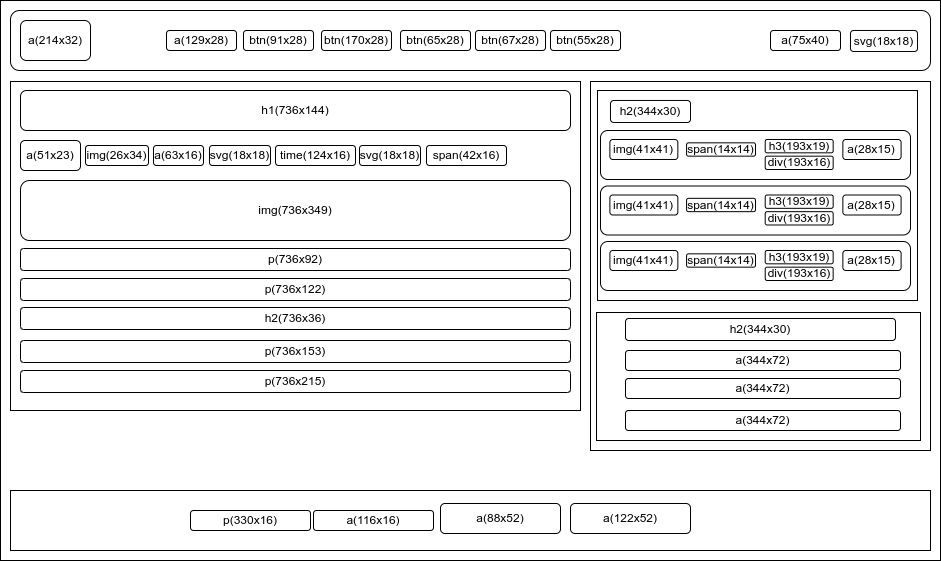}
        \end{subfigure} \\

        \midrule

        \begin{subfigure}[t]{4.8cm}
            \centering
            \includegraphics[width=\linewidth, keepaspectratio]{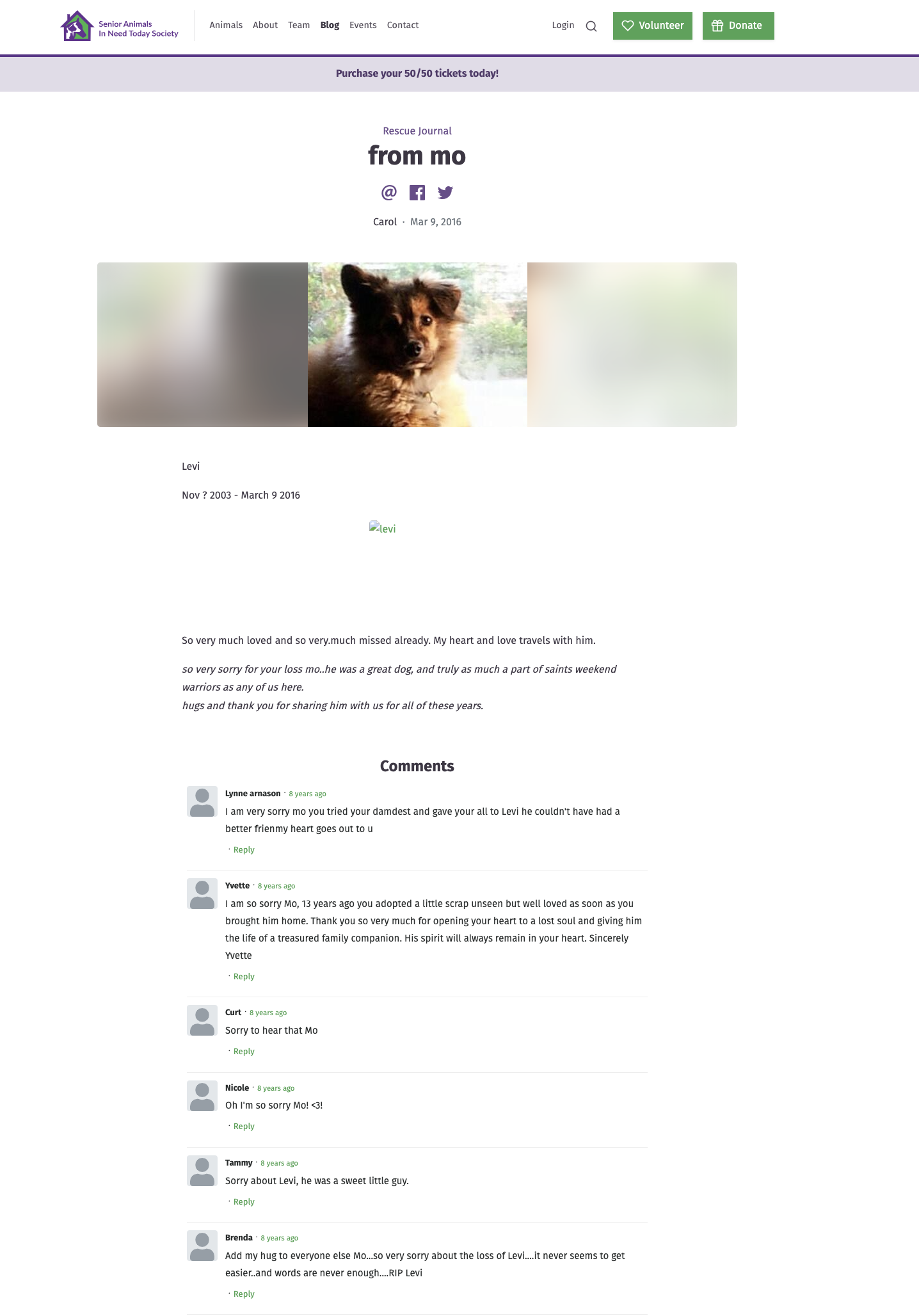}
        \end{subfigure} &
        \begin{subfigure}[t]{4.8cm}
            \centering
            \includegraphics[width=\linewidth, keepaspectratio]{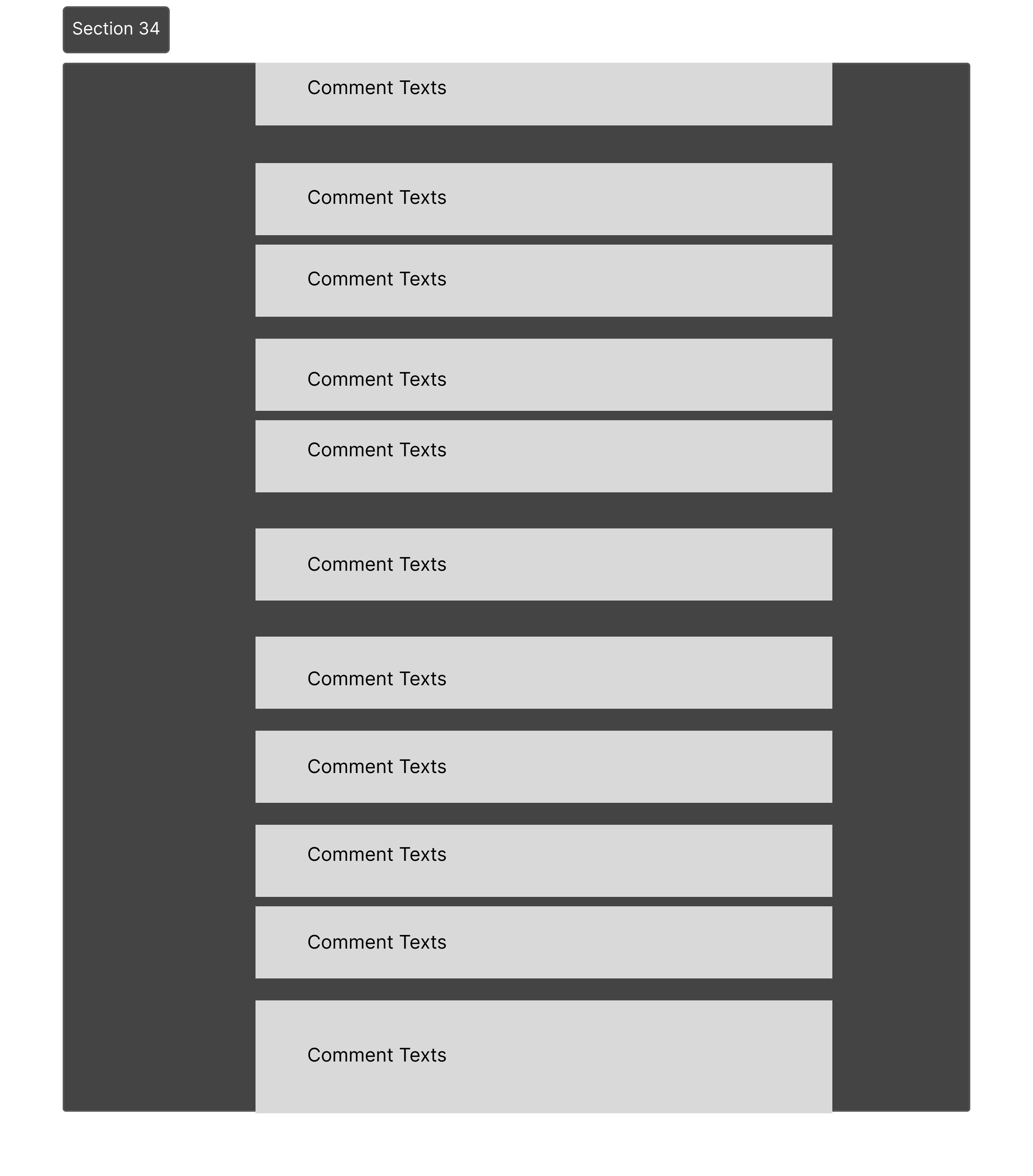}
        \end{subfigure} &
        \begin{subfigure}[t]{4.8cm}
            \centering
            \includegraphics[width=\linewidth, keepaspectratio]{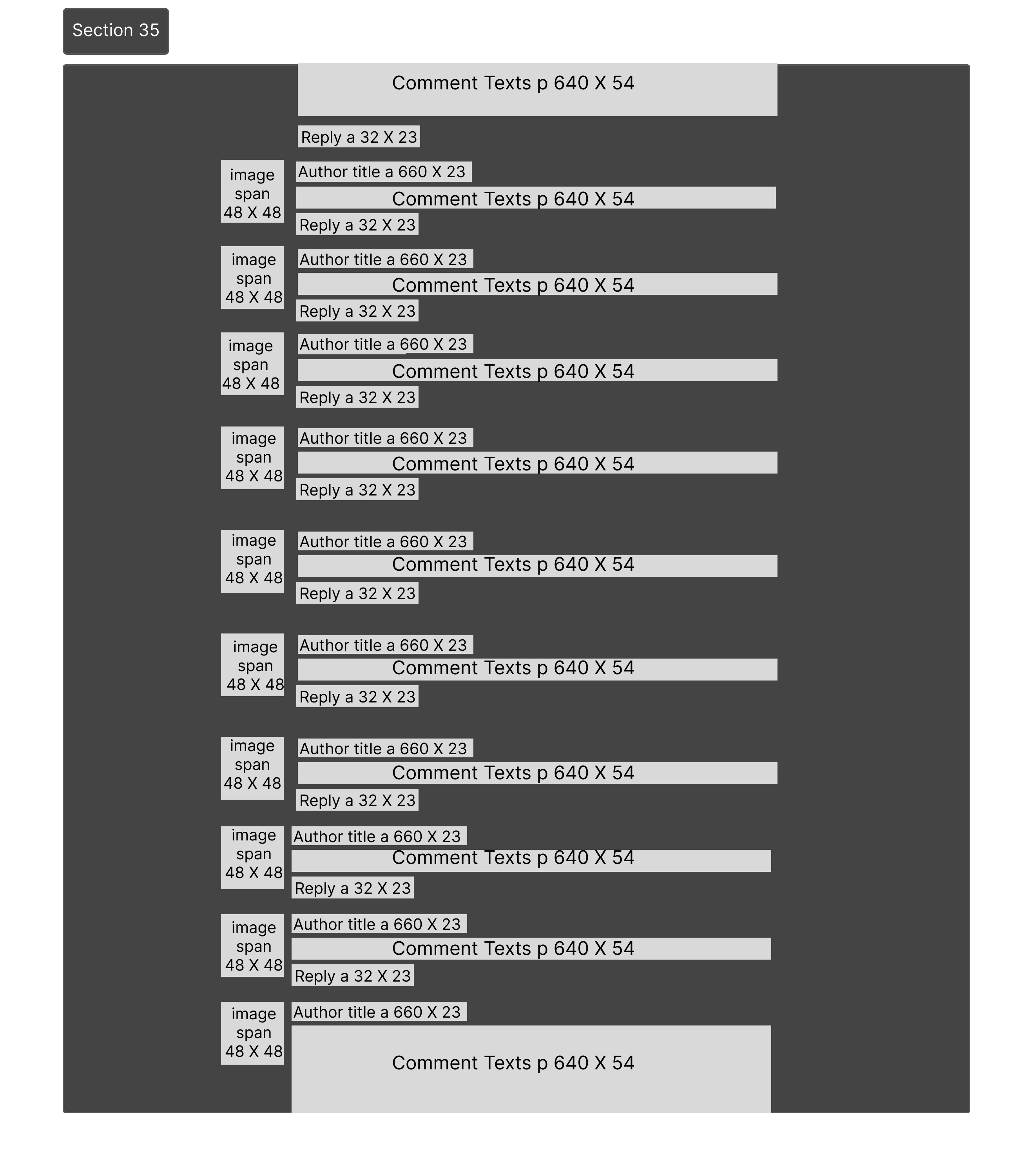}
        \end{subfigure} \\

        \midrule

        \begin{subfigure}[t]{4.8cm}
            \centering
            \includegraphics[width=\linewidth, keepaspectratio]{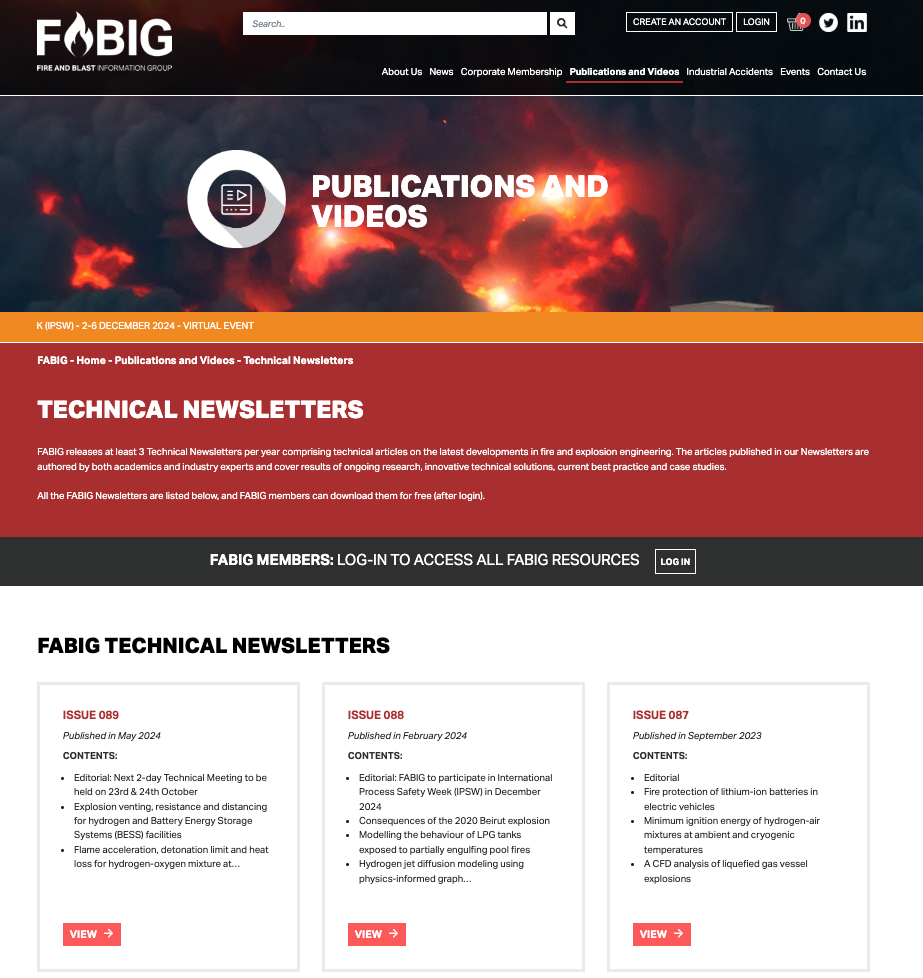}
        \end{subfigure} &
        \begin{subfigure}[t]{4.8cm}
            \centering
            \includegraphics[width=\linewidth, keepaspectratio]{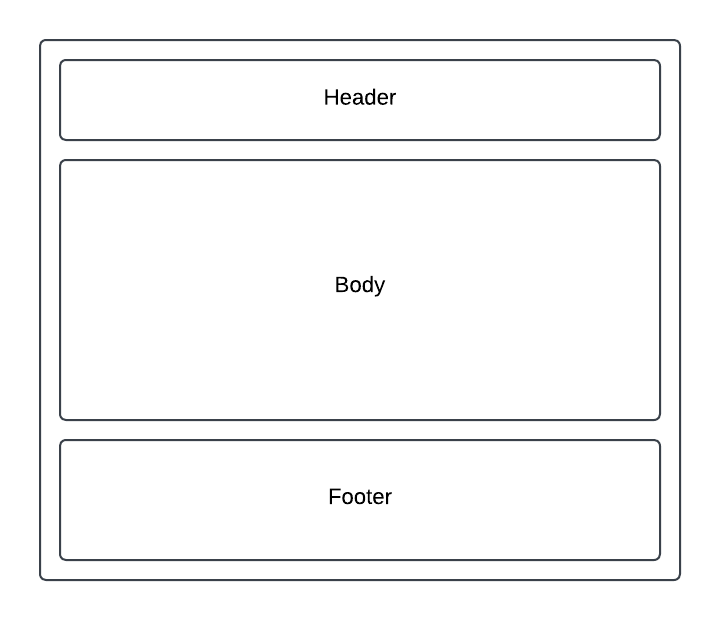}
        \end{subfigure} &
        \begin{subfigure}[t]{4.8cm}
            \centering
            \includegraphics[width=\linewidth, keepaspectratio]{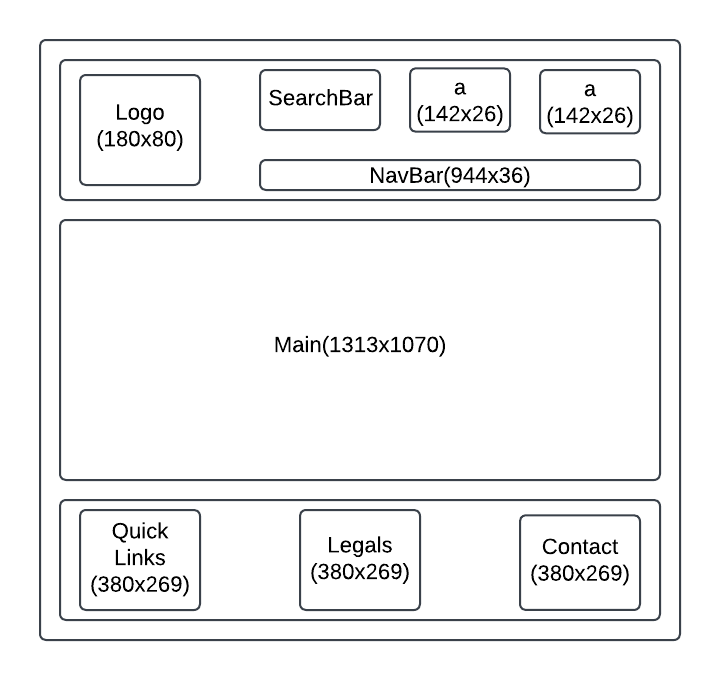}
        \end{subfigure} \\

        \bottomrule
    \end{tabular}
    \caption{\textbf{\mockupNoStyle{} Task Samples.} This table showcases examples from the \mockupNoStyle{} task, illustrating the transformation of webpage images into structured representations. Each row includes: (1) an Input Image (webpage screenshot), (2) a Simple Sketch (basic layout structure), and (3) a Complex Sketch (detailed UI components and text placements).}
    \label{tab:sketch2html_edit_task_samples}
\end{table*} 

\begin{table*}[h]
    \small
    \centering
    \renewcommand{\arraystretch}{1.5} 
    
    \resizebox{\textwidth}{!}{
    \begin{tabular}{>{\centering\arraybackslash}m{4.5cm} >{\arraybackslash}m{6cm} >{\centering\arraybackslash}m{4.5cm}}
        \toprule
        \textbf{Input Image} & \textbf{Task Description} & \textbf{Rendered Image} \\
        \midrule

        \includegraphics[width=4cm, keepaspectratio]{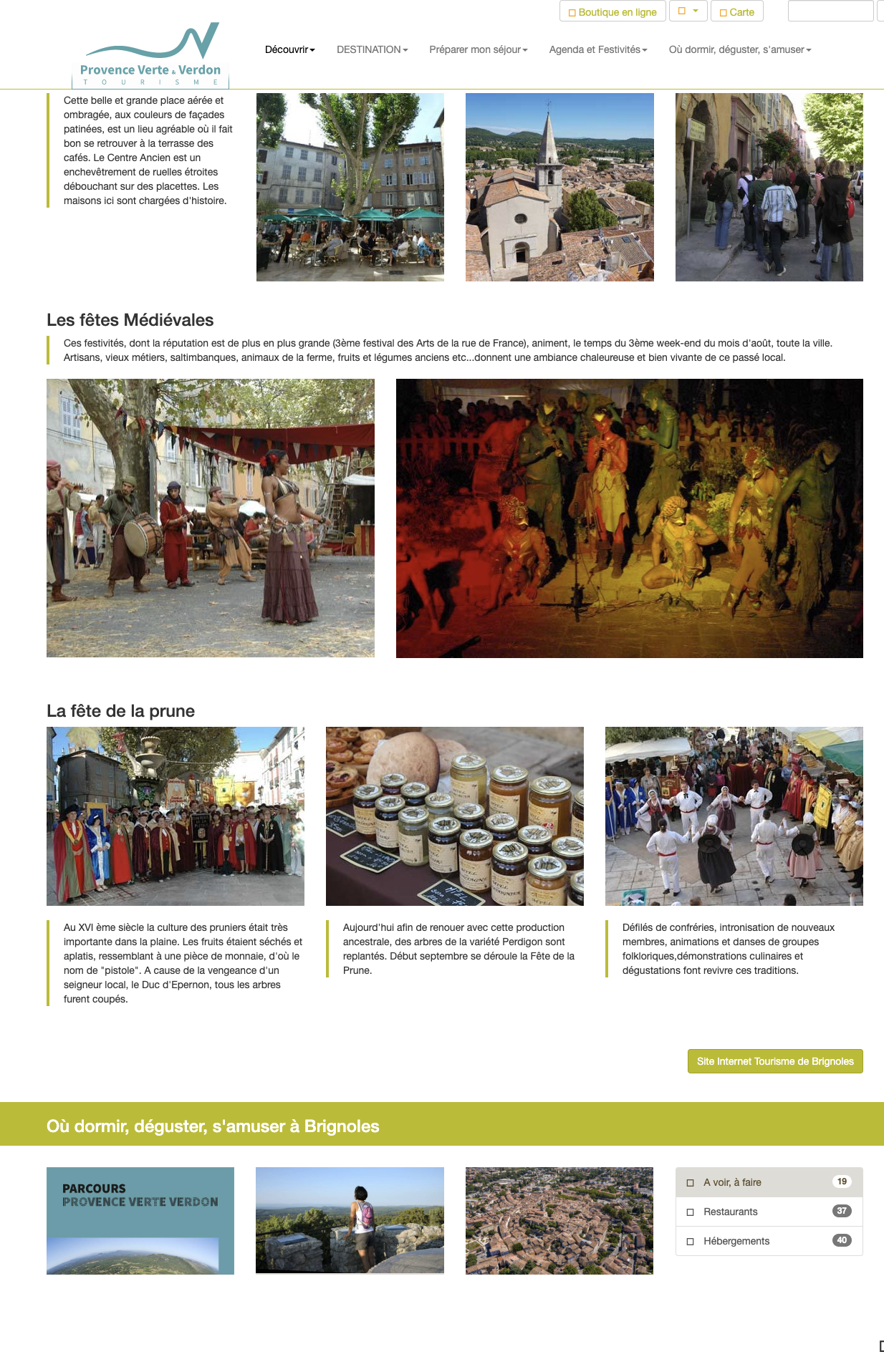} 
        & Comment faire pour afficher les différentes sections de l'article ``La fête de la prune'' en colonne et agrandir les images ?
        & \includegraphics[width=4cm, keepaspectratio]{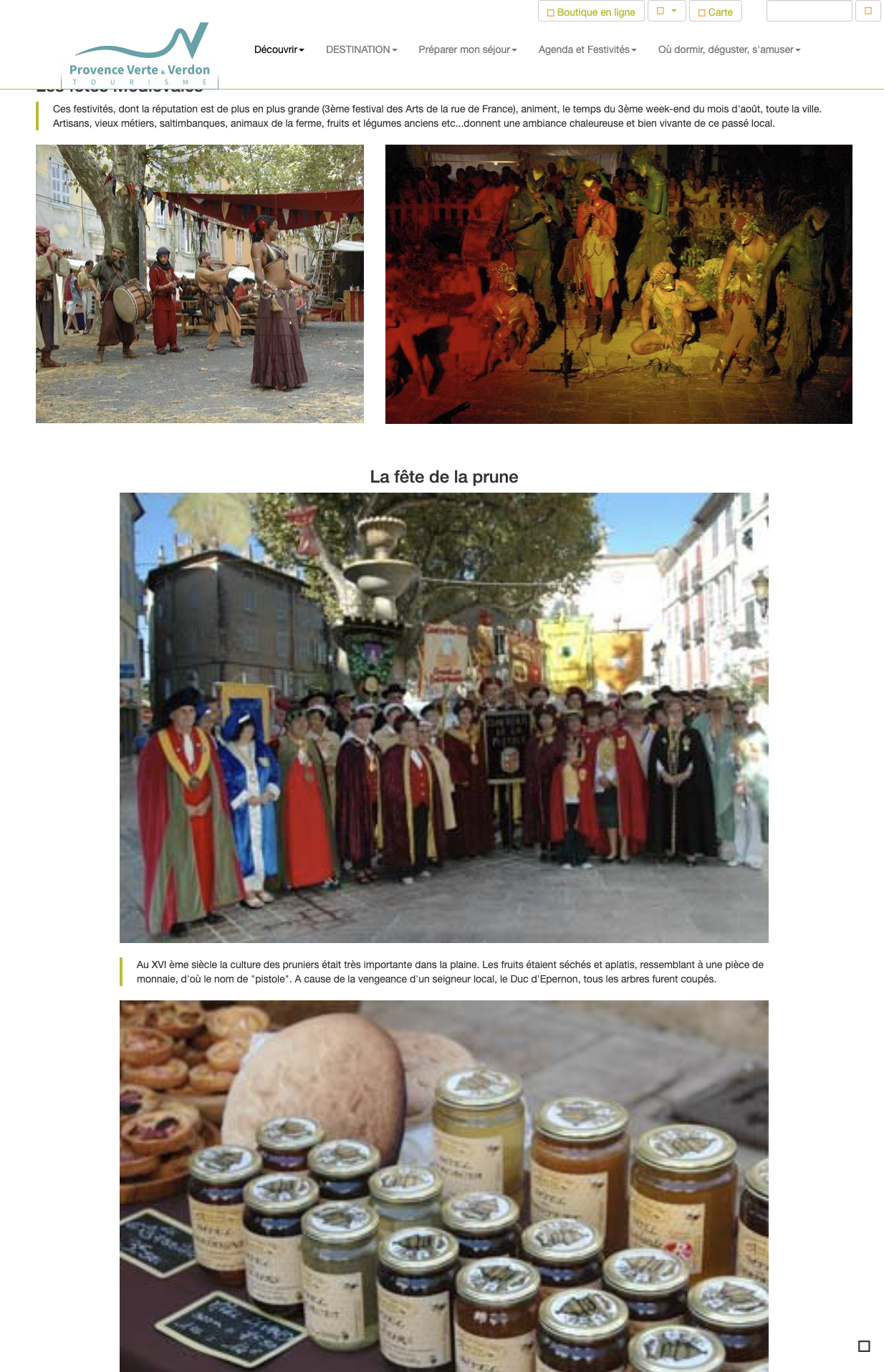} \\
        \midrule

        \includegraphics[width=4cm, keepaspectratio]{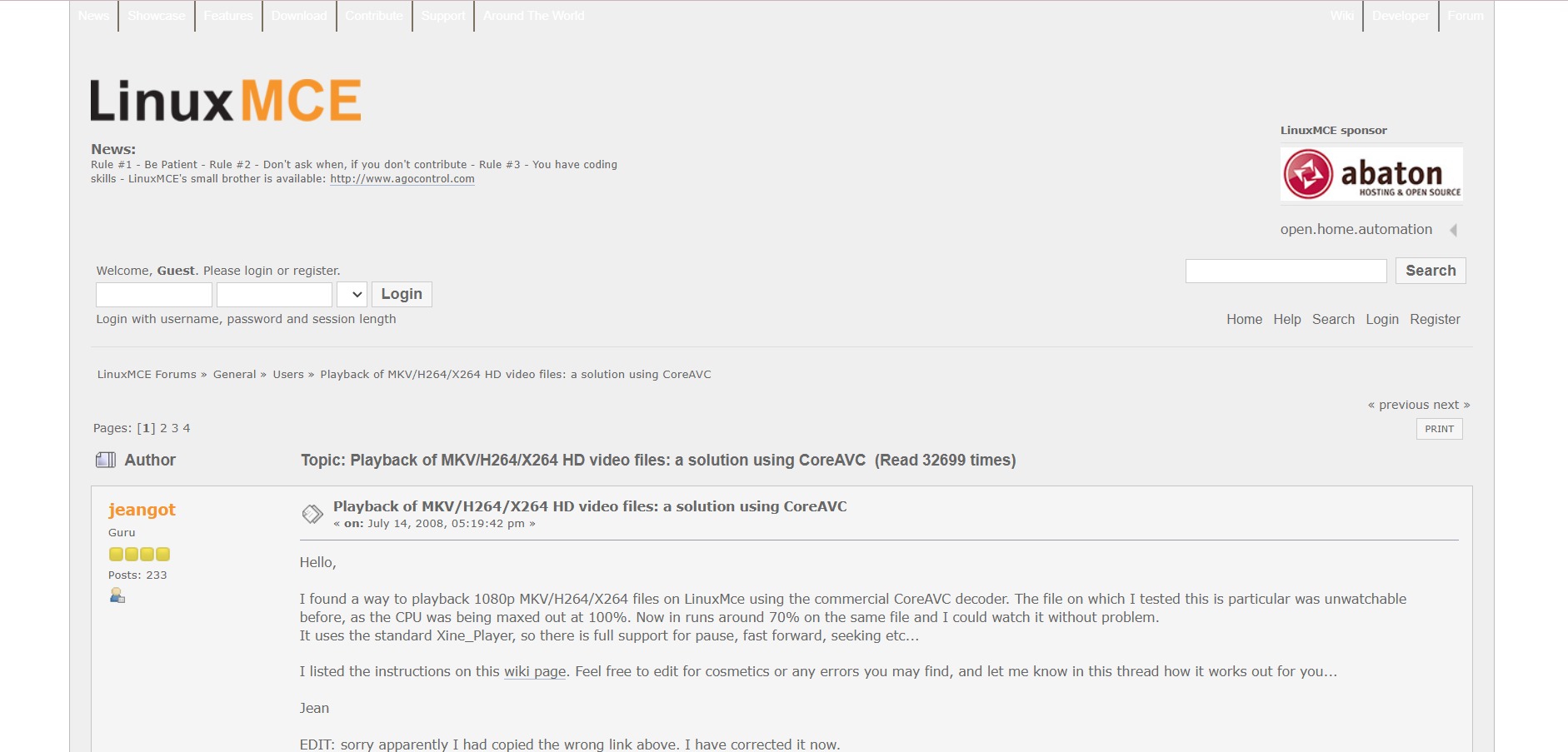} 
        & Which changes should be made in the HTML code to improve the UI of the login form and navbar?
        & \includegraphics[width=4cm, keepaspectratio]{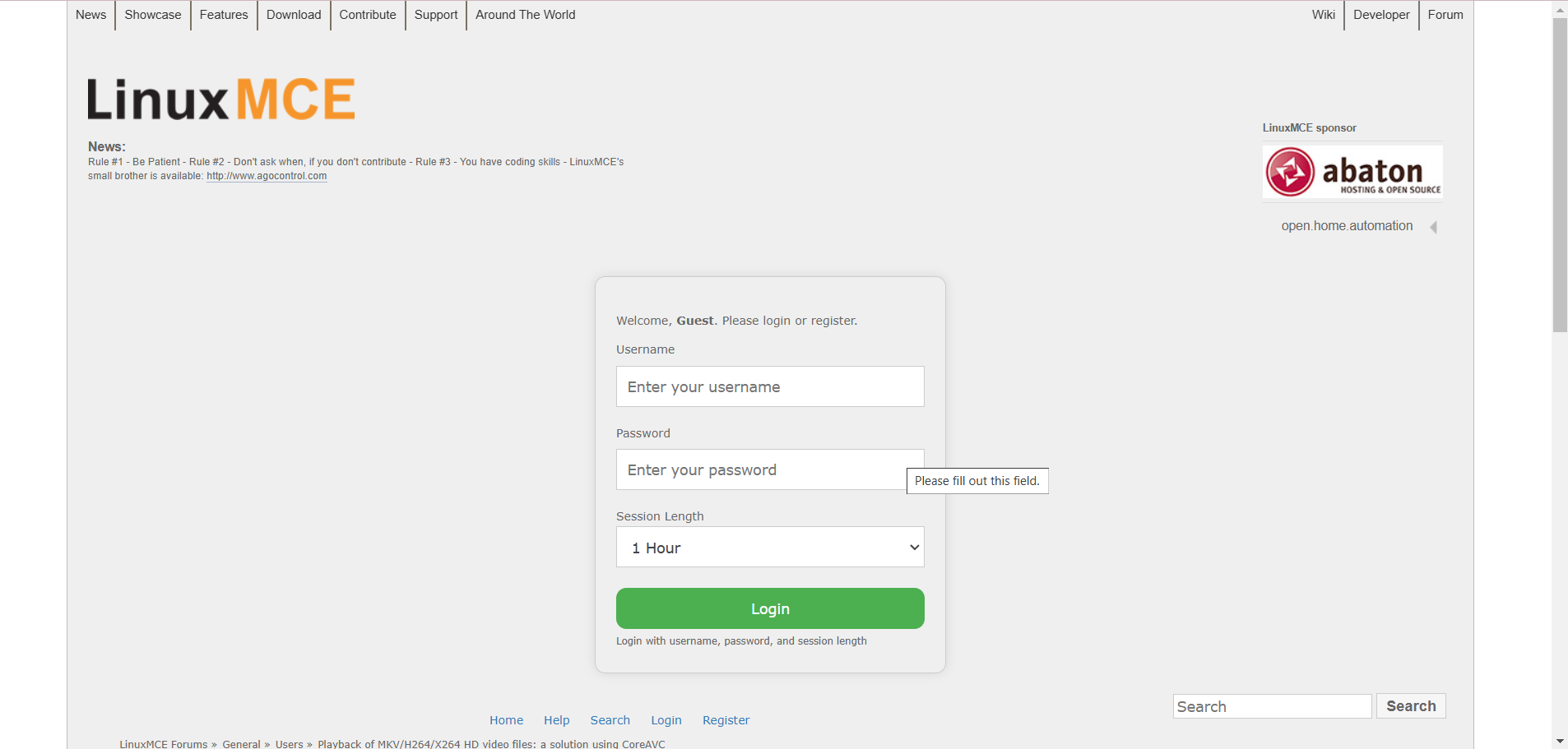} \\
        \midrule

        \includegraphics[width=4cm, keepaspectratio]{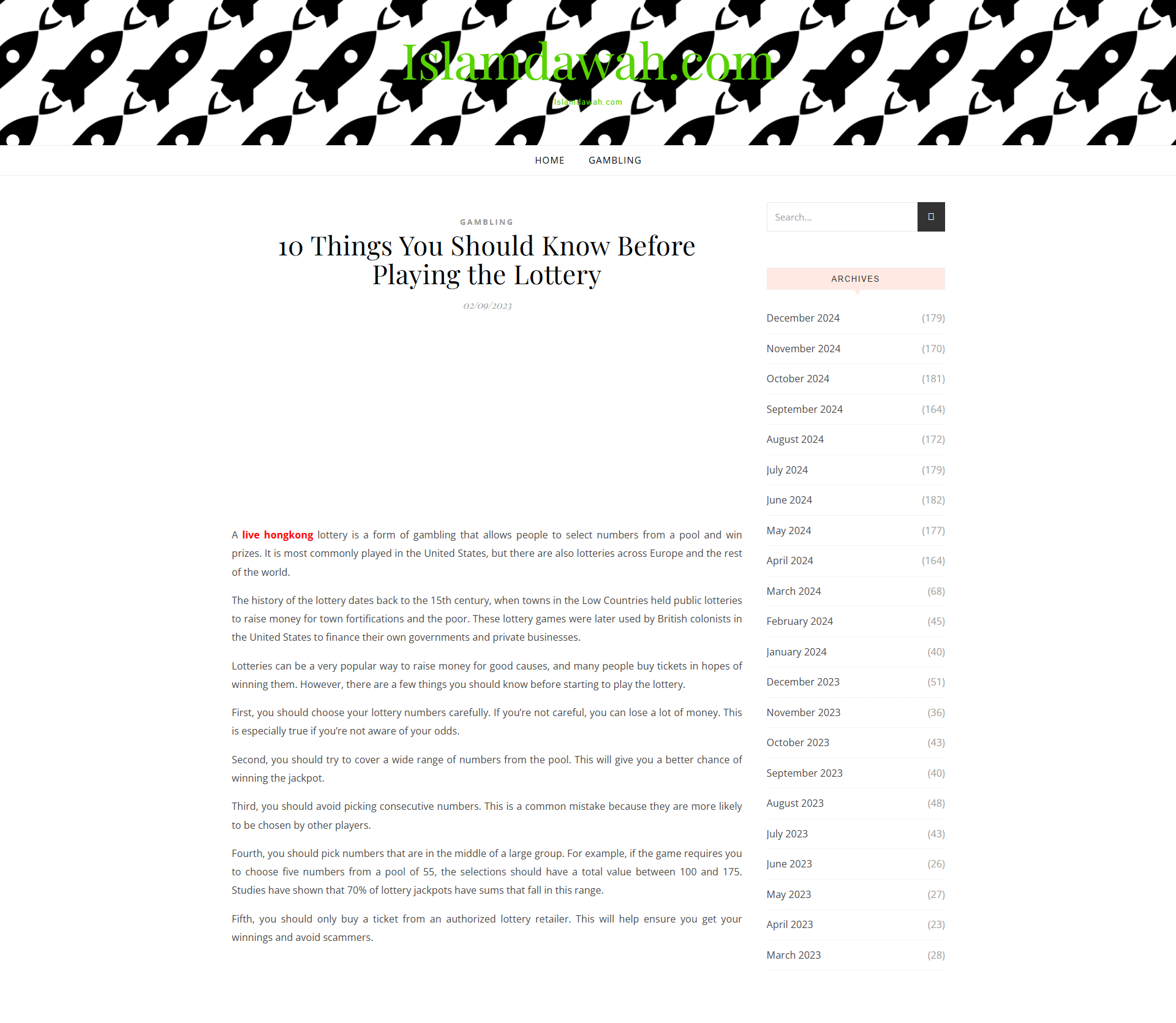} 
        & How can I fix the header element by adding a black overlay over the image, changing the font color to white, and setting the font family to ``Lucida Sans''?
        & \includegraphics[width=4cm, keepaspectratio]{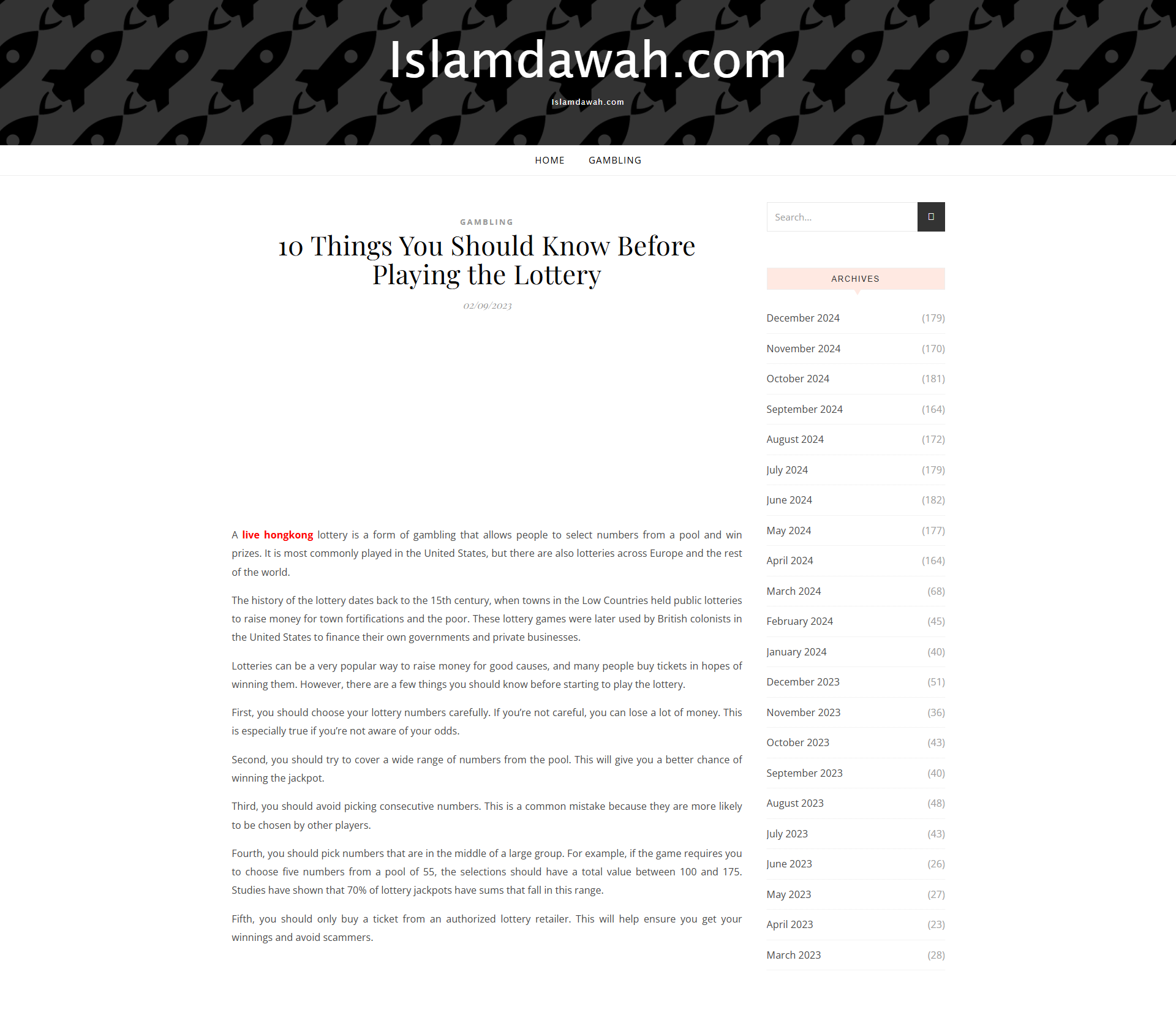} \\
        \midrule

        \includegraphics[width=4cm, keepaspectratio]{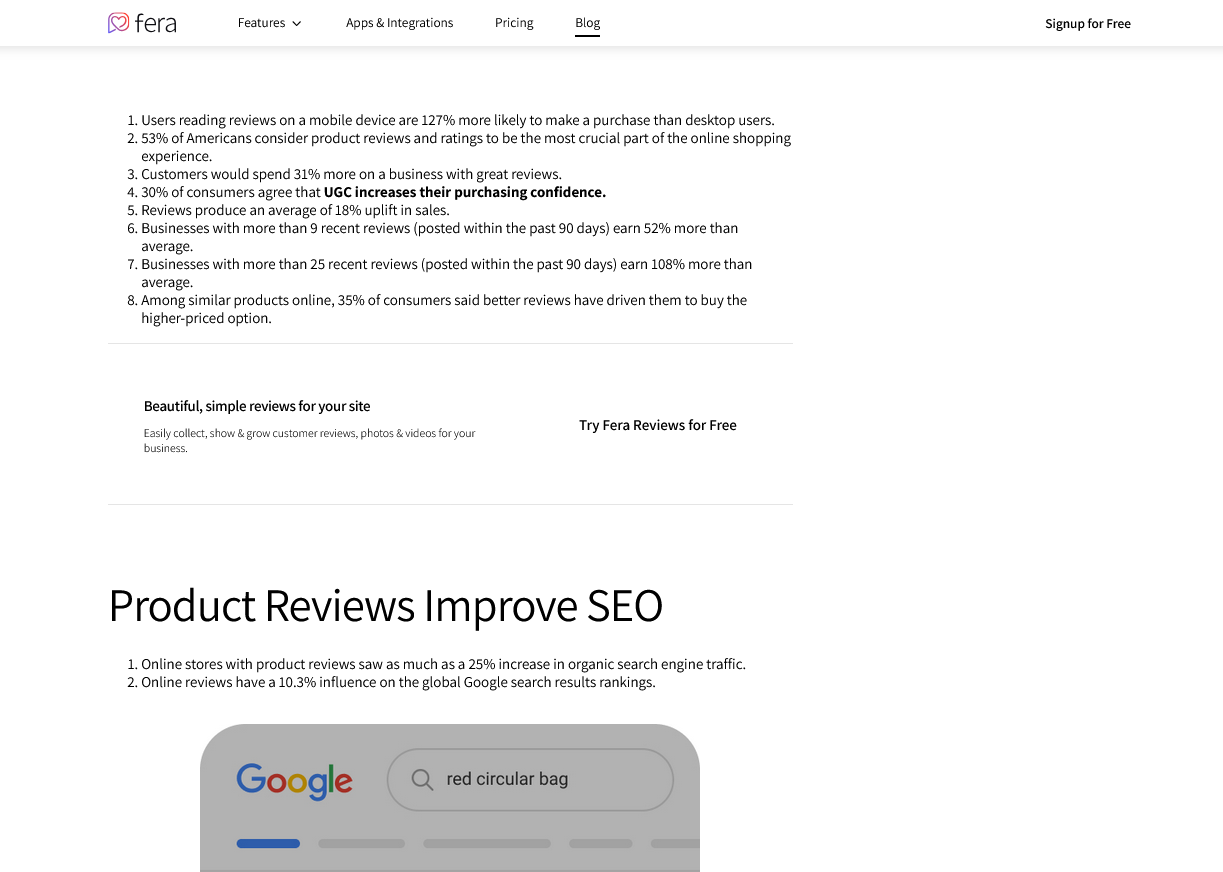} 
        & For a visually appealing design, enhance the navbar with hover and shadow effects, add hover interactions to buttons and links, and apply a card effect to containers.
        & \includegraphics[width=4cm, keepaspectratio]{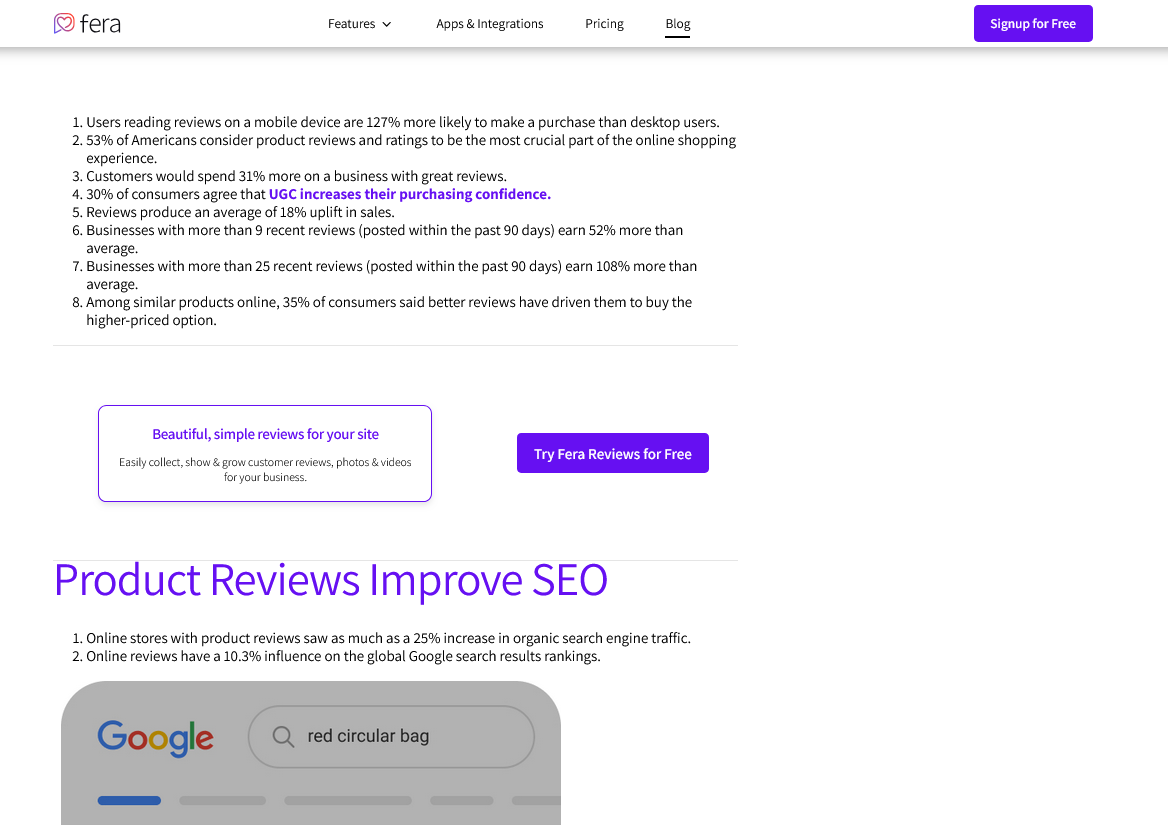} \\
        \midrule

        \includegraphics[width=4cm, keepaspectratio]{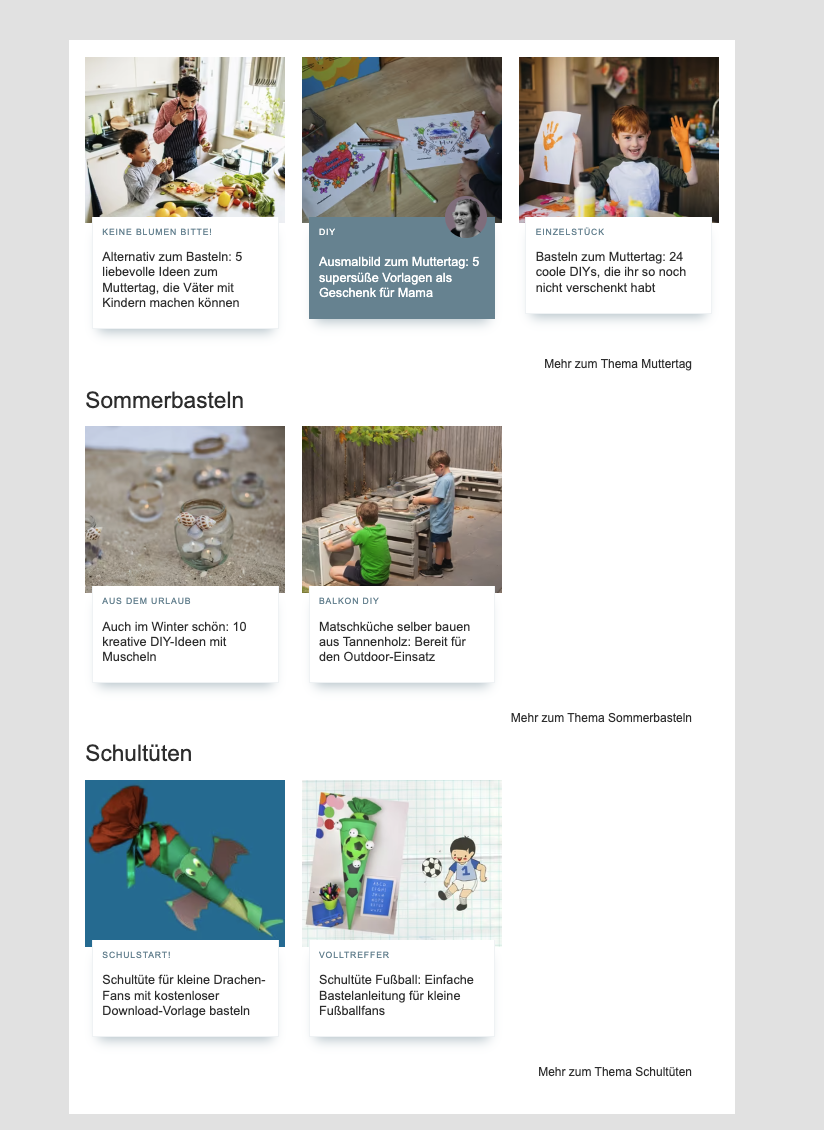} 
        & Mache die drei Felder ``link-next" auffälliger, indem du ihre Farbe, Größe oder Schriftstil anpasst.
        & \includegraphics[width=4cm, keepaspectratio]{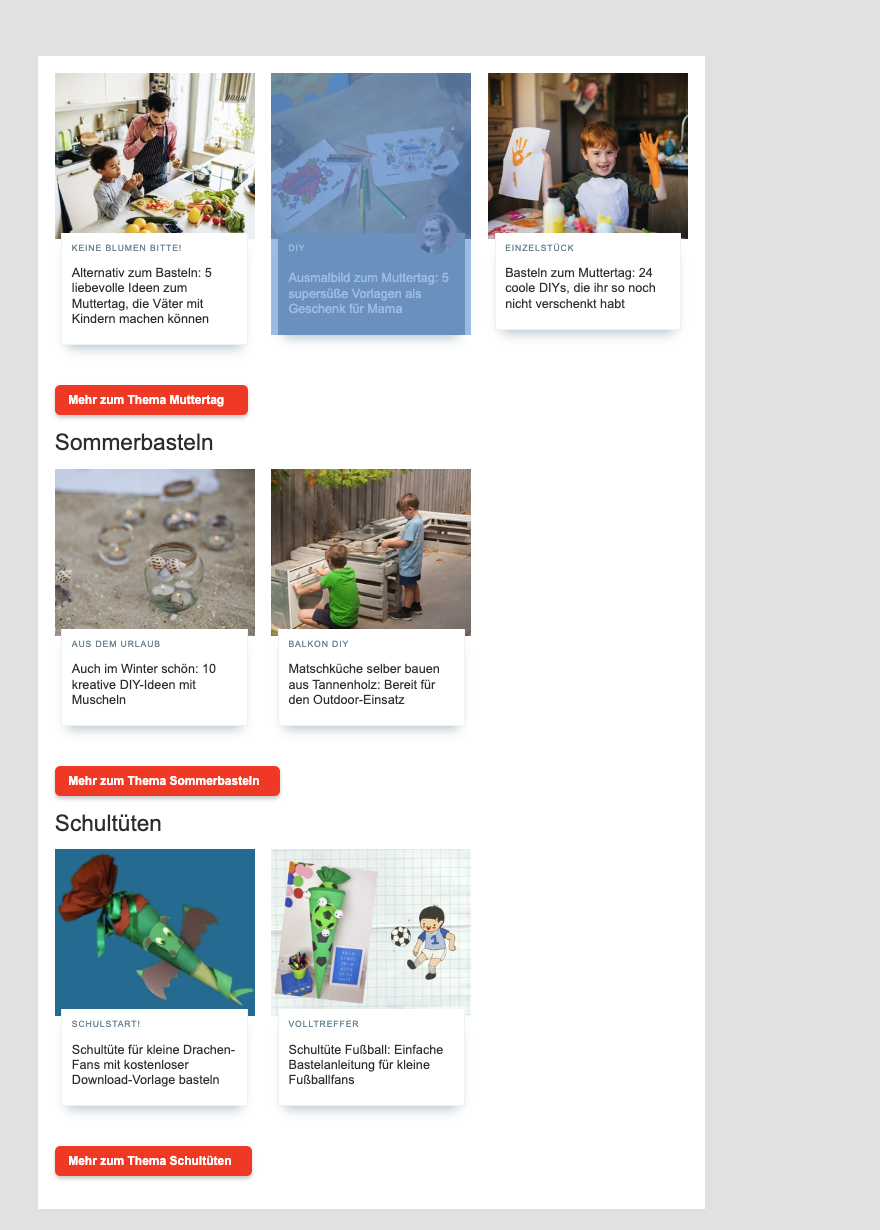} \\

        \bottomrule
    \end{tabular}
    } 
    \caption{\textbf{\codeeditNoStyle{} Task Samples.} Code edition before and after screenshot of webpage.}
    \label{tab:code_edit_task_samples}
\end{table*}

\section{Model Output Generation Prompts}
\subsection{\wqaNoStyle{} Task Completion Prompt}
\label{sec:wqa_generation_prompt}
We present the prompt used for the \wqaNoStyle{} task in Figure~\ref{fig:vqa-generation-prompt}. The prompt instructs the model to analyze a website screenshot and provide a concrete answer to the given question. When the question requires identifying or interacting with specific elements on the screen, the model is asked to include the bounding box coordinates in its response.

\begin{figure}[h]
  \centering
\begin{tcolorbox}[mypromptstyle, title={Web QA Inference}]
    Analyze the website screenshot and provide a detailed answer to the question. 
    If the question involves locating or interacting with specific elements on the screen, 
    include the bounding box coordinates [x\_min, y\_min, x\_max, y\_max] in your response.
\end{tcolorbox}
\caption{Prompt for Generating Output of \wqaNoStyle{} task}
\label{fig:vqa-generation-prompt}
\end{figure}

\subsection{\codeeditNoStyle{} Task Completion Prompt}
\label{sec:code_edition_generation_prompt}
This prompt guides a model in modifying the source code based on a modification instruction given by the user. The model outputs changes using the \texttt{git diff} format, highlighting additions and deletions with `+'s and `-'s respectively. This ensures clear and structured documentation of code edits. 
The prompt template can be seen in Figure ~\ref{fig:web-code-edit-generation-prompt}.

\begin{figure*}[h]
\centering
\begin{tcolorbox}[mypromptstyle, title={\codeeditNoStyle{} Generation Prompt}]

You are an expert web developer specializing in identifying and applying modifications to web code. You will receive a website's screenshot and a combination of it's HTML, CSS, and/or JavaScript code, formatted as follows:

\begin{itemize}[leftmargin=2em, itemsep=1pt, parsep=0pt]
    \item \textbf{HTML Code:} \texttt{html\_code}
    \item \textbf{CSS Code:} \texttt{css\_code}
    \item \textbf{JavaScript Code:} \texttt{javascript\_code}
\end{itemize}
You will also receive a modification prompt describing the required changes.
Your task is to produce the necessary code modifications using `git diff' format, even if some or all sections are missing. Follow these guidelines:
\begin{enumerate}[leftmargin=2em, itemsep=1pt, parsep=0pt]
    \item \textbf{Input code:} \texttt{<input\_code>}
    \item \textbf{Modification Prompt:} \texttt{<edit\_prompt>}
    \item \textbf{Output Diff:}  
    \begin{itemize}
        \item Use `+' for additions and `-' for deletions.
        \item Modify only the relevant parts while preserving structure.  
        \item In case the code is missing, generate the necessary block of code from scratch.
        \item Ensure readability and correctness in the modifications. 
    \end{itemize}
\end{enumerate}
Only output the necessary diff; do not repeat the input code.
\end{tcolorbox}
\caption{\codeeditNoStyle{} generation prompt}
\label{fig:web-code-edit-generation-prompt}
\end{figure*}

\subsection{\mockupNoStyle{} Task Completion Prompt}
The \mockupNoStyle{} task involves generating HTML/CSS code from an input sketch (see Figure~\ref{fig:ux-generation-prompt}). Given a visual layout, the model must produce accurate, well-structured HTML and CSS that replicate the design. The prompt guides the model to interpret elements, hierarchy, and styling for faithful image-to-code conversion.

\begin{figure*}[h]
  \centering
  \begin{tcolorbox}[mypromptstyle, title={\textbf{\mockupNoStyle{} Generation Prompt}}]
    \textbf{You are an expert website developer.} Analyze the provided webpage sketch and generate a single, fully structured HTML file with embedded CSS that accurately reflects the design.
    
    The output must be a self-contained HTML document with internal \texttt{<style>} tags for CSS. Ensure all elements are structured exactly as seen in the sketch—\textbf{no extra elements, no missing elements}.
    
    \textbf{HTML Requirements:}
    \begin{itemize}[leftmargin=2em, itemsep=1pt, parsep=0pt]
      \item \textbf{Components:} Include all necessary components such as headers, paragraphs, buttons, forms, and images, maintaining the correct hierarchy and placement.
      \item \textbf{Images:} Use images generated from \url{https://placehold.co/} with exact dimensions matching the sketch, a neutral background color, and centered “Image” text. For example: <img src="https://placehold.co/300x200?text=Image\&bg=cccccc" alt="Placeholder Image">
      
      \item \textbf{Placeholder Text:} Use Lorem Ipsum for placeholder text where needed.
    \end{itemize}
    
    \textbf{CSS Requirements:}
    \begin{itemize}[leftmargin=2em, itemsep=1pt, parsep=0pt]
      \item Implement CSS directly within the HTML file (inside a \texttt{<style>} block) to match the sketch, covering spacing, font sizes, colors, alignments, and element positioning.
      \item Use CSS Grid or Flexbox where appropriate to replicate the exact design layout.
      \item Apply styling for readability and interactive elements (e.g., fonts, button appearance).
      \item Ensure placeholder images maintain proper dimensions and design consistency.
    \end{itemize}
    
    \textbf{Code Output:}
    \begin{itemize}[leftmargin=2em, itemsep=1pt, parsep=0pt]
      \item Provide a single, complete HTML file with internal CSS (do not separate them into different files).
      \item Do not include explanations, comments, or any extra formatting outside the code itself.
    \end{itemize}
    
  \end{tcolorbox}
  \caption{\mockupNoStyle{} Generation Prompt: It takes input sketch and outputs HTML/CSS code of the given input}
  \label{fig:ux-generation-prompt}
\end{figure*}

\section{Evaluation Prompts}
\label{appendix:prompt_formulation}
This section provides details on the prompt formulations used throughout this work. These prompts guide the multimodal large language models in generating and evaluating responses across different tasks. The prompts are categorized based on their usage, including code modification, VQA evaluation, and UX scoring.

\subsection{\wqaNoStyle{} Evaluation Prompt}
\label{sec:vqa_judge_prompts}
These prompts are used for evaluating model responses in VQA tasks. The model rates answers as 1 (Correct and Complete) or 0 (Incorrect or Irrelevant) based on factual accuracy and completeness. Example cases are provided to guide the evaluation. The prompt template can be seen in Figure ~\ref{fig:vqa_judge_prompt}.

\begin{figure*}[h]
  \centering
\begin{tcolorbox}[mypromptstyle, title={Web QA Evaluation Prompt}]
\small
\texttt{
examples = [\\
\hspace*{4mm} \{\\
\hspace*{8mm} "INPUT": \{\\
\hspace*{12mm} "question": "What is the capital of France?",\\
\hspace*{12mm} "model\_answer": "Paris",\\
\hspace*{12mm} "ground\_truth": "Paris",\\
\hspace*{8mm} \},\\
\hspace*{8mm} "OUTPUT": \{\\
\hspace*{12mm} "rating": 1,\\
\hspace*{12mm} "rationale": "The model's answer matches the reference answer exactly."\\
\hspace*{8mm} \}\\
\hspace*{4mm} \},\\
\hspace*{4mm} \{\\
\hspace*{8mm} "INPUT": \{\\
\hspace*{12mm} "question": "What is in the left of the image?",\\
\hspace*{12mm} "model\_answer": "A bus is in the left of the image.",\\
\hspace*{12mm} "ground\_truth": "A dog is in the left of the image.",\\
\hspace*{8mm} \},\\
\hspace*{8mm} "OUTPUT": \{\\
\hspace*{12mm} "rating": 0,\\
\hspace*{12mm} "rationale": "The model's answer is incorrect because the reference answer is 'A dog'."\\
\hspace*{8mm} \}\\
\hspace*{4mm} \},\\
\hspace*{4mm} \{\\
\hspace*{8mm} "INPUT": \{\\
\hspace*{12mm} "question": "Where is the burger on the table? Tell me the coordinates.",\\
\hspace*{12mm} "model\_answer": "The burger is on the table.",\\
\hspace*{12mm} "ground\_truth": "The burger is on the table at (50, 10, 150, 60).",\\
\hspace*{8mm} \},\\
\hspace*{8mm} "OUTPUT": \{\\
\hspace*{12mm} "rating": 0,\\
\hspace*{12mm} "rationale": "The predicted answer is incomplete because it does not provide the coordinates as requested in the question."\\
\hspace*{8mm} \}\\
\hspace*{4mm} \}\\
]\\
test\_case = \{\\
\hspace*{4mm} "INPUT": \{\\
\hspace*{8mm} "question": question,\\
\hspace*{8mm} "model\_answer": model\_answer,\\
\hspace*{8mm} "ground\_truth": ground\_truth\\
\hspace*{4mm} \}\\
\}\\
}

You are evaluating a Visual Question Answering (VQA) system's response. Compare the model's answer with the ground truth and rate its accuracy.

\textbf{Rating Scale (1 or 0):}\\
1 - \textbf{Correct and Complete}:
- The predicted answer fully matches the ground truth.
- No factual errors or missing details.
- Addresses the question with the correct level of specificity.

0 - \textbf{Incorrect or Irrelevant}:
- Any factual errors or mismatches with the reference answer.
- Does not address the question properly.
- Provides misleading or irrelevant information.

\textbf{Examples for reference:}
{json.dumps(examples, indent=4)}

\textbf{Question, Model Answer, and Ground Truth:}
{json.dumps(test\_case, indent=4)}

You must provide your evaluation in the following JSON format (without any extra text):
{json.dumps({"rating": 0 or 1, "rationale": "[Brief explanation of why this rating was chosen]"})}
"""
\end{tcolorbox}
  \caption{LLM-as-judge prompt for \wqaNoStyle{} task using few-shot examples}
  \label{fig:vqa_judge_prompt}
\end{figure*}

\subsection{\mockupNoStyle{} Evaluation Prompt}
\label{sec:mockup_evaluation_prompt}

The \mockupNoStyle{} evaluation task involves assessing the accuracy of an MLLM-generated website based on an input sketch (see Figure~\ref{fig:mockup_evaluation_prompt}). The evaluation prompt directs the annotator to compare the AI-generated HTML/CSS output with the given visual layout, ensuring that the generated website accurately replicates the design in terms of structure, styling, and layout. The evaluation criteria focus on layout structure, spacing, proportions, and alignment, allowing for a detailed assessment of how closely the generated output matches the intended design. The goal is to evaluate the model's ability to interpret and transform the sketch into a functional, visually consistent website.

\begin{figure*}[h]
  \centering
  \small
  \begin{tcolorbox}[    colback=white,
    colframe=prompt1-frame,
    arc=2pt,
    left=4pt, right=4pt, top=4pt, bottom=4pt,
    before skip=10pt, after skip=10pt,
    bottomrule=0pt, 
    fonttitle=\bfseries,
    coltitle=white, title={\textbf{\mockupNoStyle{} Evaluation Prompt}}]

\textbf{Task Overview:} 
Your task is to evaluate the accuracy of an AI-generated website by comparing it against a provided input sketch. The AI-generated website is provided as an image rendering of the HTML/CSS output. Your goal is to assess how well this rendered image replicates the intended layout from the sketch.

\textbf{Provided Inputs:} You will receive two images: 
\begin{enumerate}[leftmargin=2em, itemsep=1pt, parsep=0pt]
    \item \textbf{Input Sketch} – A wireframe illustrating the intended layout.
    \item \textbf{Predicted AI-Rendered Website Image} – A screenshot of the website generated from AI-created HTML/CSS based on the sketch.
\end{enumerate}

Since the AI-generated website is provided as an image, your evaluation must be based entirely on visual accuracy, disregarding the underlying code implementation.

\textbf{Step 1: Detailed Description of Both Images} \\
For each image (\textbf{Input Sketch} and \textbf{AI-Rendered Website}), provide a highly-detailed breakdown based on the following categories. Ensure that descriptions follow the same format for both images to facilitate a precise comparison.

\textbf{1. Identify All Structural Sections:}\\ 
Describe in detail the overall structure of the webpage layout, covering the following: 
\begin{itemize}[leftmargin=2em, itemsep=1pt, parsep=0pt]
  \item \textbf{Header} – Does it contain a logo, navigation menu, search bar, or other elements? 
  \item \textbf{Navigation Bar} – Describe the menu items. How many items are there? Is the navigation horizontal or vertical? 
  \item \textbf{Main Content Area} – Identify distinct sections such as hero banners, text areas, images, or interactive components. 
  \item \textbf{Sidebars (if applicable)} – Is there a sidebar for additional navigation, filters, or widgets?
  \item \textbf{Footer} – What content is present (e.g., links, social icons, contact information)?
\end{itemize}

For the AI-rendered website, note any differences compared to the sketch (e.g., missing sections, extra sections, missing items, misplaced content).

\textbf{2. List and Describe All Elements:} \\
List all key elements present in the \textbf{Input Sketch} and \textbf{AI-Rendered Website}:
\begin{itemize}[leftmargin=2em, itemsep=1pt, parsep=0pt]
  \item \textbf{Text Elements} – Titles, paragraphs, labels, lists, captions.
  \item \textbf{Images \& Icons} – Identify all image placeholders and their intended placement.
  \item \textbf{Buttons \& Links} – Describe all interactive elements like CTAs, navigation links, or form buttons.
  \item \textbf{Forms \& Inputs} – Search bars, text fields, dropdowns, checkboxes, radio buttons, etc.
  \item \textbf{Tables \& Lists} – If present, describe their structure and formatting.
\end{itemize}
For the AI-rendered website, specify any elements that are missing, added, or incorrectly placed.

\textbf{3. Layout \& Positioning Details:} \\
Describe and analyze the spatial arrangement of elements in both images:
\begin{itemize}[leftmargin=2em, itemsep=1pt, parsep=0pt]
  \item \textbf{Column Structure} – Is the design single-column, multi-column, or grid-based?
  \item \textbf{Alignment} – Are elements aligned left, center, right, or justified?
  \item \textbf{Spacing \& Proportions} – Are elements evenly spaced? Are margins, padding, and gaps consistent?
  \item \textbf{Relative Proportions} – Are certain sections (e.g., hero banners, sidebars) larger than others?
\end{itemize}
For the AI-rendered website, describe any deviations from the sketch (e.g., elements’ size differences, elements too large/small, uneven spacing, misalignments).

\textbf{Step 2: Evaluation of the AI-Rendered Website} \\
After describing both images, evaluate the AI-generated website’s accuracy using the following criteria. Assign a score from 1 to 5 for each.

\textbf{1. Layout Structure Accuracy (1-5):} \\
Does the generated HTML structure strictly follow the wireframe in layout, hierarchy, and element grouping? This includes the correct placement, nesting, and semantic usage of standard structural elements: \texttt{<header>}, \texttt{<nav>}, \texttt{<main>}, \texttt{<section>}, \texttt{<aside>}, \texttt{<article>}, \texttt{<footer>}, \texttt{<div>}, and content containers like \texttt{<img>}, \texttt{<p>}.
\begin{itemize}[leftmargin=2em, itemsep=1pt, parsep=0pt]
  \item \textbf{5} $\rightarrow$ 100\% match. All elements are correctly placed, properly nested, fully grouped, and semantically accurate. No missing, misplaced, or extra elements.
  \item \textbf{4} $\rightarrow$ Mostly accurate, but minor structural inconsistencies exist (e.g., an unnecessary wrapper, slightly misplaced section, or minor redundancy). No missing elements.
  \item \textbf{3} $\rightarrow$ Some structural errors — at least one missing or misused element, multiple misplaced sections, or noticeable grouping issues.
  \item \textbf{2} $\rightarrow$ Major deviations — multiple missing, misplaced, or incorrectly nested elements, affecting hierarchy and readability.
  \item \textbf{1} $\rightarrow$ Severe structural failure — multiple core sections are absent or completely misstructured, making the output unrecognizable compared to the wireframe.
\end{itemize}

  \end{tcolorbox}
\end{figure*}

\begin{figure*}[h]
  \centering
  \small
  \begin{tcolorbox}[    colback=white,
    colframe=prompt1-frame,
    arc=2pt,
    left=4pt, right=4pt, top=4pt, bottom=4pt,
    before skip=10pt, after skip=10pt,
    fonttitle=\bfseries,
    toprule=0pt, 
    coltitle=white, title={}]

\textbf{2. Spacing \& Proportions (1-5):} \\
Do margins, paddings, and element dimensions (e.g., \texttt{width}, \texttt{height}, \texttt{max-width}, \texttt{min-width}, \texttt{max-height}, \texttt{min-height}, gap for flex/grid layouts) precisely match the wireframe?
\begin{itemize}[leftmargin=2em, itemsep=1pt, parsep=0pt]
  \item \textbf{5} $\rightarrow$ 100\% correct. All elements have precise margins, paddings, widths, heights, and spacing. No deviations.
  \item \textbf{4} $\rightarrow$ Minor inconsistencies exist (e.g., slightly incorrect padding/margin values or minor width/height variations).
  \item \textbf{3} $\rightarrow$ Noticeable discrepancies — some elements are too large, too small, or unevenly spaced, affecting visual balance.
  \item \textbf{2} $\rightarrow$ Significant spacing issues — multiple elements have incorrect dimensions, margins, or paddings, leading to a visibly distorted layout.
  \item \textbf{1} $\rightarrow$ Severe inaccuracies — most elements have incorrect proportions or spacing, making the layout visually broken and inconsistent with the wireframe.
\end{itemize}

\textbf{3. Alignment \& Grid Consistency (1-5):} \\
Are elements precisely aligned according to the wireframe, following the expected grid/flex structure and ensuring uniform positioning?
\begin{itemize}[leftmargin=2em, itemsep=1pt, parsep=0pt]
  \item \textbf{5} $\rightarrow$ Perfect alignment. Every element follows the wireframe’s grid, flex, or positioning structure exactly. No misalignments.
  \item \textbf{4} $\rightarrow$ Mostly aligned, but minor deviations exist (e.g., slightly off-center text or small pixel variations in placement).
  \item \textbf{3} $\rightarrow$ Some clear misalignments — at least one noticeably off-grid or misplaced element that affects overall balance.
  \item \textbf{2} $\rightarrow$ Major alignment issues, with multiple elements misaligned, overlapping, or not following the expected structure.
  \item \textbf{1} $\rightarrow$ Severe disorganization — the output fails to follow the wireframe’s grid or positioning, making the layout appear chaotic.
\end{itemize}

\textbf{Final Score Calculation:} \\
Final Score = (Layout Structure Accuracy + Spacing \& Proportions + Alignment \& Grid Consistency) / 3

\textbf{Output Format:} \\
Your response must follow this JSON structure:
\texttt{\\
\hspace*{4mm} \{\\
\hspace*{8mm} "descriptions": \{\\
\hspace*{12mm} "input sketch": "provide the description of sketch here",\\
\hspace*{12mm} "AI-rendered website": "provide the description of website here"\\
\hspace*{8mm} \},\\
\hspace*{8mm} "scores": \{\\
\hspace*{12mm} "layout\_structure\_accuracy": [1-5],\\
\hspace*{12mm} "spacing\_proportions": [1-5],\\
\hspace*{12mm} "alignment\_grid\_consistency": [1-5]\\
\hspace*{8mm} \},\\
\hspace*{8mm} "final\_score": [calculated average score],\\
\hspace*{8mm} "reasoning": "[Concise evaluation highlighting key strengths and weaknesses]"\\
\hspace*{4mm} \}\\
}
  \end{tcolorbox}

  \caption{ LLM-as-Judge input prompt: It evaluates the model output and the ground truth among some detailed criteria given in the prompt.}
  \label{fig:mockup_evaluation_prompt}
\end{figure*}

\subsection{Code Edit Evaluation Prompt}
\label{app:judge_prompts}

This prompt is used to evaluate model responses in code edition tasks. The model rates answers as 1-5 (5 refers to the most correct and complete, and 1 refers to incorrect or irrelevant) based on factual accuracy and completeness. Example cases guide the evaluation. The prompt template can be seen in Figure ~\ref{fig:web-code-edit-eval-prompt}.

\begin{figure*}[!htbp]
\small
\centering
\begin{tcolorbox}[mypromptstyle, title={\textbf{\codeeditNoStyle{} Evaluation Prompt}}]
  
You are evaluating a system that generates HTML code based on a given task. Compare the predicted code with the ground truth code and rate its correctness based on functionality rather than exact syntax. If the code performs the intended task correctly, even if formatted differently or using a different approach, it should receive a high score.

\textbf{Rating Scale:}
\begin{itemize}
    \item 5 - PERFECT
    - Fully achieves the required functionality as described in the reference output.
    - May have differences in syntax or structure, but effectively performs the same task with no missing elements.
    \item 4 - CORRECT BUT WITH MINOR ISSUES
    - Achieves the intended functionality but has small flaws (e.g., slightly different behavior, minor inefficiencies).
    \item 3 - PARTIALLY CORRECT
    - Achieves part of the intended functionality but is missing key aspects or has notable issues.
    \item 2 - MOSTLY INCORRECT
    - Fails to accomplish most of the required functionality but shows some partial effort.
    \item 1 - COMPLETELY INCORRECT
    - The solution does not fulfill the required functionality or is entirely off-target.
\end{itemize}
    \textbf{Examples for reference:}
        \texttt{\\
        examples = [\\
        \hspace*{4mm} \{\\
        \hspace*{8mm} "INPUT": \{\\
        \hspace*{12mm} "question": "Change the header's background color to blue.",\\
        \hspace*{12mm} "model\_answer":"+<style>header\{background-color:blue;\}</style> <header>Welcome</header>",\\
        \hspace*{12mm} "ground\_truth": "<header style='background-color: blue;'>Welcome</header>"\},\\
        \hspace*{8mm} "OUTPUT": \{\\
        \hspace*{12mm} "rating": 5,\\
        \hspace*{12mm} "rationale": "The model answer correctly implements the change by ensuring the header displays with a blue background. Despite using a style tag in the model answer versus inline styling in the ground truth, both approaches deliver the exact intended functionality." \}\\
        \hspace*{4mm} \} ]\\
    }

    \textbf{Task for Evaluation:}
    \texttt{\\
    \hspace*{4mm} \{\\
    \hspace*{8mm} "INPUT": \{ \\
    \hspace*{12mm} "question": "<question>",\\
    \hspace*{12mm} "model\_answer": "<model\_answer>",\\
    \hspace*{12mm} "ground\_truth": "<ground\_truth>"\\
    \hspace*{8mm} \}\\
    \hspace*{4mm} \}\\
    }

Provide your evaluation in the following JSON format (using json delimiters, do not include any extra text):
\texttt{\\
\hspace*{4mm} \{\\
\hspace*{8mm} "rating": "1 or 2 or 3 or 4 or 5",\\
\hspace*{8mm} "rationale": "[Brief explanation of why this rating was chosen]"\\
\hspace*{4mm} \}
}
  \end{tcolorbox}
  \caption{LLM-as-judge prompt for \codeeditNoStyle{} task using few shot examples.}
  \label{fig:web-code-edit-eval-prompt}
\end{figure*}

\section{Case Studies of Model Performance}

\subsection{Case Studies for the \codeeditNoStyle{}}
\begin{figure*}[h]
  \centering
  \includegraphics[width=\linewidth]{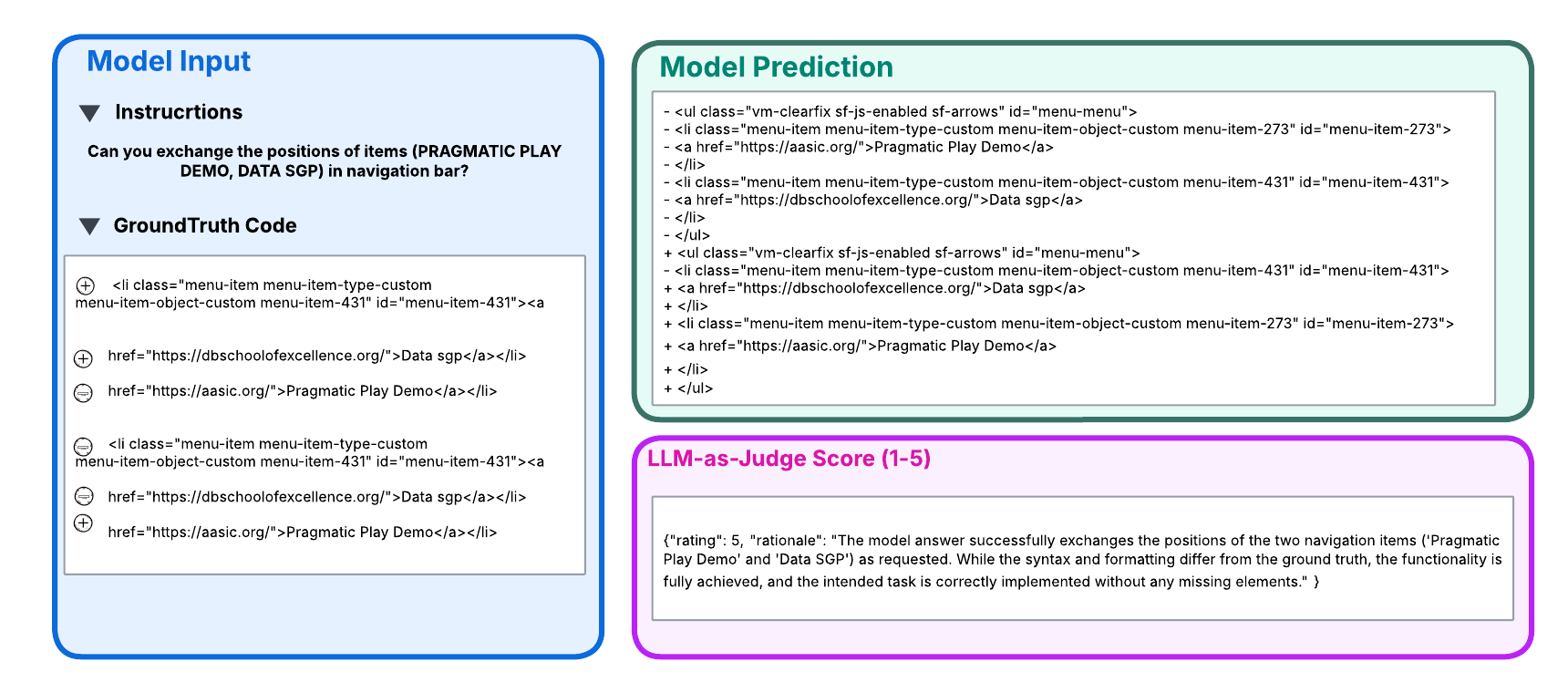}
  \caption{Code edit case study: \textbf{Success (Claude-3.5)}. The generated code accurately addresses the instructions given as input.}
  \label{fig:code_edit_success_claude}
\end{figure*}

\begin{figure*}[h]
  \centering
  \includegraphics[width=\linewidth]{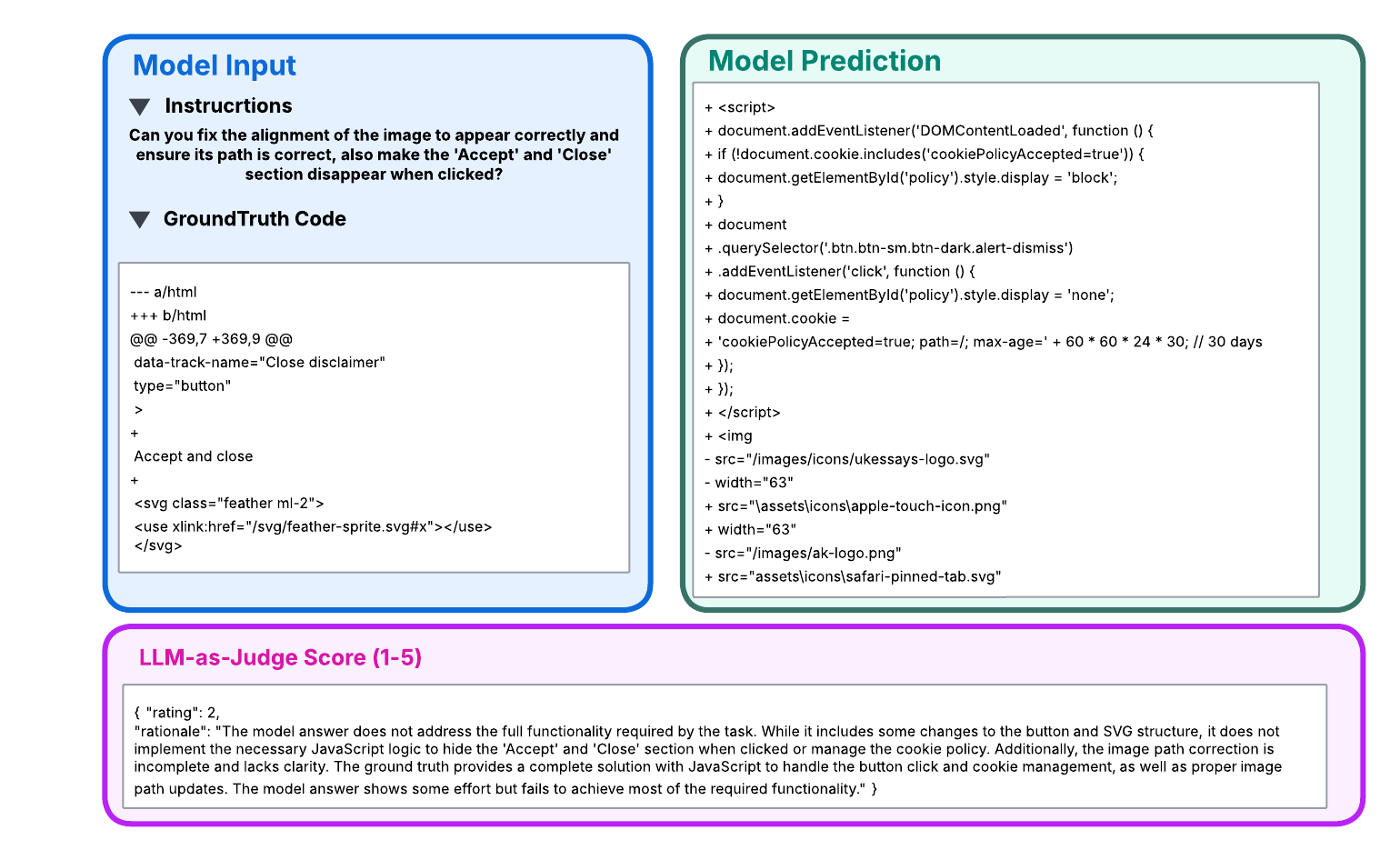}
  \caption{Code edit case study: \textbf{Failure (Gemini-2.0-Flash)}. The generated answer by the model skips main modifications requested by the user.}
  \label{fig:code_edit_failure_gemini}
\end{figure*}

We present case studies for the \codeeditNoStyle{} task illustrating both success and failure examples. Figure~\ref{fig:code_edit_success_claude} shows a success case where the Claude-3.5 model generates code that accurately follows the user’s instructions. In contrast, Figure~\ref{fig:code_edit_failure_gemini} highlights a failure case for the Gemini-2.0-Flash model, which overlooks key modifications requested by the user.


\begin{table*}[h]
  \centering
  \setlength{\tabcolsep}{6pt}  
  \renewcommand{\arraystretch}{1.1}  
  \resizebox{\textwidth}{!}{%
  \begin{tabular}{lcccccccccccc}
    \toprule
    \multirow{2}{*}{Model} & \multicolumn{3}{c}{English} & \multicolumn{3}{c}{French} & \multicolumn{3}{c}{German} & \multicolumn{3}{c}{Spanish} \\
    \cmidrule(lr){2-4} \cmidrule(lr){5-7} \cmidrule(lr){8-10} \cmidrule(lr){11-13}
     & \textbf{BLEU} & \textbf{TreeBLEU} & \textbf{LLM-as-Judge} & \textbf{BLEU} & \textbf{TreeBLEU} & \textbf{LLM-as-Judge} & \textbf{BLEU} & \textbf{TreeBLEU} & \textbf{LLM-as-Judge} & \textbf{BLEU} & \textbf{TreeBLEU} & \textbf{LLM-as-Judge} \\
    \midrule
    \rowcolor{open_models_below_8B!50} QwenVL-7B          & 7.89 & 22.1 & 2.2 & 5.34 & 15.92 & 1.95 & 5.15 & 19.54 & 1.94 & 5.07 & 17.65 & 1.96 \\
    \rowcolor{open_models_below_8B!50} Molmo-7B           & 1.25 & 5.99 & 1.6 & 2.0 & 8.98 & 1.55 & 1.14 & 8.97 & 1.39 & 2.03 & 6.38 & 1.46 \\
    \rowcolor{open_models_below_8B!50} Phi-3.5-VI         & 0.01 & 0.0 & 1.01 & 0.0 & 0.0 & 1.01 & 0.03 & 0.0 & 1.0 & 0.03 & 0.0 & 1.0 \\
    \rowcolor{open_models_below_8B!50} Llava-OV-7B         & 7.78 & 20.34 & 2.19 & 3.66 & 13.6 & 1.93 & 5.14 & 19.26 & 1.65 & 3.54 & 14.8 & 1.8 \\
    \rowcolor{open_models_below_12B!50} Fuyu-8B            & 0.01 & 0.14 & 1.05 & 0.01 & 0.14 & 1.0 & 0.0 & 0.04 & 1.0 & 0.03 & 0.0 & 1.02 \\
    \rowcolor{open_models_below_12B!50} InternVL-2.5-8B & 8.21 & 17.02 & 2.04 & 5.18 & 14.64 & 1.72 & 6.02 & 19.3 & 1.85 & 5.62 & 14.03 & 1.9 \\
    \rowcolor{open_models_below_12B!50} Glm-4v-9B         & 4.65 & 13.9 & 2.34 & 4.25 & 18.03 & 2.31 & 3.09 & 12.82 & 2.07 & 3.83 & 11.67 & 2.18 \\
    \rowcolor{open_models_below_12B!50} Pixtral-12B        & 11.88 & 21.78 & 2.86 & 5.88 & 12.8 & 2.66 & 6.5 & 18.74 & 2.33 & 6.32 & 13.33 & 2.72 \\
    \rowcolor{open_models_over_12B!50} InternVL-2.5-38B & 13.71 & 26.86 & 3.85 & 9.51 & \textbf{24.75} & 3.61 & 11.29 & 27.46 & 3.54 & 8.95 & \textbf{23.08} & 3.62 \\
    \rowcolor{open_models_over_12B!50} QwenVL-72B      & 12.8 & 26.47 & 3.37 & 11.08 & 22.88 & 3.22 & 10.23 & 30.27 & 3.05 & 9.17 & 18.22 & 2.99 \\
    \midrule
    \rowcolor{closed_models!50} Claude 3.5 Sonnet               & 15.14 & 24.4 & \textbf{4.62} & \textbf{16.25} & 19.3 & \textbf{4.39} & \textbf{17.16} & \textbf{34.6} & \textbf{4.45} & \textbf{13.53} & 16.21 & \textbf{4.47} \\
    \rowcolor{closed_models!50} Gemini-2.0-Flash        & \textbf{17.02} & \textbf{28.54} & 3.26 & 11.34 & 13.32 & 3.45 & 11.98 & 23.08 & 3.43 & 11.09 & 17.46 & 3.79 \\
    \rowcolor{closed_models!50} GPT-4o (1120)             & 13.95 & 22.94 & 4.26 & 10.32 & 9.93 & 4.35 & 12.87 & 22.63 & 4.28 & 11.15 & 12.81 & 4.39 \\
    \bottomrule
  \end{tabular}%
  }
\caption{Results of \codeeditNoStyle{} on different languages.}
\label{tab:codeedit}
\end{table*}

\subsection{Case Studies for the \mockupNoStyle{}}
\begin{figure*}[h]
  \begin{subfigure}{\textwidth}
    \begin{tcolorbox}[
      colback=white,
      colframe=black!5,
      arc=2pt,
      boxrule=0.5pt,
      width=\textwidth,
      top=1pt,
      bottom=1pt
    ]
    
    \begin{tcolorbox}[
      colback=lightblue,
      colframe=lightblue,
      width=1.02\textwidth,
      boxsep=2pt,
      top=1pt,
      bottom=1pt,
      left=4pt,
      right=4pt,
      enlarge left by=-0.005\textwidth,
      enlarge right by=-0.005\textwidth
    ]
      \textcolor{ecologygreen}{\textbf{\oOne}}
    \end{tcolorbox}
    
    \begin{minipage}{\textwidth}
      \begin{minipage}{\textwidth}
        \begin{minipage}{0.31\textwidth}
          \centering
          \parbox[c][4.3cm][t]{0.98\textwidth}{
            \centering
            \vspace{0.1cm}
            \includegraphics[height=3.7cm]{Figures/mockup_o1_failure_original.png}
            \vspace{0.1cm}
          }
          \centering\textbf{\footnotesize Original Page}
        \end{minipage}
        \begin{minipage}{0.31\textwidth}
          \centering
          \parbox[c][4.3cm][t]{0.98\textwidth}{
            \centering
            \vspace{0.1cm}
            \includegraphics[height=3.7cm]{Figures/mockup_o1_failure_sketch.png}
            \vspace{0.1cm}
          }
          \centering\textbf{\footnotesize Mockup Image}
        \end{minipage}
        \begin{minipage}{0.31\textwidth}
          \centering
          \parbox[c][4.3cm][t]{0.98\textwidth}{
            \centering
            \vspace{0.1cm}
            \includegraphics[height=3cm]{Figures/mockup_o1_failure_generated.png}
            \vspace{0.1cm}
          }
          \centering\textbf{\footnotesize Generation}
        \end{minipage}
      \end{minipage}

      \vspace{0.05cm}
      
      \begin{tcolorbox}[
        colback=codebg,
        colframe=gray!15,
        width=1.02\textwidth,
        boxsep=2pt,
        top=1pt,
        bottom=1pt,
        left=3pt,
        right=3pt,
        enlarge left by=-0.005\textwidth,
        enlarge right by=-0.005\textwidth,
        title={\textcolor{black}{\textbf{GPT Score Evaluation}}},
        fonttitle=\footnotesize
      ]
        \begin{lstlisting}[basicstyle=\scriptsize\ttfamily, escapeinside={(*}{*)}]
(*\textbf{Alignment: 2.}*) Key elements (e.g., input box) misaligned, deviating from intended grid.
(*\textbf{Layout: 2.}*) Two-column structure poorly represented, essential sections missing/merged.
(*\textbf{Spacing: 2.}*) Uneven element distribution results in inconsistent spacing and imbalance.
(*\textbf{Overall Score: 2}*)
        \end{lstlisting}
      \end{tcolorbox}
    \end{minipage}
    
    \end{tcolorbox}
  \end{subfigure}
  
  \vspace{0.1cm}
  
  \begin{subfigure}{\textwidth}
    \begin{tcolorbox}[
      colback=white,
      colframe=black!5,
      arc=2pt,
      boxrule=0.5pt,
      width=\textwidth,
      top=1pt,
      bottom=1pt
    ]
    
    \begin{tcolorbox}[
      colback=lightblue,
      colframe=lightblue,
      width=1.02\textwidth,
      boxsep=2pt,
      top=1pt,
      bottom=1pt,
      left=4pt,
      right=4pt,
      enlarge left by=-0.005\textwidth,
      enlarge right by=-0.005\textwidth
    ]
      \textcolor{ecologygreen}{\textbf{InternVL-38B}}
    \end{tcolorbox}
    
    \begin{minipage}{\textwidth}
      \begin{minipage}{\textwidth}
        \begin{minipage}{0.31\textwidth}
          \centering
          \parbox[c][4.1cm][t]{0.98\textwidth}{
            \centering
            \vspace{0.1cm}
            \includegraphics[height=3.9cm]{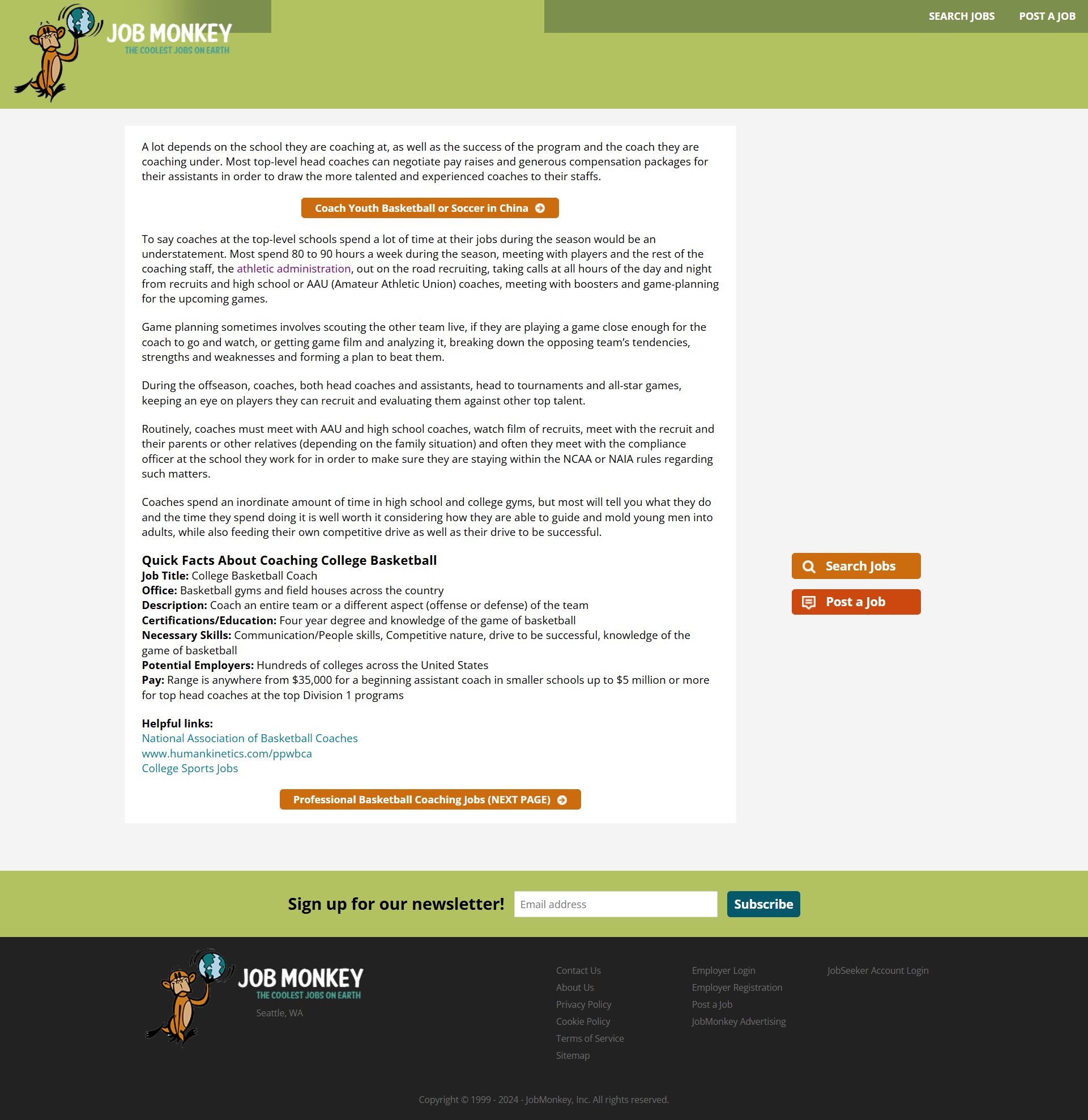}
          }
          \vspace{0.05cm}
          \centering\textbf{\footnotesize Original Page}
        \end{minipage}
        \begin{minipage}{0.31\textwidth}
          \centering
          \parbox[c][4.1cm][t]{0.98\textwidth}{
            \centering
            \vspace{0.1cm}
            \includegraphics[height=3.9cm]{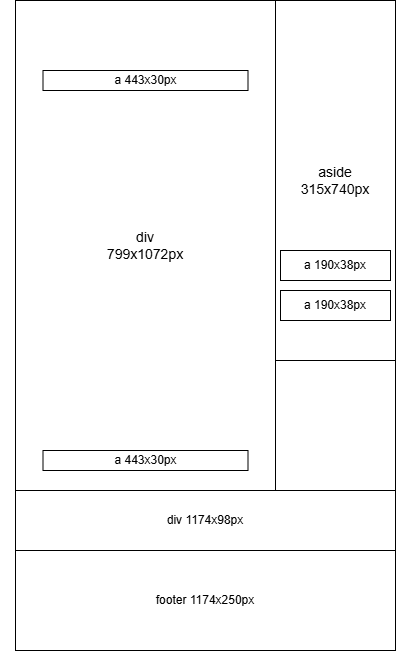}
          }
        \vspace{0.05cm}
          \centering\textbf{\footnotesize Mockup Image}
        \end{minipage}
        \begin{minipage}{0.31\textwidth}
          \centering
          \parbox[c][4.1cm][t]{0.98\textwidth}{
            \centering
            \vspace{0.1cm}
            \includegraphics[height=2.7cm]{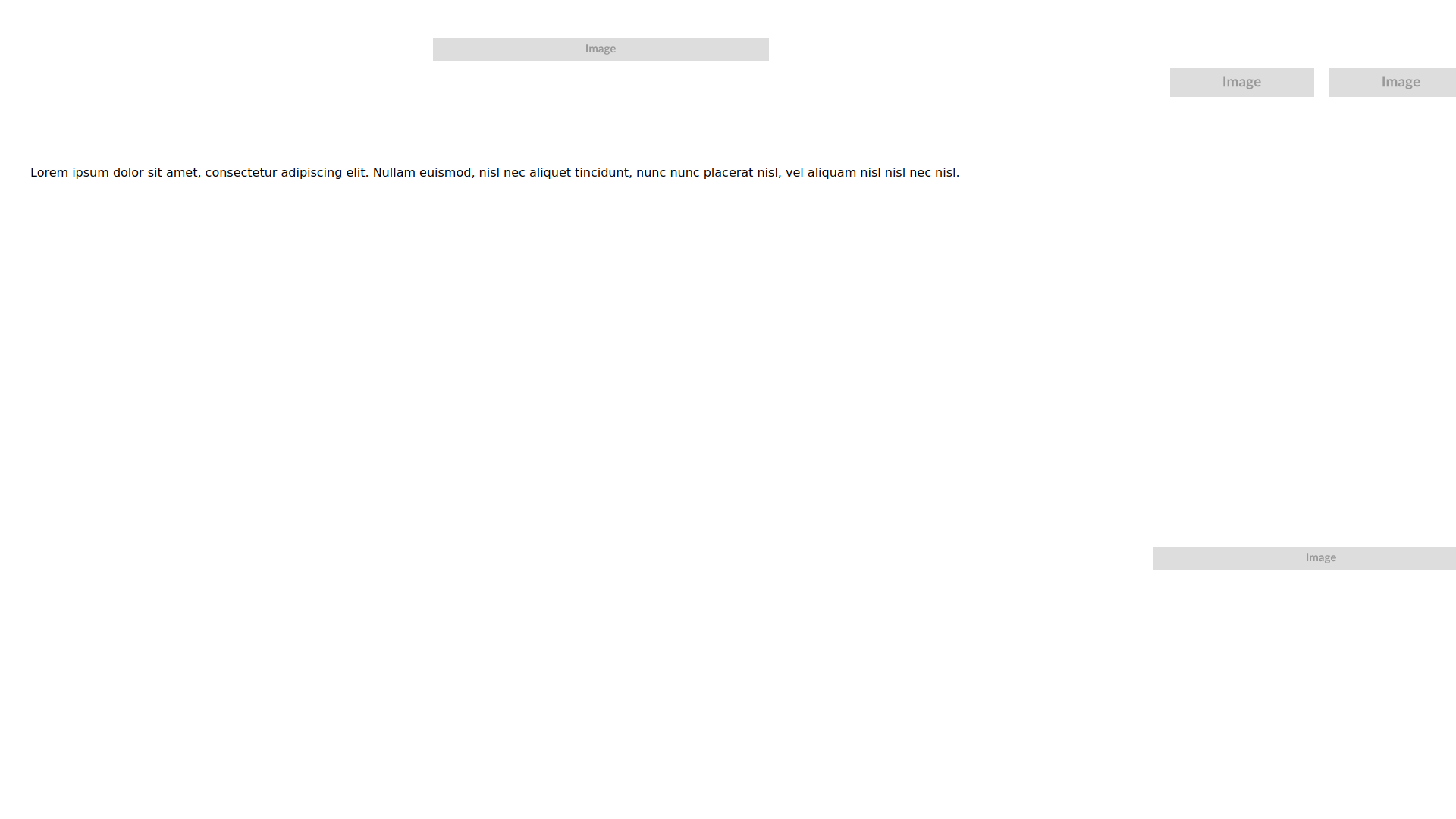}
          }
        \vspace{0.05cm}
          \centering\textbf{\footnotesize Generation}
        \end{minipage}
      \end{minipage}

      \vspace{0.05cm}
      
      \begin{tcolorbox}[
        colback=codebg,
        colframe=gray!15,
        width=1.02\textwidth,
        boxsep=2pt,
        top=1pt,
        bottom=1pt,
        left=3pt,
        right=3pt,
        enlarge left by=-0.005\textwidth,
        enlarge right by=-0.005\textwidth,
        title={\textcolor{black}{\textbf{GPT Score Evaluation}}},
        fonttitle=\footnotesize
      ]
        \begin{lstlisting}[basicstyle=\scriptsize\ttfamily, escapeinside={(*}{*)}]
(*\textbf{Alignment: 2.}*) The main content is misaligned; also, key elements like the aside and buttons are missing.
(*\textbf{Layout: 2.}*) The intended two-column structure is replaced by a single-column layout, significantly deviating from the sketch.
(*\textbf{Spacing: 2.}*) Inconsistent spacing and proportions arise from the absence of the aside and smaller placeholders.
(*\textbf{Overall Score: 2}*)
        \end{lstlisting}
      \end{tcolorbox}
    \end{minipage}
    
    \end{tcolorbox}
  \end{subfigure}
  
  \caption{Examples of the \textbf{failure cases on the \mockupNoStyle{} task} for the best closed-source model (\oOne) and the best open-source model (InternVL2.5-38B).}
  \label{fig:mockup2code_failure_case_o1_and_internvl}
\end{figure*}

\begin{figure*}[h]
  \begin{subfigure}{\textwidth}
    \begin{tcolorbox}[
      colback=white,
      colframe=black!5,
      arc=2pt,
      boxrule=0.5pt,
      width=\textwidth,
      top=1pt,
      bottom=1pt
    ]
    
    \begin{tcolorbox}[
      colback=lightblue,
      colframe=lightblue,
      width=1.02\textwidth,
      boxsep=2pt,
      top=1pt,
      bottom=1pt,
      left=4pt,
      right=4pt,
      enlarge left by=-0.005\textwidth,
      enlarge right by=-0.005\textwidth
    ]
      \textcolor{ecologygreen}{\textbf{\oOne}}
    \end{tcolorbox}
    
    \begin{minipage}{\textwidth}
      \begin{minipage}{\textwidth}
        \begin{minipage}{0.31\textwidth}
          \centering
          \parbox[c][4.3cm][t]{0.98\textwidth}{
            \centering
            \vspace{0.1cm}
            \includegraphics[height=3.7cm]{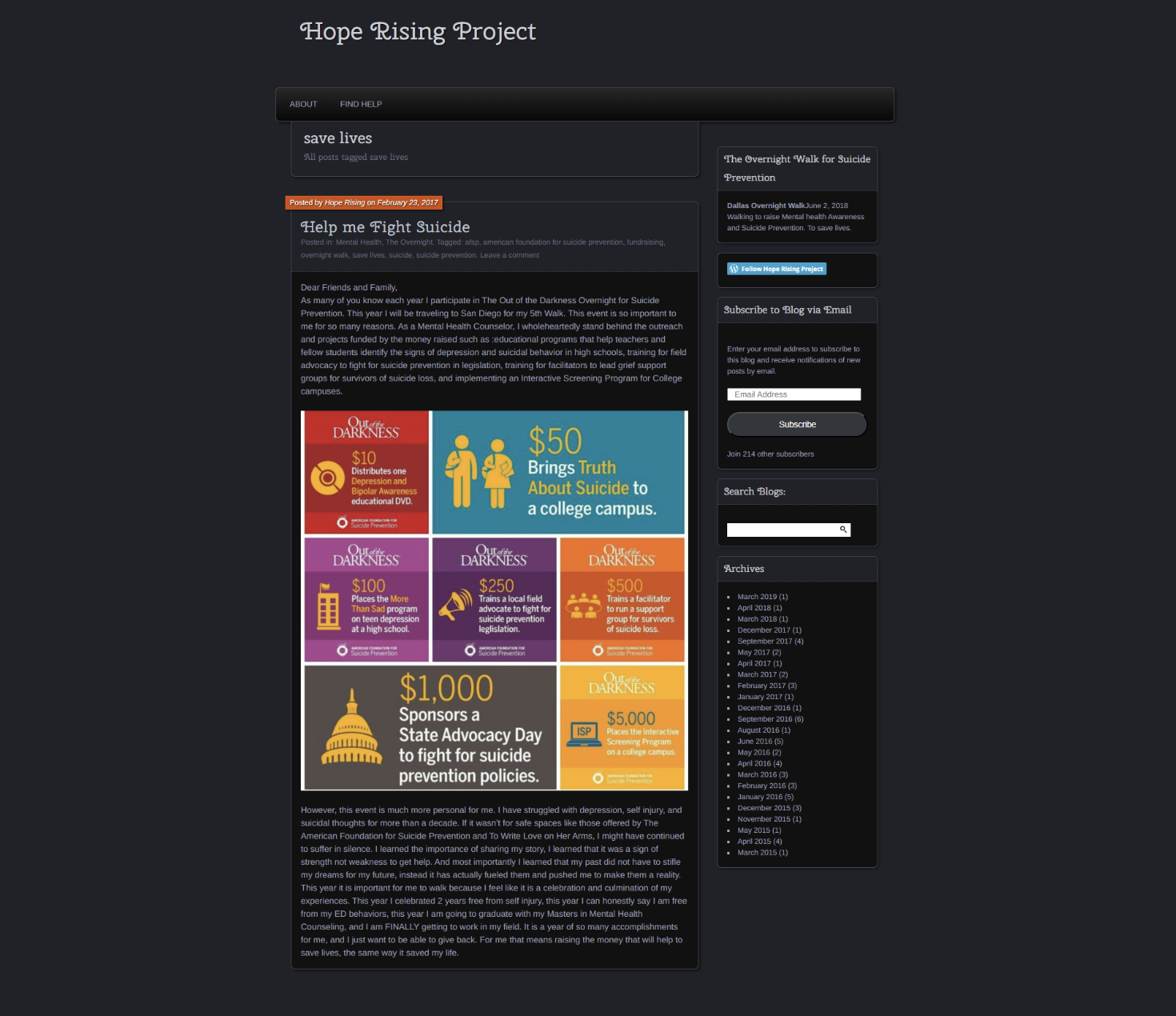}
            \vspace{0.1cm}
          }
          \centering\textbf{\footnotesize Original Page}
        \end{minipage}
        \begin{minipage}{0.31\textwidth}
          \centering
          \parbox[c][4.3cm][t]{0.98\textwidth}{
            \centering
            \vspace{0.1cm}
            \includegraphics[height=3.7cm]{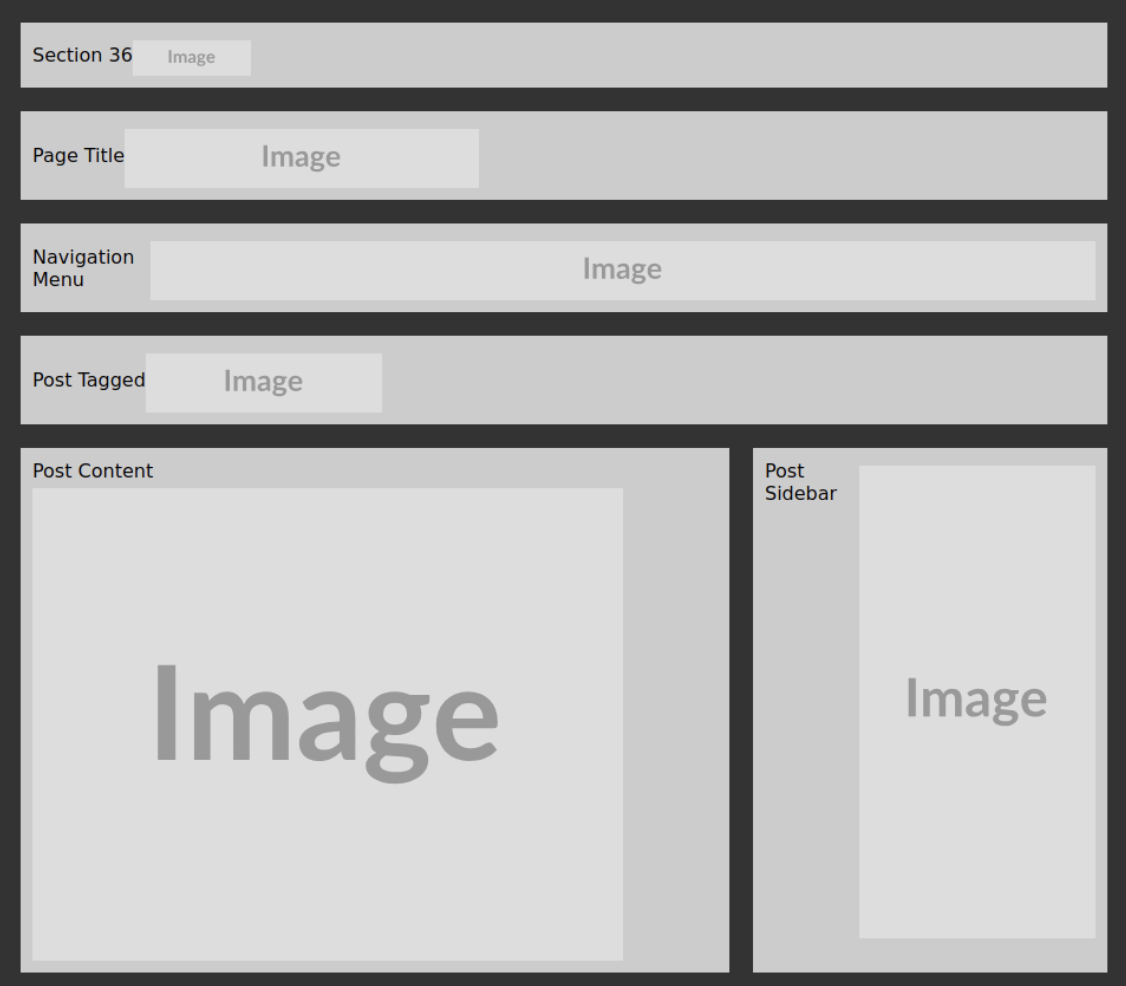}
            \vspace{0.1cm}
          }
          \centering\textbf{\footnotesize Mockup Image}
        \end{minipage}
        \begin{minipage}{0.31\textwidth}
          \centering
          \parbox[c][4.3cm][t]{0.98\textwidth}{
            \centering
            \vspace{0.1cm}
            \includegraphics[height=3cm]{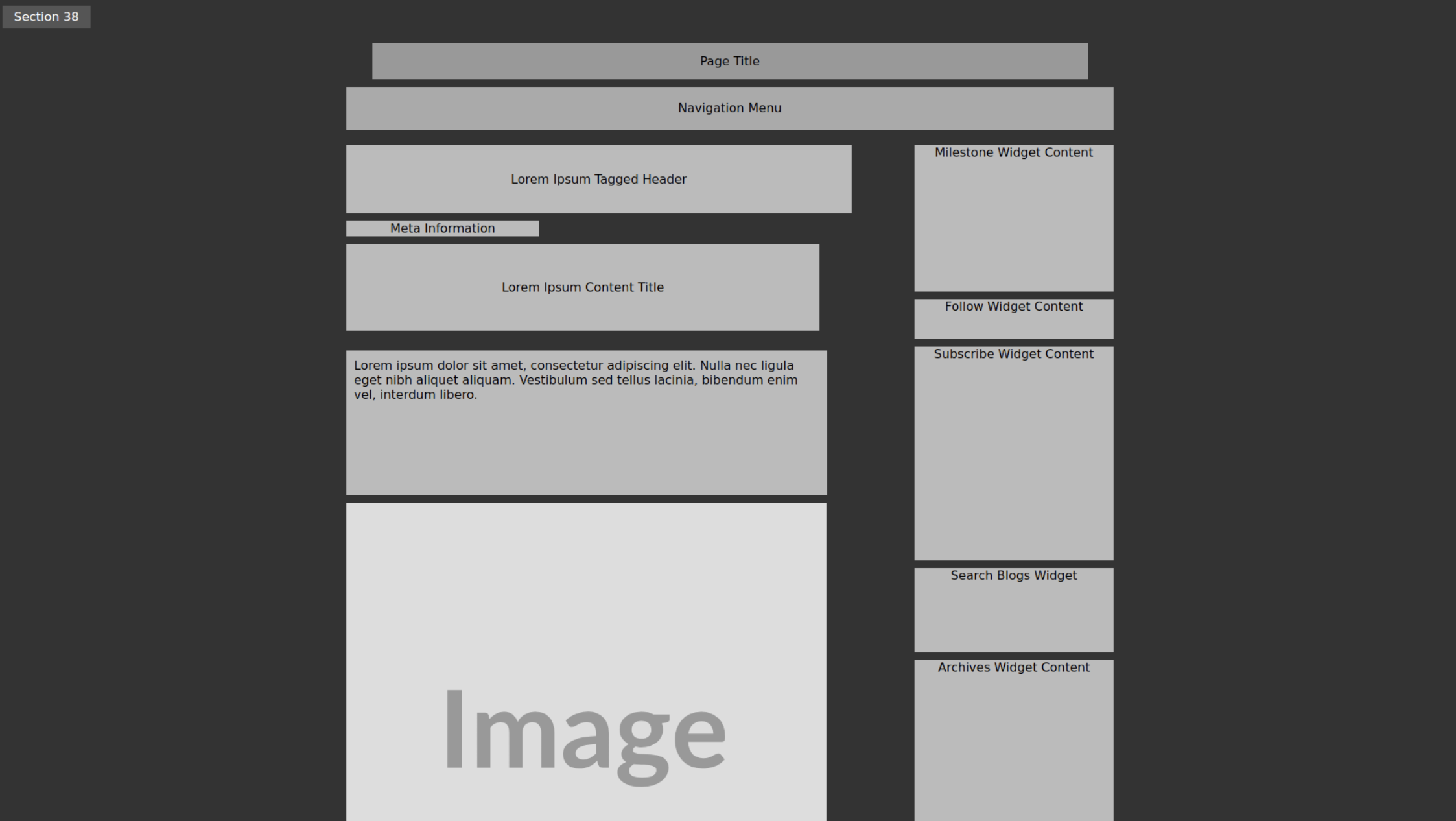}
            \vspace{0.1cm}
          }
          \centering\textbf{\footnotesize Generation}
        \end{minipage}
      \end{minipage}

      \vspace{0.05cm}
      
      \begin{tcolorbox}[
        colback=codebg,
        colframe=gray!15,
        width=1.02\textwidth,
        boxsep=2pt,
        top=1pt,
        bottom=1pt,
        left=3pt,
        right=3pt,
        enlarge left by=-0.005\textwidth,
        enlarge right by=-0.005\textwidth,
        title={\textcolor{black}{\textbf{GPT Score Evaluation on Simple sketch}}},
        fonttitle=\footnotesize
      ]
\begin{lstlisting}[basicstyle=\scriptsize\ttfamily, escapeinside={(*}{*)}]
(*\textbf{Layout: 5.}*) Structure perfectly replicated.
(*\textbf{Spacing: 4.}*) Minor inconsistencies in spacing.
(*\textbf{Alignment: 5.}*) Precise alignment maintained.
(*\textbf{Overall Score: 4.67}*)
\end{lstlisting}
      \end{tcolorbox}
    \end{minipage}
    
    \end{tcolorbox}
  \end{subfigure}
  
  \vspace{0.1cm}
  
  \begin{subfigure}{\textwidth}
    \begin{tcolorbox}[
      colback=white,
      colframe=black!5,
      arc=2pt,
      boxrule=0.5pt,
      width=\textwidth,
      top=1pt,
      bottom=1pt
    ]
    
    \begin{tcolorbox}[
      colback=lightblue,
      colframe=lightblue,
      width=1.02\textwidth,
      boxsep=2pt,
      top=1pt,
      bottom=1pt,
      left=4pt,
      right=4pt,
      enlarge left by=-0.005\textwidth,
      enlarge right by=-0.005\textwidth
    ]
      \textcolor{ecologygreen}{\textbf{OpenAI-o1}}
    \end{tcolorbox}
    
    \begin{minipage}{\textwidth}
      \begin{minipage}{\textwidth}
        \begin{minipage}{0.31\textwidth}
          \centering
          \parbox[c][4.1cm][t]{0.98\textwidth}{
            \centering
            \vspace{0.1cm}
            \includegraphics[height=3.9cm]{Figures-appendix/1-original.png}
          }
          \vspace{0.05cm}
          \centering\textbf{\footnotesize Original Page}
        \end{minipage}
        \begin{minipage}{0.31\textwidth}
          \centering
          \parbox[c][4.1cm][t]{0.98\textwidth}{
            \centering
            \vspace{0.1cm}
            \includegraphics[height=3.9cm]{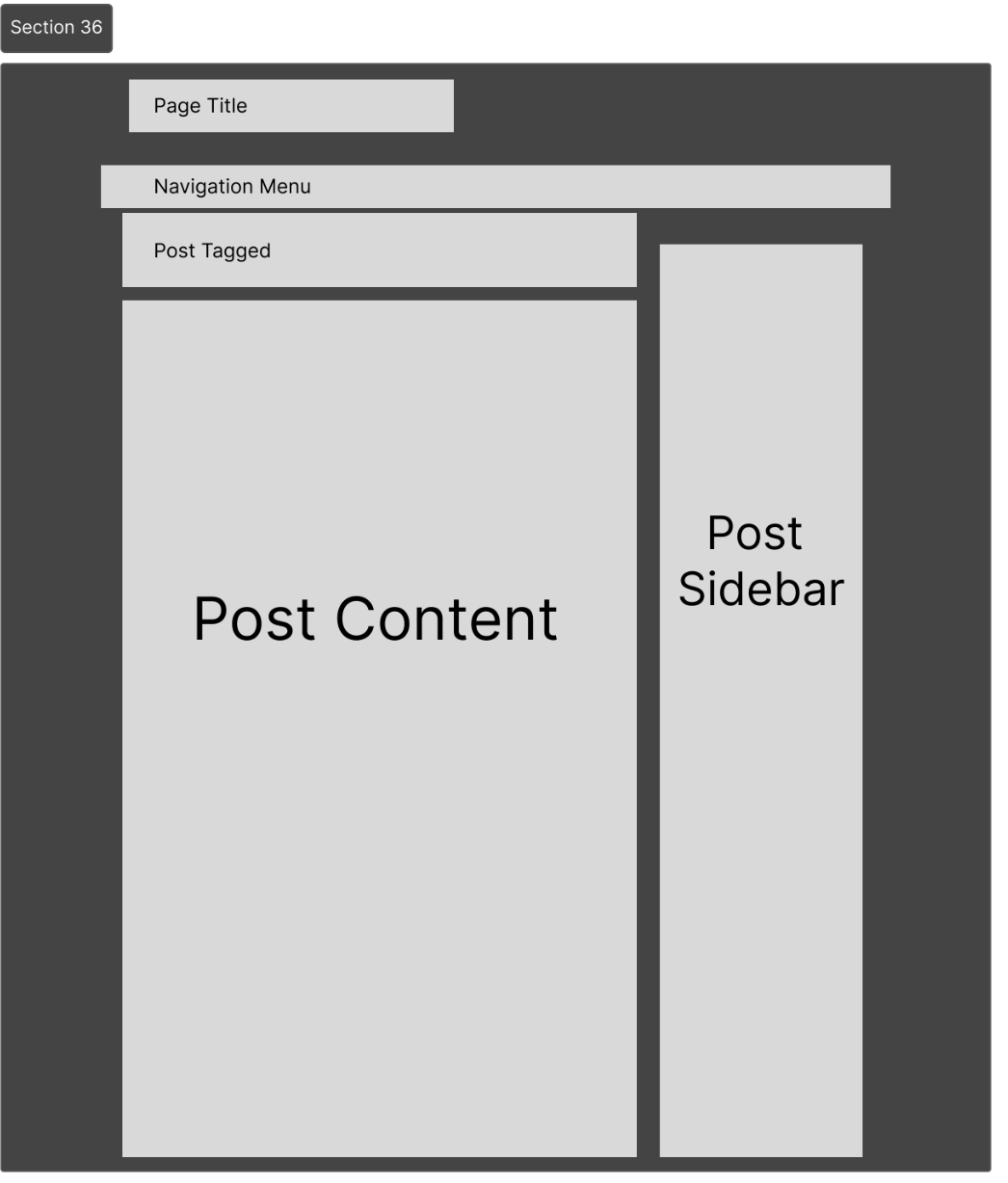}
          }
        \vspace{0.05cm}
          \centering\textbf{\footnotesize Mockup Image}
        \end{minipage}
        \begin{minipage}{0.31\textwidth}
          \centering
          \parbox[c][4.1cm][t]{0.98\textwidth}{
            \centering
            \vspace{0.1cm}
            \includegraphics[height=2.7cm]{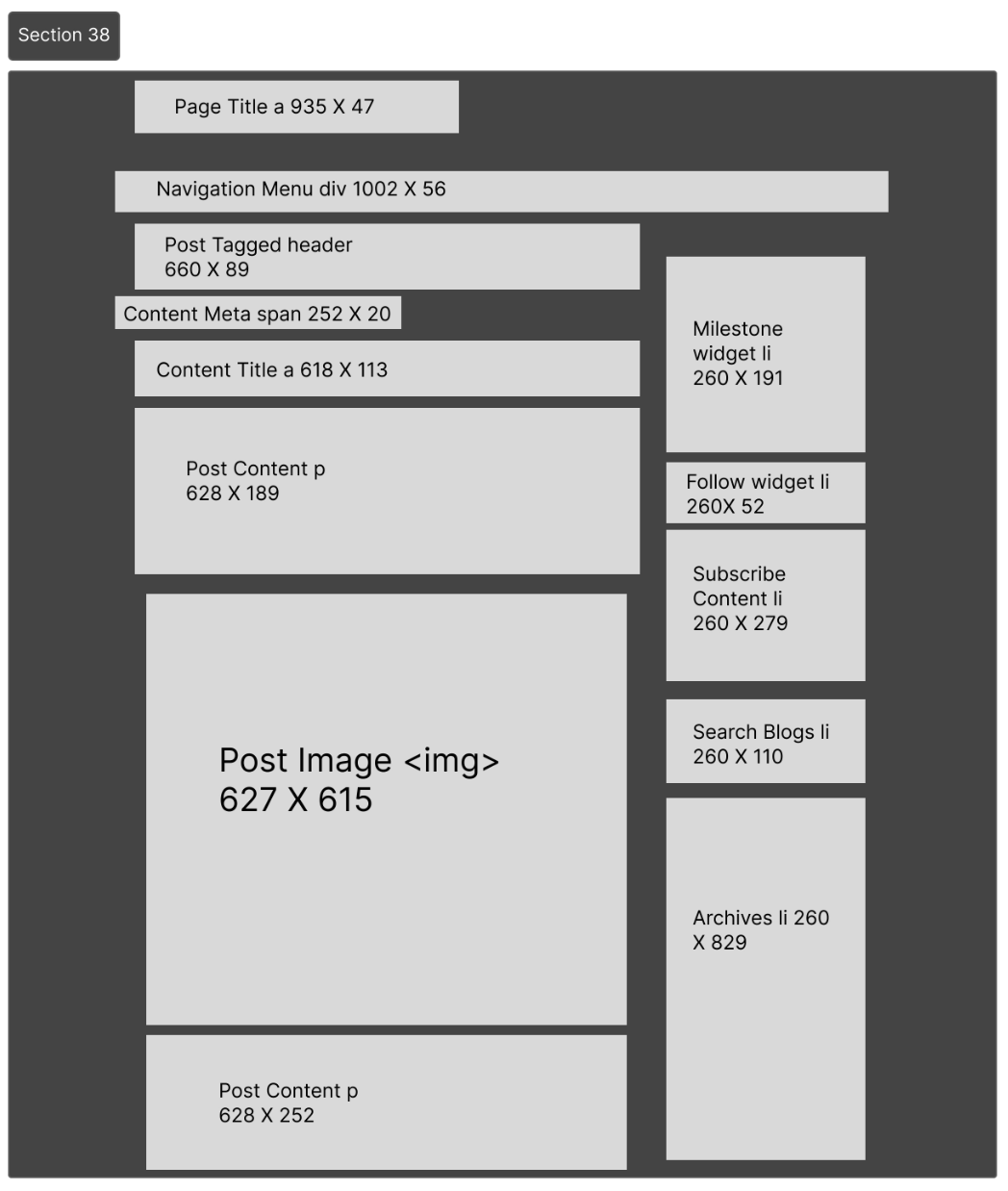}
          }
        \vspace{0.05cm}
          \centering\textbf{\footnotesize Generation}
        \end{minipage}
      \end{minipage}

      \vspace{0.05cm}
      
      \begin{tcolorbox}[
        colback=codebg,
        colframe=gray!15,
        width=1.02\textwidth,
        boxsep=2pt,
        top=1pt,
        bottom=1pt,
        left=3pt,
        right=3pt,
        enlarge left by=-0.005\textwidth,
        enlarge right by=-0.005\textwidth,
        title={\textcolor{black}{\textbf{GPT Score Evaluation on Complex Sketch}}},
        fonttitle=\footnotesize
      ]
\begin{lstlisting}[basicstyle=\scriptsize\ttfamily, escapeinside={(*}{*)}]
(*\textbf{Layout: 5.}*) Structure and sections perfectly replicated.
(*\textbf{Spacing: 5.}*) Spacing and proportions are consistent.
(*\textbf{Alignment: 5.}*) Alignment is precise and matches the sketch.
(*\textbf{Overall Score: 5}*)
\end{lstlisting}

      \end{tcolorbox}
    \end{minipage}
    
    \end{tcolorbox}
  \end{subfigure}
  
  \caption{Examples of the \textbf{success cases on the \mockupNoStyle{} task} for the best closed-source model (\oOne) for both simple and complex mockups.}
  \label{fig:mockup2code_success_case_o1_1}
\end{figure*}

\begin{figure*}[h]
  \begin{subfigure}{\textwidth}
    \begin{tcolorbox}[
      colback=white,
      colframe=black!5,
      arc=2pt,
      boxrule=0.5pt,
      width=\textwidth,
      top=1pt,
      bottom=1pt
    ]
    
    \begin{tcolorbox}[
      colback=lightblue,
      colframe=lightblue,
      width=1.02\textwidth,
      boxsep=2pt,
      top=1pt,
      bottom=1pt,
      left=4pt,
      right=4pt,
      enlarge left by=-0.005\textwidth,
      enlarge right by=-0.005\textwidth
    ]
      \textcolor{ecologygreen}{\textbf{\InternVLTwoFiveEightB}}
    \end{tcolorbox}
    
    \begin{minipage}{\textwidth}
      \begin{minipage}{\textwidth}
        \begin{minipage}{0.31\textwidth}
          \centering
          \parbox[c][4.3cm][t]{0.98\textwidth}{
            \centering
            \vspace{0.1cm}
            \includegraphics[height=3.7cm]{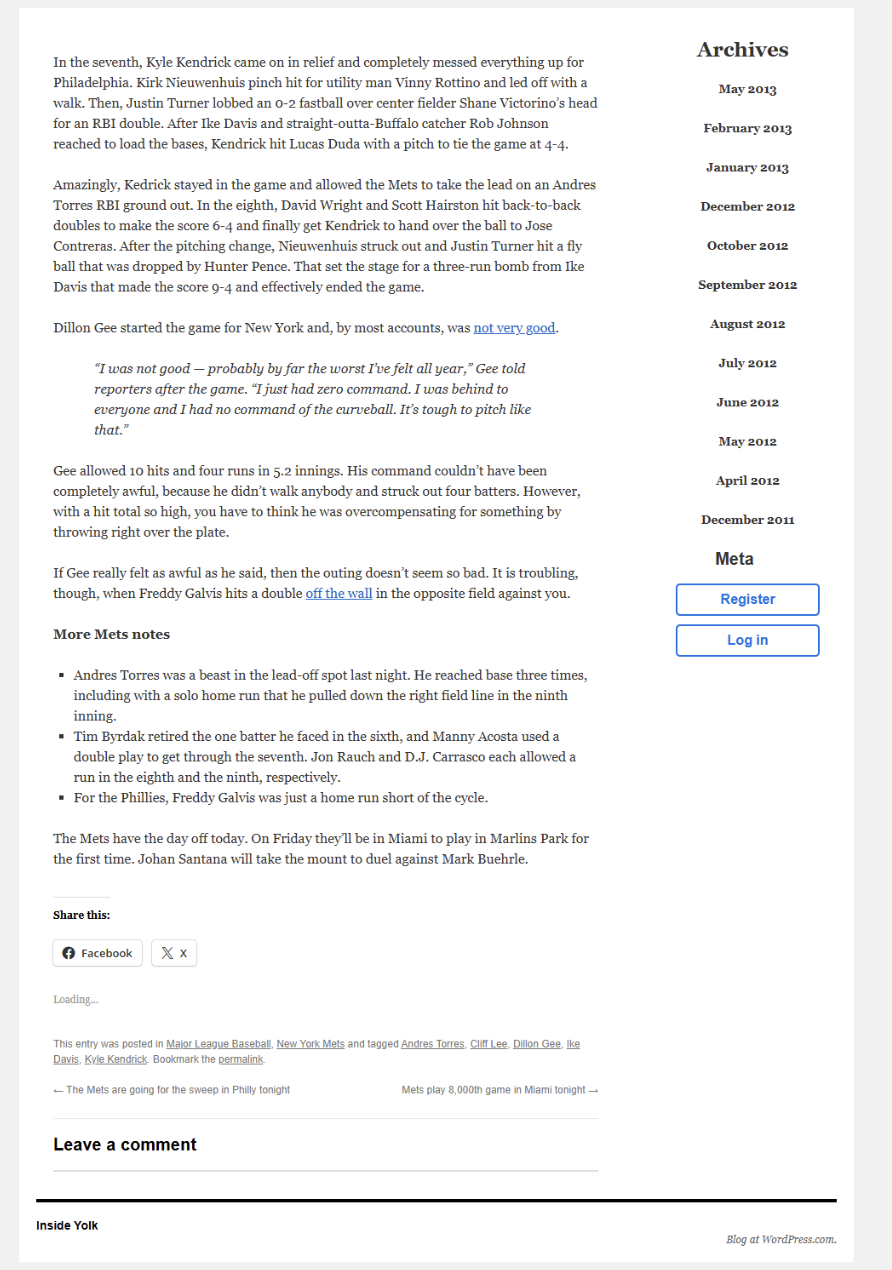}
            \vspace{0.1cm}
          }
          \centering\textbf{\footnotesize Original Page}
        \end{minipage}
        \begin{minipage}{0.31\textwidth}
          \centering
          \parbox[c][4.3cm][t]{0.98\textwidth}{
            \centering
            \vspace{0.1cm}
            \includegraphics[width=2.4cm,height=3.7cm]{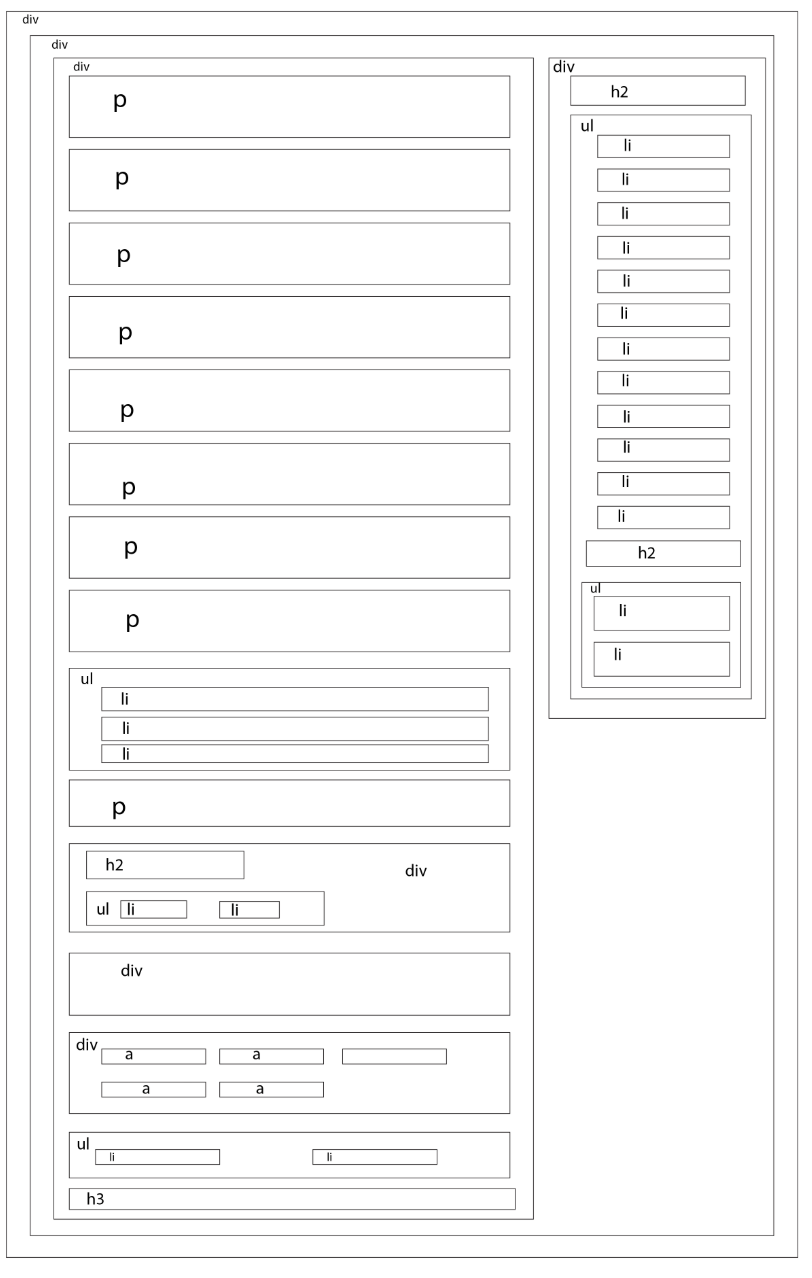}
            \vspace{0.1cm}
          }
          \centering\textbf{\footnotesize Mockup Image}
        \end{minipage}
        \begin{minipage}{0.31\textwidth}
          \centering
          \parbox[c][4.3cm][t]{0.98\textwidth}{
            \centering
            \vspace{0.1cm}
            \includegraphics[width = 2cm ,height=3cm]{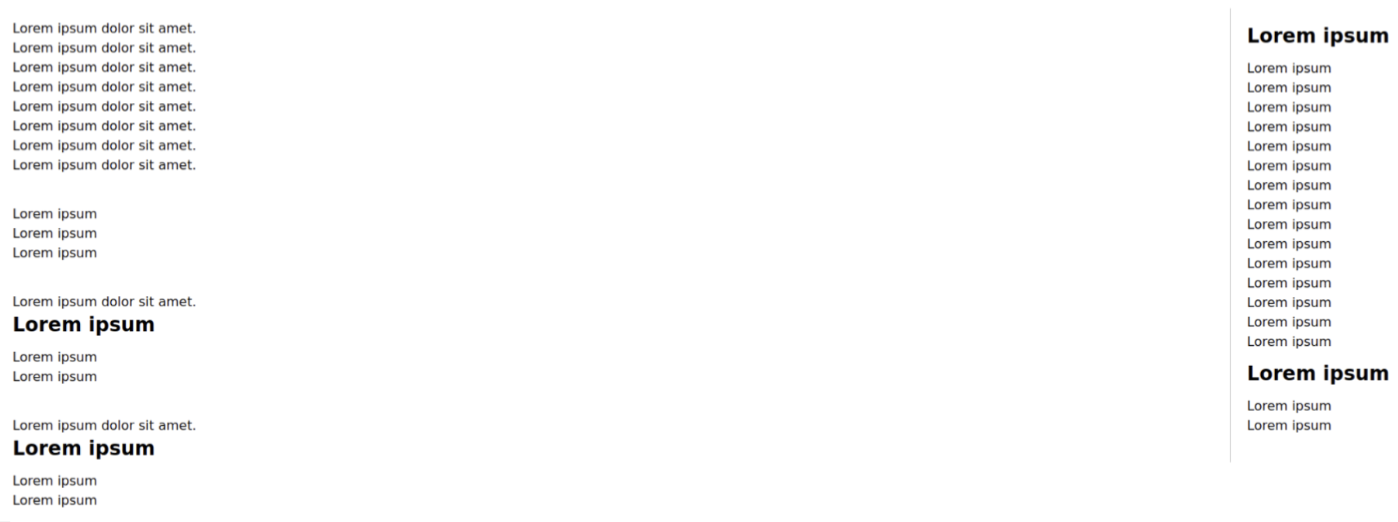}
            \vspace{0.1cm}
          }
          \centering\textbf{\footnotesize Generation}
        \end{minipage}
      \end{minipage}

      \vspace{0.05cm}
      
      \begin{tcolorbox}[
        colback=codebg,
        colframe=gray!15,
        width=1.02\textwidth,
        boxsep=2pt,
        top=1pt,
        bottom=1pt,
        left=3pt,
        right=3pt,
        enlarge left by=-0.005\textwidth,
        enlarge right by=-0.005\textwidth,
        title={\textcolor{black}{\textbf{GPT Score Evaluation on Simple Sketch}}},
        fonttitle=\footnotesize
      ]
\begin{lstlisting}[basicstyle=\scriptsize\ttfamily, escapeinside={(*}{*)}]
(*\textbf{Layout: 2.}*) Many key sections, headers, and lists are missing, leading to a poor structural match with the sketch.
(*\textbf{Spacing: 2.}*) Uneven gaps and inconsistent spacing cause improper grouping and distorted proportions.
(*\textbf{Alignment: 2.}*) Misaligned elements and an irregular grid result in a disorganized layout.
(*\textbf{Overall Score: 2}*)
\end{lstlisting}

      \end{tcolorbox}
    \end{minipage}
    
    \end{tcolorbox}
  \end{subfigure}
  
  \vspace{0.1cm}
  
  \begin{subfigure}{\textwidth}
    \begin{tcolorbox}[
      colback=white,
      colframe=black!5,
      arc=2pt,
      boxrule=0.5pt,
      width=\textwidth,
      top=1pt,
      bottom=1pt
    ]
    
    \begin{tcolorbox}[
      colback=lightblue,
      colframe=lightblue,
      width=1.02\textwidth,
      boxsep=2pt,
      top=1pt,
      bottom=1pt,
      left=4pt,
      right=4pt,
      enlarge left by=-0.005\textwidth,
      enlarge right by=-0.005\textwidth
    ]
      \textcolor{ecologygreen}{\textbf{\InternVLTwoFiveEightB}}
    \end{tcolorbox}
    
    \begin{minipage}{\textwidth}
      \begin{minipage}{\textwidth}
        \begin{minipage}{0.31\textwidth}
          \centering
          \parbox[c][4.1cm][t]{0.98\textwidth}{
            \centering
            \vspace{0.1cm}
            \includegraphics[height=3.9cm]{Figures-appendix/2-original.png}
          }
          \vspace{0.05cm}
          \centering\textbf{\footnotesize Original Page}
        \end{minipage}
        \begin{minipage}{0.31\textwidth}
          \centering
          \parbox[c][4.1cm][t]{0.98\textwidth}{
            \centering
            \vspace{0.1cm}
            \includegraphics[height=3.9cm]{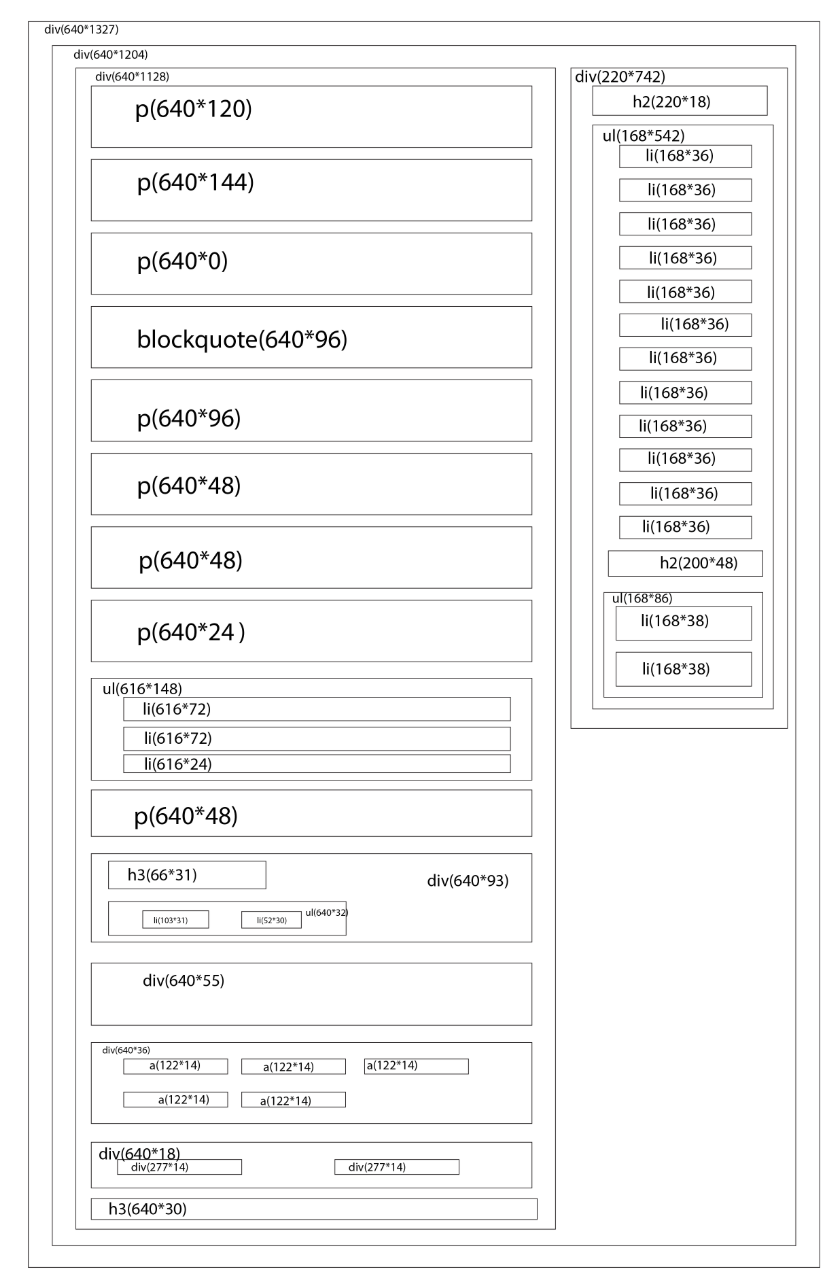}
          }
        \vspace{0.05cm}
          \centering\textbf{\footnotesize Mockup Image}
        \end{minipage}
        \begin{minipage}{0.31\textwidth}
          \centering
          \parbox[c][4.1cm][t]{0.98\textwidth}{
            \centering
            \vspace{0.1cm}
            \includegraphics[height=2.7cm]{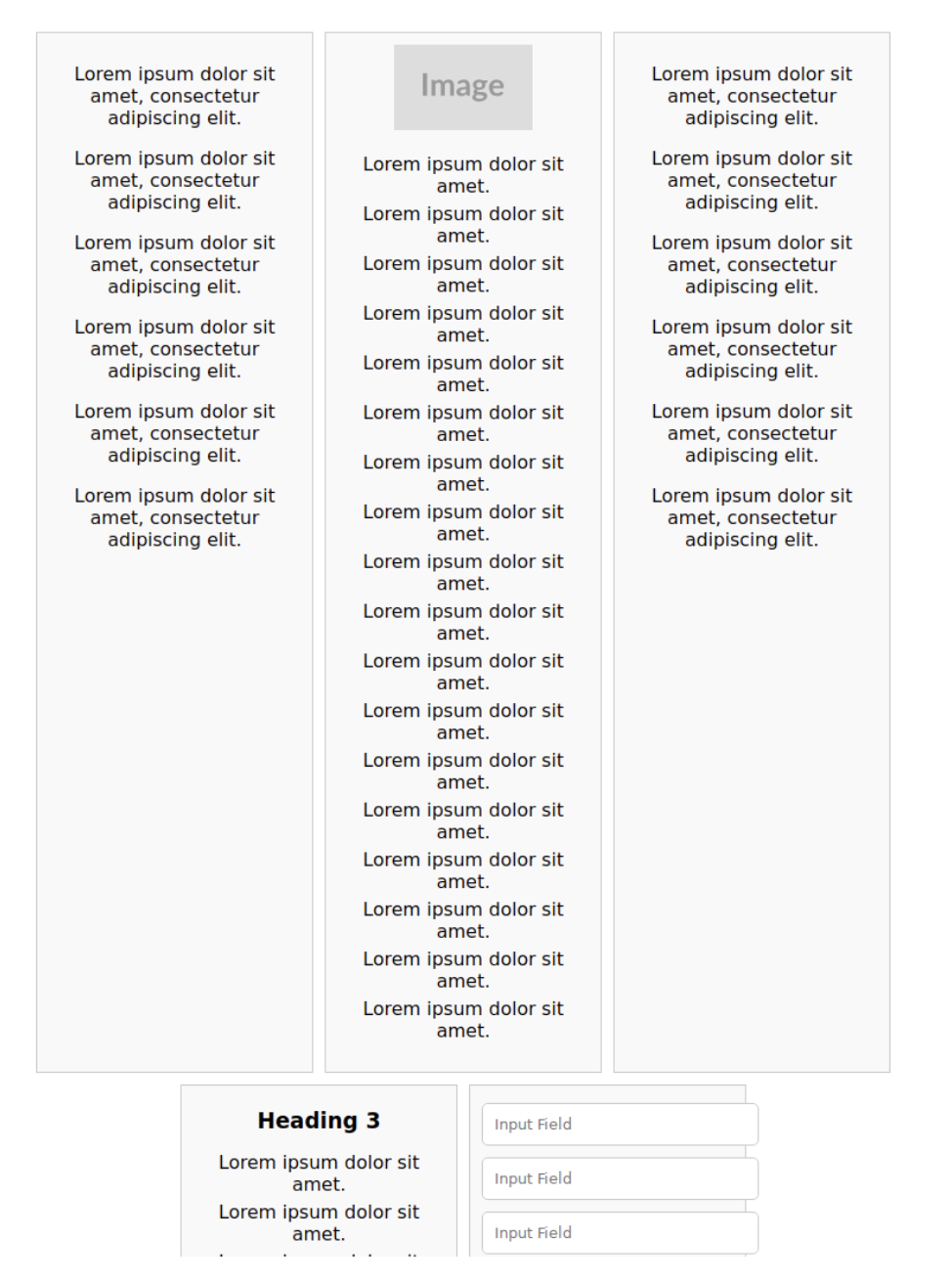}
          }
        \vspace{0.05cm}
          \centering\textbf{\footnotesize Generation}
        \end{minipage}
      \end{minipage}

      \vspace{0.05cm}
      
      \begin{tcolorbox}[
        colback=codebg,
        colframe=gray!15,
        width=1.02\textwidth,
        boxsep=2pt,
        top=1pt,
        bottom=1pt,
        left=3pt,
        right=3pt,
        enlarge left by=-0.005\textwidth,
        enlarge right by=-0.005\textwidth,
        title={\textcolor{black}{\textbf{GPT Score Evaluation on Complex Sketch}}},
        fonttitle=\footnotesize
      ]
\begin{lstlisting}[basicstyle=\scriptsize\ttfamily, escapeinside={(*}{*)}]
(*\textbf{Layout: 4.}*) Two-column layout is largely reproduced with most elements accurately placed, though an extra image placeholder appears.
(*\textbf{Spacing: 3.}*) Inconsistent spacing leads to uneven gaps that affect the overall balance.
(*\textbf{Alignment: 3.}*) Some elements are misaligned, disrupting the intended grid structure.
(*\textbf{Overall Score: 3.33}*)
\end{lstlisting}

      \end{tcolorbox}
    \end{minipage}
    
    \end{tcolorbox}
  \end{subfigure}
  
  \caption{Examples of the \textbf{failure cases on the \mockupNoStyle{} task} for the open-source model (\InternVLTwoFiveEightB) for both simple and complex mockups.}
  \label{fig:mockup2code_failure_case_internvl}
\end{figure*}

\begin{figure*}[h]
  \begin{subfigure}{\textwidth}
    \begin{tcolorbox}[
      colback=white,
      colframe=black!5,
      arc=2pt,
      boxrule=0.5pt,
      width=\textwidth,
      top=1pt,
      bottom=1pt
    ]
    
    \begin{tcolorbox}[
      colback=lightblue,
      colframe=lightblue,
      width=1.02\textwidth,
      boxsep=2pt,
      top=1pt,
      bottom=1pt,
      left=4pt,
      right=4pt,
      enlarge left by=-0.005\textwidth,
      enlarge right by=-0.005\textwidth
    ]
      \textcolor{ecologygreen}{\textbf{\oOne}}
    \end{tcolorbox}
    
    \begin{minipage}{\textwidth}
      \begin{minipage}{\textwidth}
        \begin{minipage}{0.40\textwidth}
          \centering
          \parbox[c][4.3cm][t]{0.98\textwidth}{
            \centering
            \vspace{0.1cm}
            \includegraphics[height=3.7cm]{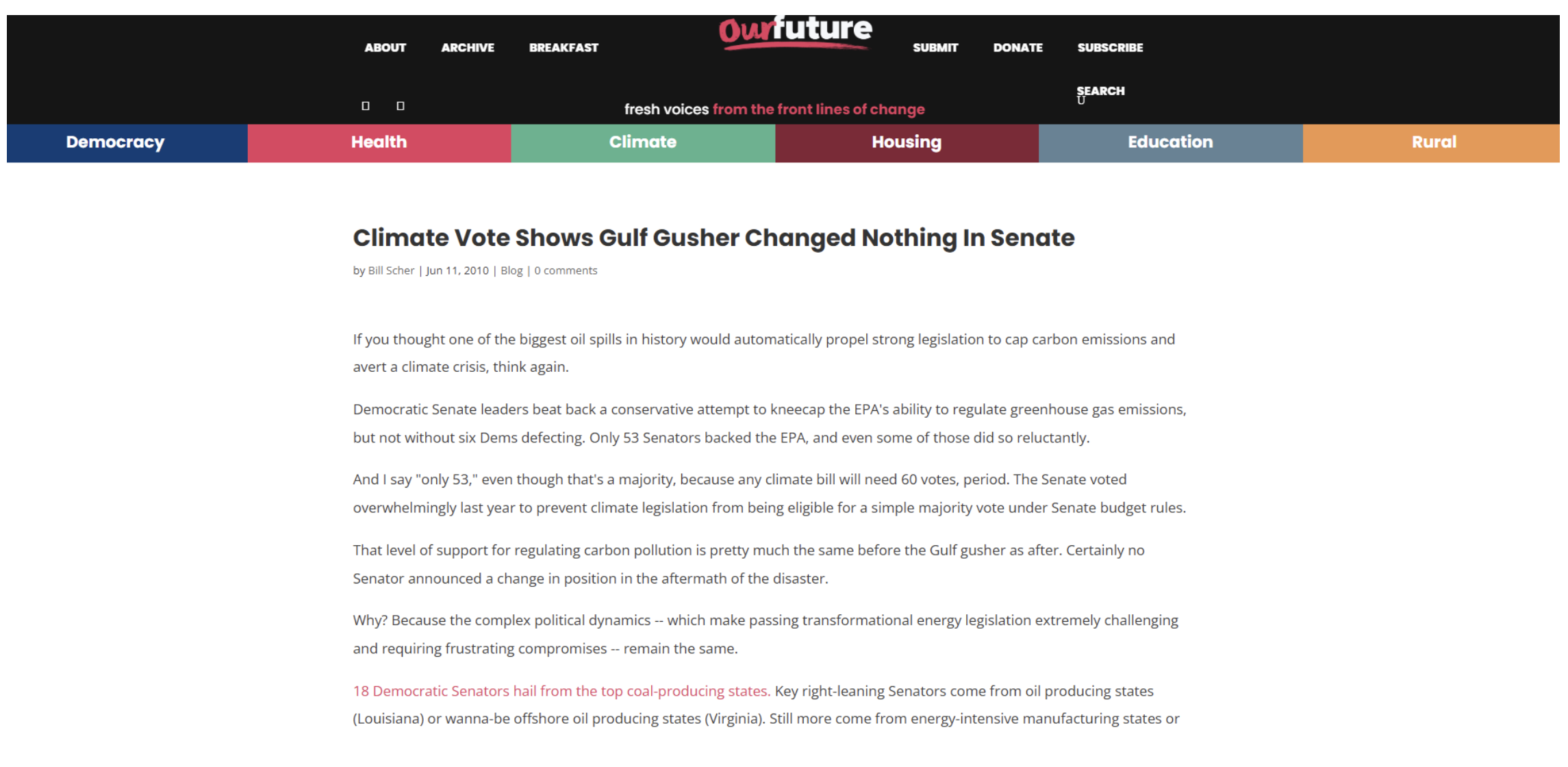}
            \vspace{0.1cm}
          }
          \centering\textbf{\footnotesize Original Page}
        \end{minipage}
        \begin{minipage}{0.18\textwidth}
          \centering
          \parbox[c][4.3cm][t]{0.98\textwidth}{
            \centering
            \vspace{0.1cm}
            \includegraphics[height=3.7cm]{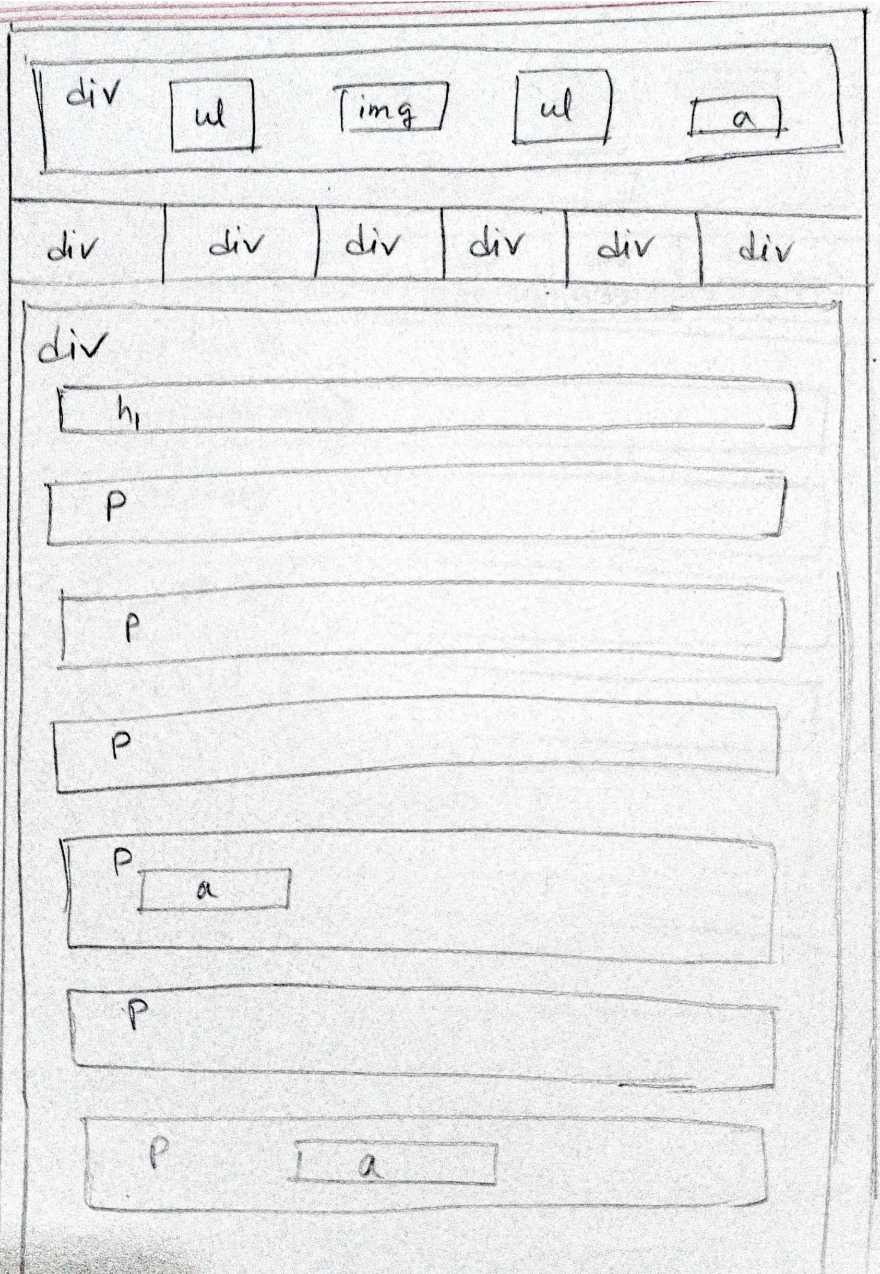}
            \vspace{0.1cm}
          }
          \centering\textbf{\footnotesize Mockup Image}
        \end{minipage}
        \begin{minipage}{0.32\textwidth}
          \centering
          \parbox[c][4.3cm][t]{0.98\textwidth}{
            \centering
            \vspace{0.1cm}
            \includegraphics[height=3cm]{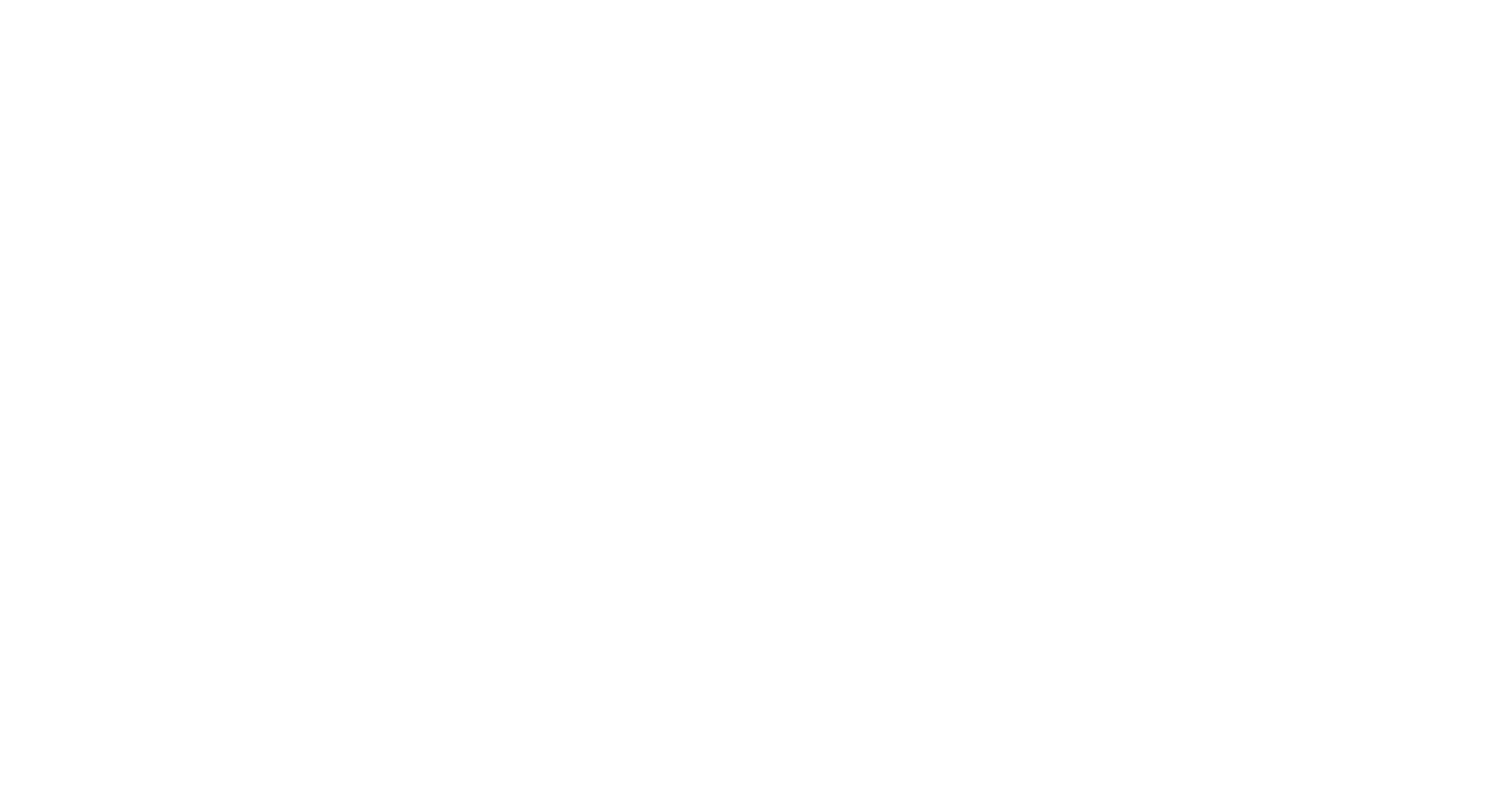}
            \vspace{0.1cm}
          }
          \centering\textbf{\footnotesize Generation}
        \end{minipage}
      \end{minipage}

      \vspace{0.05cm}
      
      \begin{tcolorbox}[
        colback=codebg,
        colframe=gray!15,
        width=1.02\textwidth,
        boxsep=2pt,
        top=1pt,
        bottom=1pt,
        left=3pt,
        right=3pt,
        enlarge left by=-0.005\textwidth,
        enlarge right by=-0.005\textwidth,
        title={\textcolor{black}{\textbf{GPT Score Evaluation}}},
        fonttitle=\footnotesize
      ]
        \begin{lstlisting}[basicstyle=\scriptsize\ttfamily, escapeinside={(*}{*)}]
(*\textbf{Alignment: 2.}*) Key elements (e.g., input box) misaligned, deviating from intended grid.
(*\textbf{Layout: 2.}*) Two-column structure poorly represented, essential sections missing/merged.
(*\textbf{Spacing: 2.}*) Uneven element distribution results in inconsistent spacing and imbalance.
(*\textbf{Overall Score: 2}*)
        \end{lstlisting}
      \end{tcolorbox}
    \end{minipage}
    
    \end{tcolorbox}
  \end{subfigure}
  
  \vspace{0.1cm}
  
  \begin{subfigure}{\textwidth}
    \begin{tcolorbox}[
      colback=white,
      colframe=black!5,
      arc=2pt,
      boxrule=0.5pt,
      width=\textwidth,
      top=1pt,
      bottom=1pt
    ]
    
    \begin{tcolorbox}[
      colback=lightblue,
      colframe=lightblue,
      width=1.02\textwidth,
      boxsep=2pt,
      top=1pt,
      bottom=1pt,
      left=4pt,
      right=4pt,
      enlarge left by=-0.005\textwidth,
      enlarge right by=-0.005\textwidth
    ]
      \textcolor{ecologygreen}{\textbf{OpenAI-o1}}
    \end{tcolorbox}
    
    \begin{minipage}{\textwidth}
      \begin{minipage}{\textwidth}
        \begin{minipage}{0.39\textwidth}
          \centering
          \parbox[c][4.1cm][t]{0.98\textwidth}{
            \centering
            \vspace{0.1cm}
            \includegraphics[height=3.9cm]{Figures-appendix/3-original.png}
          }
          \vspace{0.05cm}
          \centering\textbf{\footnotesize Original Page}
        \end{minipage}
        \begin{minipage}{0.19\textwidth}
          \centering
          \parbox[c][4.1cm][t]{0.98\textwidth}{
            \centering
            \vspace{0.1cm}
            \includegraphics[width=2.7cm]{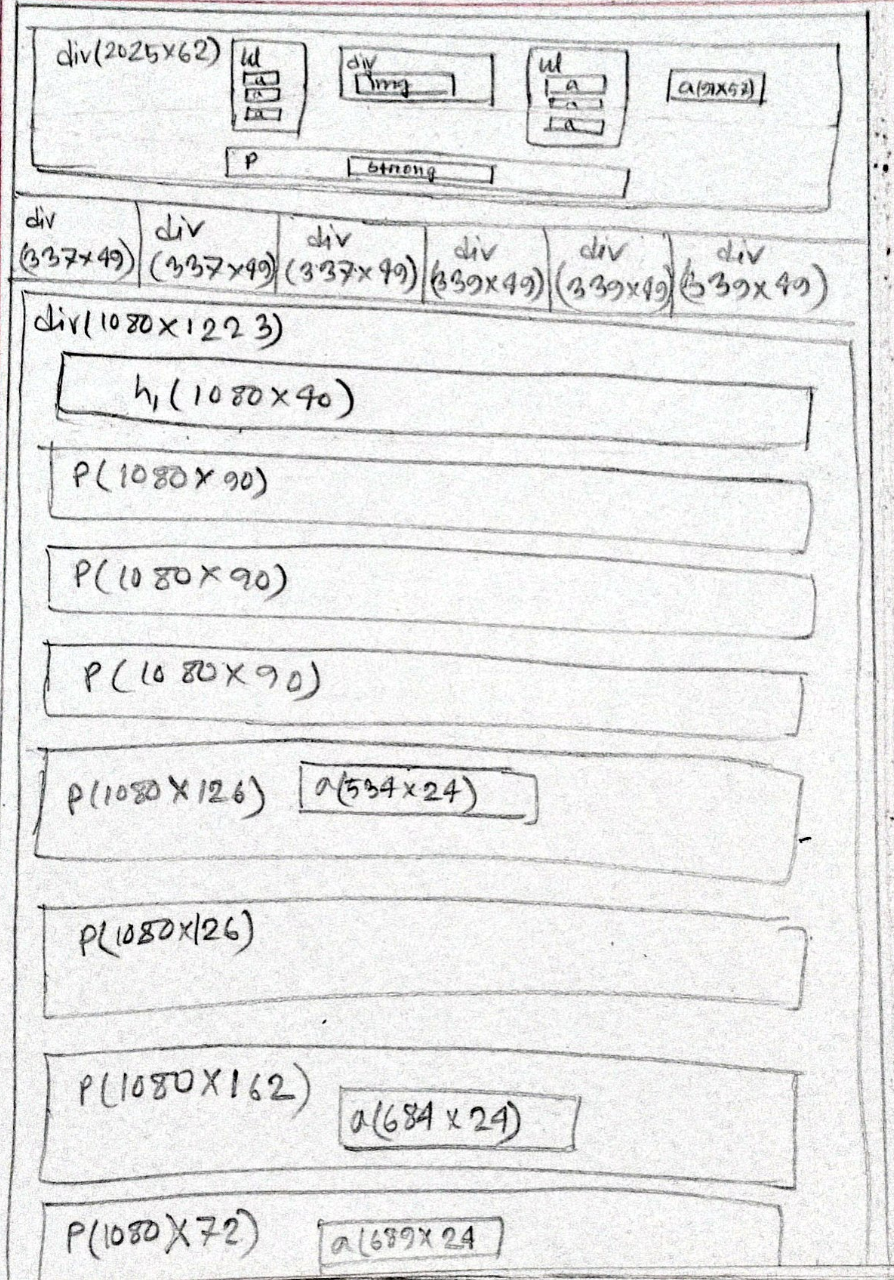}
          }
        \vspace{0.05cm}
          \centering\textbf{\footnotesize Mockup Image}
        \end{minipage}
        \begin{minipage}{0.3\textwidth}
          \centering
          \parbox[c][4.1cm][t]{0.98\textwidth}{
            \centering
            \vspace{0.1cm}
            \includegraphics[height=2.7cm]{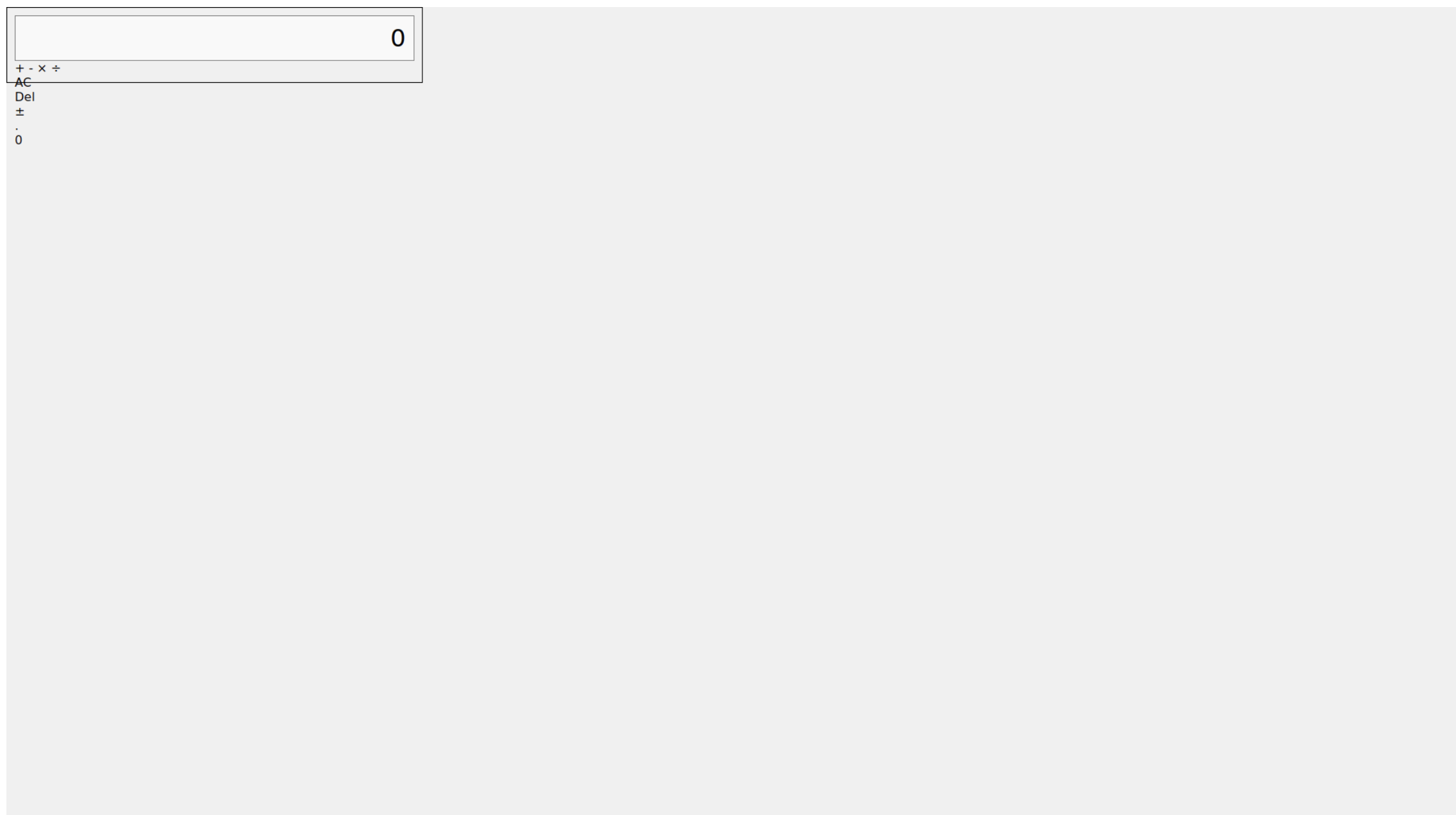}
          }
        \vspace{0.05cm}
          \centering\textbf{\footnotesize Generation}
        \end{minipage}
      \end{minipage}

      \vspace{0.05cm}
      
      \begin{tcolorbox}[
        colback=codebg,
        colframe=gray!15,
        width=1.02\textwidth,
        boxsep=2pt,
        top=1pt,
        bottom=1pt,
        left=3pt,
        right=3pt,
        enlarge left by=-0.005\textwidth,
        enlarge right by=-0.005\textwidth,
        title={\textcolor{black}{\textbf{GPT Score Evaluation}}},
        fonttitle=\footnotesize
      ]
        \begin{lstlisting}[basicstyle=\scriptsize\ttfamily, escapeinside={(*}{*)}]
(*\textbf{Alignment: 2.}*) The main content is misaligned; also, key elements like the aside and buttons are missing.
(*\textbf{Layout: 2.}*) The intended two-column structure is replaced by a single-column layout, significantly deviating from the sketch.
(*\textbf{Spacing: 2.}*) Inconsistent spacing and proportions arise from the absence of the aside and smaller placeholders.
(*\textbf{Overall Score: 2}*)
        \end{lstlisting}
      \end{tcolorbox}
    \end{minipage}
    
    \end{tcolorbox}
  \end{subfigure}
  
  \caption{Examples of the \textbf{failure cases on the \mockupNoStyle{} task} for the best closed-source model (\oOne) for both simple and complex mockups.}
  \label{fig:mockup2code_failure_case_o1_simp_complex}
\end{figure*}

We provide several examples illustrating the performance of different models on the \mockupNoStyle{} task, including both the input mockups and the generated outputs. Figure~\ref{fig:mockup2code_failure_case_o1_and_internvl} shows failure cases where both the best closed-source model (\oOne) and the best open-source model (\InternVLTwoFiveThirtyEightB) struggled to accurately reproduce the designs. In contrast, Figure~\ref{fig:mockup2code_success_case_o1_1} highlights success cases for the \oOne{} model, demonstrating its ability to handle both simple and complex mockups effectively. Additionally, Figures~\ref{fig:mockup2code_failure_case_internvl} and~\ref{fig:mockup2code_failure_case_o1_simp_complex} present failure cases specifically for the open-source model \InternVLTwoFiveEightB{} and closed-source model \oOne{}, emphasizing areas where it underperforms on varying mockup complexities.

